\documentclass[lettersize,journal]{IEEEtran}
\usepackage{amsmath,amsfonts}
\usepackage{algorithmic}
\usepackage{algorithm}
\usepackage{array}
\usepackage{textcomp}
\usepackage{stfloats}
\usepackage{url}
\usepackage{verbatim}
\usepackage{graphicx}
\usepackage{graphics}
\usepackage{cite}
\usepackage{pifont}
\usepackage{caption}
\usepackage{subcaption}
\newcommand{\cmark}{\ding{51}}%
\newcommand{\xmark}{\ding{55}}%
\newcommand{\tabincell}[2]{\begin{tabular}{@{}#1@{}}#2\end{tabular}}
\newcommand{\res}[2]{{#1}$\pm${#2}}

\newcommand{\etc}{\textit{etc.}}
\newcommand{\eg}{\textit{e.g.}}
\newcommand{\etal}{\textit{et al.}}
\newcommand{\midrule}{\hline}
\newcommand{\toprule}{\hline}
\newcommand{\bottomrule}{\hline}

\usepackage{multirow}
\usepackage{amssymb}
\usepackage{pifont}
\usepackage{hyperref}

\begin{document}

\title{Anomaly Crossing: New Horizons for Video Anomaly Detection as Cross-domain Few-shot Learning}

\author{Guangyu Sun$^{1*}$\thanks{* Equal Contribution}\quad\quad Zhang Liu$^{2*}$\quad\quad Lianggong Wen$^2$\quad\quad Jing Shi$^1$\quad\quad Chenliang Xu$^1$\\
$^1$University of Rochester \quad\quad $^2$Corning Inc.\\
{\tt\small gsun6@ur.rochester.edu, \{j.shi,chenliang.xu\}@rochester.edu} \\
{\tt\small \{LiuZ2, WenLB\}@corning.com} \quad }



\maketitle
 Developer Category
\begin{abstract}
 Video anomaly detection aims to identify abnormal events that occur in videos. Since anomalous events are relatively rare, it is not feasible to collect a balanced dataset and train a binary classifier to solve the task. Thus, most previous approaches learn only from normal videos using unsupervised or semi-supervised methods. Obviously, they are limited in capturing and utilizing discriminative abnormal characteristics, which leads to compromised anomaly detection performance. In this paper, to address this issue, we propose a new learning paradigm by making full use of both normal and abnormal videos for video anomaly detection. In particular, we formulate a new learning task: cross-domain few-shot anomaly detection, which can transfer knowledge learned from numerous videos in the source domain to help solve few-shot abnormality detection in the target domain. Concretely, we leverage self-supervised training on the target normal videos to reduce the domain gap and devise a meta context perception module to explore the video context of the event in the few-shot setting. Our experiments show that our method significantly outperforms baseline methods on DoTA and UCF-Crime datasets, and the new task contributes to a more practical training paradigm for anomaly detection.
\end{abstract}


\begin{figure}[t]
  \centering
  \includegraphics[width=0.8\linewidth ]{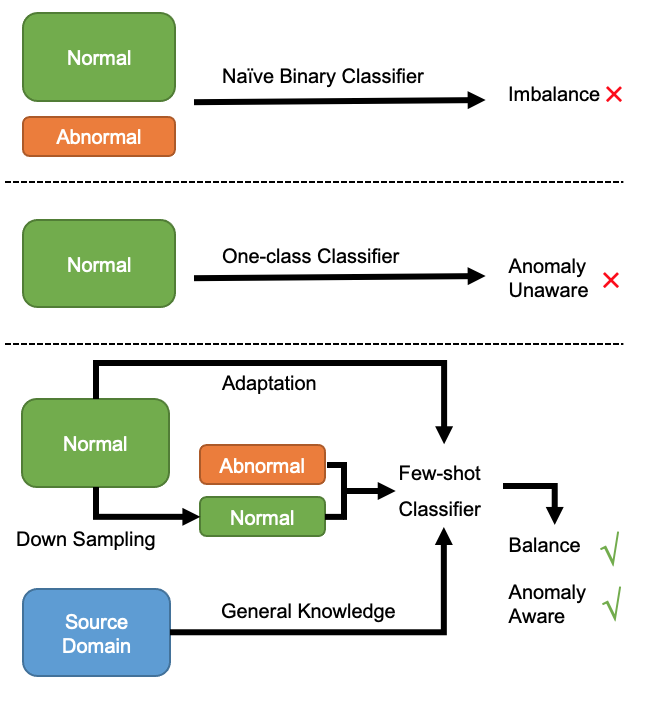}
  \caption{Different training paradigms
  for video anomaly detection: Training a naive binary classifier will fail due to the severe data imbalance. Most existing methods focus on training a one-class classifier based on only normal samples, which is anomaly unaware. To leverage the abnormal samples to learn an anomaly-aware classifier in a balanced dataset, we propose a new pipeline to train a few-shot classifier.}
  \label{fig:intro}
  \vspace{-8mm}
\end{figure}

\section{Introduction}

Video anomaly detection~\cite{chalapathy2019deep,kiran2018overview} has broad application potential in security~\cite{zhong2019graph}, industry~\cite{atha2018evaluation}, and healthcare~\cite{esteva2017dermatologist}.
The goal is to identify if there is an abnormal event that happened in a given video. This goal can be formulated as a binary classification problem. However, one unique property of anomaly detection is that the distributions of normal and anomalous events are quite different, leading to a severe data imbalance. 
Training a naive binary classifier on such an imbalanced dataset will always steer the classifier to give a negative prediction, thus failing to detect the anomaly (Fig.~\ref{fig:intro} top). 
To tackle unbalanced data in video anomaly detection, different training paradigms are proposed. 
An intuitive solution is to re-sample or down-sample to get an equal amount of normal and abnormal samples. However, naive re-sampling or down-sampling is insufficient for the classifier to learn all anomaly patterns under extreme data imbalance. Another solution is to collect more anomalous data and construct a balanced dataset~\cite{he2018anomaly,huo2012abnormal,sultani2019realworld,zhong2019graph,lv2021localizing,park2020learning}. However, it is not possible to collect sufficient anomalous data in many real-world applications, \emph{e.g.}, industrial production pipeline, as the defective rate can be low and the shutdown of machines suffers monetary loss.

Another line of work completely discards the abnormal data and uses only normal training samples since anomalous events are rare in practice, formulating the problem as single-class classification~\cite{kiran2018overview,cheng2015video} (Fig.~\ref{fig:intro} middle).
Such a paradigm tries to approximate the distribution of normal samples and regards the out-of-distribution samples as anomalous; however, not all normal behaviors can be observed. Thus, some normal events may deviate from the approximated distribution, triggering false anomalies.

These two lines of works are either infeasible or neglect the discriminative abnormal characteristics, leading to compromised anomaly detection performance.
Therefore, a new training paradigm is needed to construct a balanced dataset and 
get an anomaly-aware classifier for anomaly detection.

We propose a new task \emph{cross-domain few-shot video anomaly detection}~(CD-FSVAD) for anomaly detection (Fig.~\ref{fig:intro} bottom). To address the insufficiency of abnormal samples, we resort to few-shot learning~\cite{fei2006one,vinyals2016matching} to better utilize the limited abnormal samples. 
However, recent few-shot learning largely relies on extensive annotated data for meta-learning, where base classes are sampled from the same domain as the novel classes~\cite{vinyals2016matching,finn2017model}, clearly infeasible for anomaly detection. On the other hand, general knowledge may be feasibly gained from outside the in-domain data. Inspired by cross-domain few-shot learning (CD-FSL)~\cite{guo2020broader}, we expect to learn such general knowledge from another large-scale dataset as the source domain. In this new task, we can leverage a large amount of source-domain videos to help target-domain anomaly detection with only a few abnormal videos and abundant normal videos, which is a more realistic setting for traffic accident detection, surveillance, industrial production pipeline, \etc.

This new task is challenging owing to the large gap between the source and the target domain and the data imbalance between normal and abnormal samples.
To tackle these challenges in CD-FSVAD, we devise a novel baseline for anomaly detection called \textit{Anomaly Crossing}, where knowledge will be learned from the source domain, adapted from the normal samples, and finally fit the target domain through only a few abnormal samples. Two novel modules, Domain Adaptation Module (DAM) and Meta Context Perception Module (MCPM), are proposed in our pipeline.
To reduce the domain gap, we devise the DAM that can use a large amount of target-domain normal videos. Then, inspired by the power of self-supervised learning (SSL)~\cite{alwassel2019self}, 
given the backbone trained from source-domain videos, we fine-tune it via SSL using normal videos in the target domain to achieve the unsupervised domain adaptation.
Furthermore, the video temporal context is crucial for anomaly detection, and we also expect to enable the model with adaptive context modeling ability in different novel scenes.
However, the events in different scenes may need different temporal context modeling; for example, the pedestrian and vehicle surveillance has different motion patterns; thus, they need different context modeling.
Therefore, the model should automatically adapt its contextual modeling to novel events.
To achieve this goal, we propose the MCPM consisting of a learnable Graph Convolution Network (GCN) to perceive the meta-context that can be adapted to the target domain under a few-shot setting. 
We evaluate the performance of our pipeline on two different target domains with diverse scenes: a traffic dataset DoTA~\cite{yao2020dota} and a surveillance dataset UCF-Crime~\cite{sultani2019realworld}.
Our pipeline consistently outperforms the comparison methods on both datasets, achieving 15\% higher accuracy than the best-compared ones on the DoTA dataset.

The contributions of this paper can be summarized as follows. 
Firstly, we introduce a more practical CD-FSVAD task to address the extreme data imbalance issue in anomaly detection and propose a novel pipeline Anomaly Crossing as a new training paradigm to tackle this task.
Secondly, we propose to leverage video self-supervised learning tasks on the target-domain normal samples to reduce the gap between source and target domain.
Thirdly, we enable our Meta Context Perception Module to adapt its contextual modeling to different scenes automatically.
Finally, Anomaly Crossing consistently outperforms the comparison methods on DoTA and UCF-Crime datasets.

\begin{figure*}[t]
  \centering
  \includegraphics[width=\linewidth]{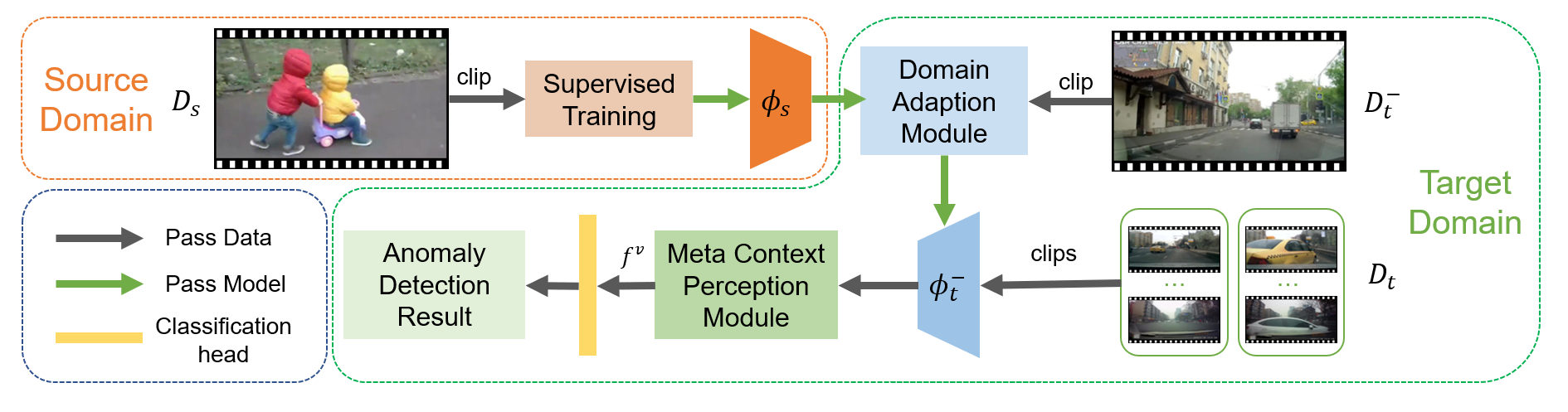}
  \caption{The pipeline of Anomaly Crossing.  The left of the first row shows a supervised training process to get the encoder $\phi_s$ in the source domain. The right of the first row shows the Domain Adaption Module (DAM) of the normal samples in the target domain to get the adapted encoder $\phi_t^-$. The second row shows the meta-testing process in the target domain. Clips of each video in the support set are fed into $\phi_t^-$, and the clip features will be passed to the Meta Context Perception Module (MCPM) to get the video feature $f^v$. Anomaly detection result will be output by a classification head given $f^v$. Blue and green arrows mean passing the data or entire model into the module. 
  }

  \vspace{-5mm}
  \label{fig:pipeline}
\end{figure*}

\section{Related Work}

\noindent\textbf{Anomaly Detection.} Anomaly detection is a challenging problem that has been studied for years~\cite{fei2006one,kratz2009anomaly,zhao2011online,li2013anomaly,tian2021weaklysupervised,a,c,g,h}. Methods out of different perspectives are proposed. Among them the mainstream methods completely discard the anomalous data and try to learn a representation for only normal events due to the rarity of abnormal events and the intrinsic data imbalance. Then,are construction loss is used as anomaly score to identify abnormal events~\cite{a,b,c,d,e} or a one-class classier is used instead
~\cite{kiran2018overview,cheng2015video,sabokrou2018adversarially}. Another stream of works design models for video prediction learnings and detect anomaly based on difference between the predicted frame and the observed frame~\cite{ramachandra2020survey,ramachandra2020survey,f}. All these methods suffer limited accuracy especially for subtle abnormal cases which are similar with normal cases due to the ignorance of patterns in abnormal samples. However, such kind of cases are quite common in many applications such as in industry. Furthermore, these methods are lack of the capability to evolve when more abnormal samples are collected later on. To resolve these issues, several recent works try to leverage abnormal samples following weekly supervised setting~\cite{feng2021mist,liu2019few,tian2021weaklysupervised,g,h}. However, information in abnormal samples is still not fully exploited due to the limitation of weekly supervised learning. Different with these methods we formalize anomaly detection as a few-shot classification problem under supervised learning setting. Meanwhile a Domain Adaptation Module (DAM) is designed to adapt knowledge learned from a source domain dataset (large) to the target domain (small) by leveraging excessive normal samples. Thus, information in both abnormal and normal samples is fully exploited, and a deep neural network (R(2+1)D) backbone is possible to be used to extract better spatial-temporal representations of video clips instead of shallow networks used in most of the existing works. Finally, a Meta Context Perception Module (MCPM) is designed to enhance the understanding of scene context which is critical for VAD especially for cases with complicated behaviors and patterns instead of a simple reconstruction loss or a FC classification head.

\noindent\textbf{Few-shot Video Classification.} 
 Few-shot Video Classification. Few-shot learning is usually achieved by meta-training with a large amount of labeled data~\cite{NIPS2016_90e13578,finn2017model,ravi2016optimization}. Based on such data, most of the methods are metric-based to extract a generalized video metric~\cite{lu2021fewshot}. These metrics can be obtained by pooling~\cite{careaga2019metric,kumar2019protogan,xian2020generalized,zhu2021few,ben2020taen,cao2020few}, adaptive fusion~\cite{fu2020depth,bo2020few}, dynamic images~\cite{tan2019learning}, or attention~\cite{bishay2019tarn,zhang2020few}.
 Significantly, with the help of progress in video representation learning, few-shot learning can be achieved even without meta-training. For example, after training a video encoder(e.g., 3D-CNN, R(2+1)D-CNN) on largescale data with supervision, few-shot learning can be achieved by simply learning a classifier using the few shots~\cite{xian2020generalized}. Other researchers focus on the feature extraction from video using Graph Neural Network (GNN) to get a better feature representation of the video~\cite{i} or further predict augmented video features~\cite{j}. However, in CD-FSVAD, the training dataset on the source domain has an extensive domain gap with the target domain. Therefore, based on our experiments, training a classifier on the target domain straightforwardly will not work well as before. As a result, our method includes two modules to mitigate the domain gap and achieve few-shot video classification in a cross-domain fashion.

\noindent\textbf{Cross-domain Few-shot Classification.} 
Cross-domain Few-shot Classification. CD-FSL is recently proposed mainly in the image domain~\cite{guo2020broader,k,l}. 
The goal is to strengthen the few-shot model generalizability when the base classes are sampled from the different domains as the novel classes. Most existing methods focus on the meta training process in the source domain, including ensemble modeling~\cite{liu2020feature}, meta fine-tuning~\cite{cai2020cross}, transductive multi-head model~\cite{jiang2020transductive}, representation fusion~\cite{adler2021crossdomain}, and learned feature-wised transformation~\cite{tseng2020cross}. 
These methods are domain-agnostic so that they neglect the information on the target domain. Meanwhile,
\cite{phoo2020self} is a concurrent work sharing the most similar spirit with our methods to leverage SSL on the target domain, which requires many unlabeled target-domain training data. In addition, however, we explore it in the anomaly detection task and research the more complex video modality since the video domain CD-FSL is nearly unexplored. ~\cite{gao2020pairwise} concentrates on CD-FSL for action recognition problems, which does not leverage the large-amount normal sample on the target domain as our method for domain adaptation. Compared with all the existing work, our method takes both source domain and target domain into considerations with the large-amount normal samples, while fills the blank of the study of CD-FSL on the video domain.

\section{Method}
Considering a common video anomaly detection task, \eg, the traffic accident detection, there are sufficient normal samples but only a few abnormal samples available. Therefore, the goal is to learn a model to classify whether an input video contains an anomaly event or not.
A naive binary classification model trained on all samples will fail due to extreme data imbalance. Down-sampling is a common way~\cite{aggarwal2017introduction} for data balance, but simple down-sampling neglects the considerable normal data. To learn from fewer samples, we further propose to regard anomaly detection as a few-shot detection problem, formulated as a 2-way-$K$-shot classification problem. However, typical few-shot learning~\cite{NIPS2016_90e13578,finn2017model,ravi2016optimization} requires a large dataset for meta-training whose classes are in the same domain of novel classes, which is infeasible for our anomaly detection problem since there are insufficient abnormal videos to support meta-training. To confront the few-shot challenge, we proposed to inflate our training data from a different domain where a large amount of videos are available, \eg, human action videos~\cite{ghadiyaram2019large,kay2017kinetics}.
Nonetheless, directly applying the standard meta-learning approach on a target-independent domain is incapable of solving the target-domain problem; hence, we propose a new pipeline, Anomaly Crossing (Section \ref{AnomalyCrossing}), as a baseline for this new formulation. Our Domain Adaptation Module (Section \ref{DAM}) leverages the normal samples to mitigate the domain gap via self-supervised contrastive learning. Furthermore, concerning the impact of video context for anomaly detection, we propose a Meta Context Perception Module (Section \ref{MCPM}) to perceive the context-aware video feature.

\begin{algorithm}
\caption{Training of the Anomaly Crossing Pipeline}\label{alg:pipeline}
\begin{algorithmic}[1]

\STATE \textbf{Training on the source domain:}

     Learn a clip-level encoder $\phi_s$ on $D_s$.

\STATE \textbf{Domain Adaptation Module:}

    Perform adaptation $\mathcal{A}$ on $D_t^-$ and get a refined clip-level encoder $\phi_t^-$.

\STATE \textbf{Meta Context Perception Module:}
    \begin{enumerate}[1]
        \item Sample $n$ clips from each video $x_t \in X_t$ and extract corresponding clip-level features $f^{c} = \{\phi_t^-(x_{t,n})$, for $x_t$ in $X_t\}$ from $\phi_t^-$.
        \item Extract corresponding video-level feature $f^{v}$ of $f^c$ from MCPM $\mathcal{P}$.
    \end{enumerate}
\STATE \textbf{Training:}

    Pass $f^v$ to the classification head $c_t$ and train the parameters in MCPM and $c_t$ by the loss between predicted anomaly detection result $\hat{Y_t}$ and ground truth $Y_t$ in $D_t$. 

\end{algorithmic}

\end{algorithm}

\subsection{Problem Formulation}

In the perspective of few-shot learning (FSL), the video anomaly detection task can be formulated as follows.
Here we refer to the anomaly detection dataset as the target domain and the other dataset (\eg human action) as the source domain.
Given a target-domain dataset of sufficient normal samples and $K$ abnormal samples, we denote all normal samples as a sub-dataset $D_t^-$. 
Together with the abnormal samples, we down-sample $D_t^-$ to the same size $K$ to construct a 2-way-$K$-shot dataset $D_t=\{(x_t,y_t)\in X_t \times Y_t\}$ for few-shot learning. 
To enable the FSL, we include a source-domain dataset $D_s$  composed of sufficient labeled samples $(x_s,y_s)\in X_s\times Y_s$ to learn general knowledge for classification.
This setting corresponds to the cross-domain few-shot learning~(CD-FSL), where the $D_s$ and $D_t$ have a large domain gap.

As a result, the Cross-Domain Few-shot Anomaly Detection (CD-FSVAD) task is formulated to learn a good classifier $X_t \rightarrow Y_t$ based on $D_s$,  $D_t^-$ and $D_t$.  

\subsection{The Anomaly Crossing Pipeline}\label{AnomalyCrossing}

Our Anomaly Crossing pipeline leverages both normal and abnormal samples on the target domain. 
First, as \cite{xian2020generalized} indicates that meta-training is not a necessity if a video encoder is trained on large-scale supervised data, we also directly train a source-domain video clip encoder for a good initialization of the model backbone.
Then, normal samples are leveraged for anomaly detection in a new horizon that adapts the backbone learned from source domain to the target domain instead of training from scratch. 
Finally, the down-sampled normal samples paired with the few abnormal samples are used to train only the classification head, as known as a meta-testing stage.
The pipeline is shown in Fig.~\ref{fig:pipeline} and described in Algorithm~\ref{alg:pipeline}.

\noindent\textbf{Training on the source domain.} 
We conduct supervised training on the source domain to learn a video encoder $\phi_s$ entailing the general knowledge for video classification. This process can be formalized as:
\begin{equation}
\label{eq:source}
    \min_{\phi_s,c_s} \mathcal{L}_s(c_s(\phi_s(X_s)),Y_s) \enspace,
\end{equation}
where $\mathcal{L}_s$ is the loss function for supervised training on the source domain, and $c_s$ is the classification head. An alternative is meta-training instead of standard training; however, our experiments show that these two training fashions are comparable in our cross-domain setting (Section~\ref{sec:eval_res}).
Despite that we learn from source-domain in a supervised way following the insight of \cite{xian2020generalized}, unsupervised learning fashion is also feasible for future exploration.

\noindent\textbf{Domain adaptation.}
Domain adaptation enables the knowledge learned from the source domain to better assist the target domain for anomaly detection.
Typical domain adaptation approaches require a large quantity of target domain samples for training~\cite{gebru2017fine,motiian2017unified}, which is infeasible for our case where our target domain $D_t$ only contains few-shot abnormal samples.
However, apart from $D_t$, we have numerous normal samples $D_t^-$ in the target domain, which carries rich domain information that might benefit domain adaptation, which is a unique property for the anomaly detection task.
Therefore, we devise a novel Domain Adaptation Module (DAM) to adapt the backbone parameter of $\phi_s$ to $\phi_t^-$ based on the normal samples $D_t^-$, so as to adapt representations on the source domain to the target domain.
This process can be formalized as:
\begin{equation}
\label{eq:ref}
    \phi_t^- = \mathcal{A} (\phi_s,D_t^-) 
    \enspace,
\end{equation}
where $\mathcal{A}$ refers to the DAM.

\noindent\textbf{Meta-testing on the target domain.} 
Despite that $\phi_t^-$ involves the target domain information, it is unaware of the abnormal information.
Therefore, we need a further fine-tuning of the model using $D_t$ to capture the abnormal pattern, referred to as the meta-testing step in FSL.
However, the typical meta-testing stage just adapts the last classification head with the frozen backbone, which is commonly a fully connected (FC) layer~\cite{bertinetto2018meta}.
However, a simple tuning on the FC layer may not well capture the temporal dynamics of the abnormal pattern, which is critical for detecting video anomalies such as abnormal vehicle speed.
Instead,  we further leverage a Meta Context Perception Module (MCPM) that can refine the input video clip features into a spatio-temporal contextualized video feature, and the whole parameter of MCPM are tuned in this stage, 
as shown in Eq.~(\ref{eq:f}). 
The parameter in MCPM $\phi_t$ constructs a mapping from $X_t$ to $f^v(X_t)$ ], as shown in Eq.~(\ref{eq:phi}). Therefore, the objective of this process is as shown in Eq.~(\ref{eq:target}).
\begin{equation}
\label{eq:f}
     f^c(x_t) = \{\phi_t^-(x_{t,i})\}, \enspace i = 1, 2, \dots n \enspace ,\\
\end{equation}
\begin{equation}
\label{eq:phi}
     \phi_t(X_t)= f^v(X_t) = \mathcal{P}(f^c(X_t))\enspace, \\
\end{equation}
\begin{equation}
\label{eq:target}
    \min_{\phi_t,c_t} \mathcal{L}_t(c_t(\phi_t(X_t)),Y_t)\enspace,
\end{equation}
where $\mathcal{P}$ refers to the MCPM, $x_{t,i}$ refers $i_{th}$ sampled clip in video $x_t \in X_t$, $\mathcal{L}_t$ is the loss function for classification on target domain, and $c_t$ is the classification head.

\begin{figure}[t]
  \centering
  \includegraphics[width=\linewidth]{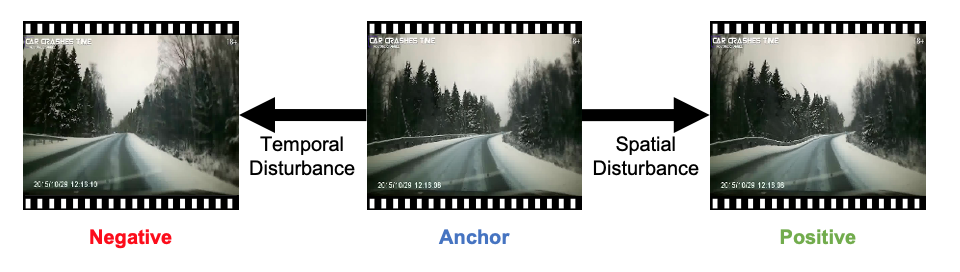}
  \caption{Construction of samples for self-supervised learning. A spatial disturbance will be performed for a positive sample to gain a clip with a similar motion but a dissimilar scene. A temporal disturbance will be performed for the negative sample to gain a clip with a similar scene but dissimilar motion.}
  \label{fig:ssl}
  \vspace{-5mm}
\end{figure}

\subsection{Domain Adaptation Module}\label{DAM}

The goal of the Domain Adaption Module (DAM) is to adapt the learned knowledge from the source domain to the target domain. 
Since we only use the large number of normal samples in the target domain for adaptation, it is an unsupervised domain adaptation.
Despite no label is available, recent progress on self-supervised learning~\cite{wang2020enhancing,yao2020video} on large scale video dataset has exhibited excellent representation and generalization capabilities.
Inspired by this, a most straightforward idea is to re-train the $\phi_s$ on $D_t^-$ using state-of-the-art self-supervised training.
However, such an idea is intuitively risky because the knowledge learned from $D_t^-$ might overwrite the modeling ability of $\phi_s$ due to well-known catastrophic forgetting~\cite{kirkpatrick2017overcoming}.
Nonetheless, we find it works well in practice, and the experiment shows that the forgetting phenomenon does not occur (Sec.~\ref{sec:exp_module_effect}).
Therefore, we select a recent SOTA self-supervised learning algorithm~\cite{wang2020enhancing} as our DAM.
Note that other self-supervised approaches might also work, and we leave the selection as future work.
The details for the DAM are as follows.


\noindent\textbf{Sample Selections.}
Given a video clip $c_0$, we randomly crop three video clips $c_1,c_2,c_3$, and we apply different transformations to them to build the data triplet containing anchor, positive, and negative samples for the later contrastive learning.
\textit{Anchor}: We apply basic augmentations including random rotation, random cropping, color jittering to $c_1$ to get the anchor sample $a$. 
\textit{Positive}: 
Every frame in the clip $c_2$ is applied with the same random warped to get the positive sample $p$, as shown in the right of Fig.~\ref{fig:ssl}.
\textit{Negative}:
We randomly shift the starting time of $c_3$ while keeping its duration to construct the negative sample $n$, as shown in the left of Fig.~\ref{fig:ssl}.

\noindent\textbf{Objective Function.}
Contrastive learning is employed to learn the representations by enhancing the affinity between the anchor and positive sample and the dissimilarity between the anchor and negative sample. 
Specifically, feed $a,p,n$ into $\phi_s$, we will get the clip feature $z_a,z_p,z_n$ and the final InfoNCE~\cite{oord2019representation} object function is written as:
$$
    \mathcal{L}_c = -\log \sum_{i=1}^{N}\frac{\exp(z_{a_i}\cdot z_{p_i})}{sim(z_{a_i}, z_{p_i}, z_{n_i})+ \sum_{j=0}^{K} \exp(z_{a_i}\cdot z_{a_j}) } \enspace,
$$
where $N$ is the numbers of video in $D_t^-$, $K$ is the number of other samples, and $sim(z_{a_i}, z_{p_i}, z_{n_i}) = \exp(z_{a_i}\cdot z_{p_i})+ \exp(z_{a_i}\cdot z_{n_i})$. 
As a result, $\phi_s$ will be adapted to $\phi_t^-$ via DAM.



\begin{figure}[t]
  \centering
  \includegraphics[width=\linewidth]{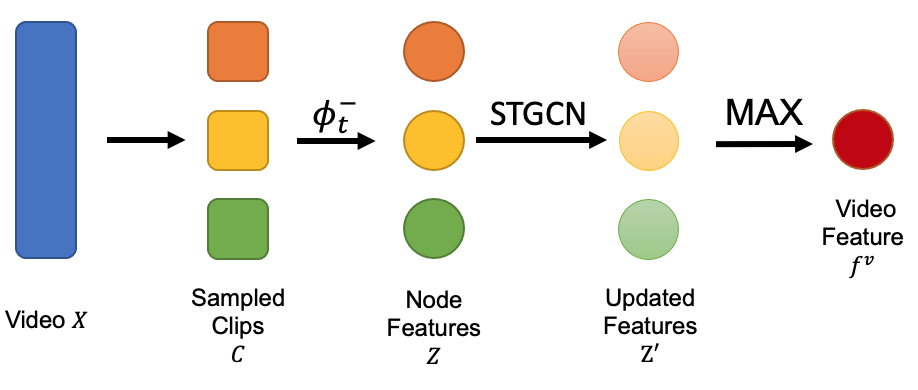}
  \caption{
    Extract context-aware video feature in Meta Context Perception Module on the target domain. Sampled video clips will be fed into $\phi_t^-$ then updated as node features in STGCN. Finally, the video feature will be given by the max of all node features.
  }
  \label{fig:ctx}
\end{figure}







\subsection{Meta Context Perception Module}\label{MCPM}

The goal of MCPM is to update video clip features given $D_t$ to get an updated video feature $f^v$ entailing temporal information to better assist the anomaly detection on the target domain. 
Considering Graph Convolutions Networks (GCN) are widely used to capture the temporal information of video~\cite{wang2018videos,xu2020g}, we apply GCN in MCPM. 
Meanwhile, the MCPM should also have the ability to be adapted to novel domains by a few shots of abnormal and normal samples. 

In addition, we desire to handle both temporal and spatial information to serve anomaly detection better. Therefore, a Semantic-Temporal Graph Convolution Network (STGCN)~\cite{xu2020g} is chosen to capture the spatio-temporal information.

\subsubsection{Construction of STGCN}
STGCN takes the clip feature as nodes and builds temporal edges $\mathcal{E}_t$ and semantic edges $\mathcal{E}_s$ based on temporal ordering and  node similarity, respectively.
For the video clip features fed into STGCN, we will sample $L$ clips $C=\{c_1,c_2,\dots,c_L\}$ with a fixed stride, then get the video clip features $z_l=\phi_t^-(c_i)$ by the video clip encoder $\phi_t^-$. 
Then we will build a video graph $\mathcal{G}=\{\mathcal{V,E}\}$, where $\mathcal{V}=\{z_l\}_{l=1}^L$ and $\mathcal{E}=\mathcal{E}_t \cup \mathcal{E}_s$. 
The overview of this module is exposed in Fig.~\ref{fig:ctx} and the details of the edges construction, the feature updating, and training of STGCN are described as follows:

\noindent\textbf{Semantic Edges $\mathcal{E}_s$.} Semantic edges are designed to connect nodes with similar semantic information. 
Each node is connected to its $k$ nearest neighbors according to their feature distances. 



\noindent\textbf{Temporal Edges $\mathcal{E}_t$.} Temporal edges are designed to connect adjacent nodes in temporal order.
A forward edge and a backward will be built for each node except the first node (no backward edge) and the last node (no forward edge). 

\noindent\textbf{Node Feature Updating.} Taking into consideration that temporal edges are in a linear structure, we use three 1d-convolution layers to update the node features from the temporal edges $Z^t$. To update the node feature from the semantic edges $Z^s$, we
use a single-layer edge convolution~\cite{wang2019dynamic} $Z^s=([Z^T, AZ^T-Z^T]W)^T$ as our graph convolution operation. $Z=\{z_1,z_2,\dots,z_L\}$ are the features of all nodes in the graph, $W$ is the trainable weight of three 2d-convolution layers, $A$ is the adjacency matrix of the graph without self-loops, and $[\cdot,\cdot]$ represents the matrix concatenation of columns. The final updated node feature is given by a fusion of $Z$, $Z^t$ and $Z^s$: $Z'=\text{ReLU}(Z+Z^t+Z^s)$.




\begin{table*}
  \caption{\textbf{2-way 5-shot result of different anomaly types in DoTA.} ST refers to ``Start-stop or stationary'', AH refers to ``Moving ahead or waiting'' \etc. The full explanation of the abbreviation is shown in Appendix~B.
The result under each type means the meta-testing set will be constructed only by this type. For example, ``All Types'' means the meta-testing set will be constructed among all anomaly types.
  }
  \label{tab:type}
  \centering
  \scalebox{1}{
  \begin{tabular}{llllll|l}
    \hline
    \multirow{2}{*}{Methods} & AH & ST& LA&TC&OC&\multirow{2}{*}{All Types}\\
    &VP&OO-LEFT&OO-RIGHT&VO&UK\\
    \hline
    \multirow{2}{*}{Deep SVDD~\cite{pmlr-v80-ruff18a}} &0.49 & 0.55& 0.49& 0.53 & 0.57 &\multirow{2}{*}{0.52}\\
    &0.50 &0.55& 0.52  & 0.56& 0.53 \\
    \hline
     \multirow{2}{*}{Liu \etal~\cite{liu2020feature}}& 0.66$\pm$0.09 & 0.55$\pm$0.09& 0.61$\pm$0.10& 0.67$\pm$0.09 & 0.66$\pm$0.10 &\multirow{2}{*}{0.66$\pm$0.10}\\
    &0.65$\pm$0.09 &0.75$\pm$0.09& 0.73$\pm$0.08  & 0.64$\pm$0.10& 0.59$\pm$0.09\\
    \hline
     \multirow{2}{*}{Xian \etal~\cite{xian2020generalized}} & 0.66$\pm$0.10 & 0.53$\pm$0.08& 0.61$\pm$0.10& 0.65$\pm$0.10 & 0.63$\pm$0.10 &\multirow{2}{*}{0.65$\pm$0.09}\\
    &0.65$\pm$0.09 &0.72$\pm$0.09& 0.70$\pm$0.10  & 0.60$\pm$0.09& 0.59$\pm$0.10& \\
    \hline
    
    \multirow{2}{*}{Ours} & \textbf{0.84$\pm$0.07} & \textbf{0.75$\pm$0.09}& \textbf{0.80$\pm$0.08}& \textbf{0.82$\pm$0.08} & \textbf{0.83$\pm$0.07} &\multirow{2}{*}{\textbf{0.81$\pm$0.08}}\\
    &\textbf{0.79$\pm$0.07} &\textbf{0.84$\pm$0.07}& \textbf{0.80$\pm$0.09}  & \textbf{0.82$\pm$0.09}& \textbf{0.69$\pm$0.10} \\
    \hline
  \end{tabular}
  }
  \vspace{-5mm}
\end{table*}
\subsection{Training of STGCN} 
With the updated node features $Z'$, the video feature $f^v=\max(Z')$ will be computed as the maximum  of all node features.
The feature will be classified by a classification head $c_t$ to get the predicted class, as shown in the second row of Fig. \ref{fig:pipeline}. 
Based on the 2-way $K$-shot dataset $D_t$, we use the cross-entropy loss to train $W$ in STGCN and the classification head $c_t$. 


\begin{table}
  \caption{2-way 5-shot result of different anomaly types in UCF-Crime. A full table with all different types is in Appendix~B.
  }
  \label{tab:type_ucf}
  \centering
  \resizebox{\linewidth}{!}{
  \begin{tabular}{llll|l}
    \hline
    \multirow{2}{*}{Methods} & Arson & Burglary& Explosion&\multirow{2}{*}{All Types}\\
    &Road Accidents&Shooting&Shoplifting\\
    \hline
    \multirow{2}{*}{Deep SVDD~\cite{pmlr-v80-ruff18a}} &0.59 & 0.55& 0.59 &\multirow{2}{*}{0.54}\\
    &0.57 &0.52& 0.63\\
    \hline
    \multirow{2}{*}{Liu \etal~\cite{liu2020feature}}&0.70$\pm$0.09 & 0.59$\pm$0.10& 0.65$\pm$0.10 &\multirow{2}{*}{0.56$\pm$0.09}\\
    &0.61$\pm$0.11 &0.56$\pm$0.10& 0.65$\pm$0.10\\
    \hline
    \multirow{2}{*}{Xian \etal~\cite{xian2020generalized}} & 0.80$\pm$0.08 & 0.62$\pm$0.08& 0.73$\pm$0.09&\multirow{2}{*}{0.62$\pm$0.10}\\ &0.68$\pm$0.09 & 0.60$\pm$0.09 &0.70$\pm$0.09 \\
    \hline
    
    \multirow{2}{*}{Ours} & \textbf{0.86$\pm$0.07} & \textbf{0.69$\pm$0.09}& \textbf{0.78$\pm$0.09}&\multirow{2}{*}{\textbf{0.66$\pm$0.10}}\\ &\textbf{0.75$\pm$0.09} & \textbf{0.72$\pm$0.09} &\textbf{0.77$\pm$0.09} \\
    \hline
  \end{tabular}}  
\end{table}
\section{Experiment}
\vspace{-2mm}
\subsection{Experimental Setting}

\noindent\textbf{Datasets.} IG-65M is a large dataset collected from Instagram, which contains more than 65 million videos. Pretrained backbones are released instead of the dataset itself. Therefore,
We choose IG-65M~\cite{ghadiyaram2019large} as our source domain. 
To test our method in different task types with diverse scenes, we choose a 
traffic dataset (Detection of Traffic Anomaly (DoTA)~\cite{yao2020dota}) and a surveillance dataset (UCF-Crime~\cite{sultani2019realworld}) as our target domains:
    Detection of Traffic Anomaly (DoTA) is a traffic dataset containing 4,677 videos with temporal, spatial, and categorical annotations. Since the anomaly start and the end time are annotated, we capture the frames before the anomaly start time as the normal sample while capturing the frames between the anomaly start and end time as the abnormal sample.
    UCF-Crime consists of 1,900 long and untrimmed real-world surveillance videos, with 13 real anomalies such as fighting, road accident, burglary, robbery, and normal activities. In order to get enough samples for evaluation, we augmented UCF-Crime by annotating more temporal segmentation. 
    

\noindent\textbf{Networks and Baselines.} We use 34-layer R(2+1)D IG-65M pretrained backbone as our feature encoder on the source domain. The clip length is 8, and the frame size is $224\times224$.
For each method, training will be based on this pretrained backbone instead of training from scratch. 
We choose a one-class classification method Deep SVDD~\cite{pmlr-v80-ruff18a}, a few-show classification method~\cite{xian2020generalized} and a cross-domain few-shot classification method on image domain~\cite{liu2020feature}  as our baselines. Deep SVDD~\cite{pmlr-v80-ruff18a} leverage only normal videos  to learn a hypersphere in feature space built by $\phi_s$.
Liu \etal~\cite{liu2020feature} applied meta-training on the source domain with batch spectral regularization. On the other hand, Xian \etal~\cite{xian2020generalized}  assumes that a model pretrained in a large and general dataset can get adequate knowledge to achieve few-shot learning by training a linear classifier without updating the model's parameters in the test phase. We choose 3DFSV as an implementation of this method.


\noindent\textbf{Domain adaptation module settings.} For the self-supervised learning task, we set the batch size as 44, the size of the memory bank as 4200, the learning rate as 0.003, and train the network for 200 epochs. 
For DoTA and UCF-Crime, 4136 and 950 samples with lengths more than 20 frames are used for DAM correspondingly.

\noindent\textbf{Meta context perception module settings.} The STGCN is implemented following the structure GCNeXt as in Xu \etal~\cite{xu2020g}. The video clip length is 8 frames, and the stride size is 4 frames. The minimum video length is set as 20 frames, 
and the maximum length is 124 frames.
The graph output dimension is 32 with 4 paths. The graph is trained with a learning rate of 0.001 and a batch size of 4 for 60 epochs.

\begin{table}
  \caption{Evaluation on vary number of shots}
  \label{tab:shots}
  \centering
  \resizebox{\linewidth}{!}{
  \begin{tabular}{lllllllll}
    \hline
     DAM& \tabincell{l}{MCPM} & 5 shot & 10 shot &15 shot &20 shot\\
    \hline
      $\checkmark$ & \xmark & 0.76$\pm$0.09 & 0.78$\pm$0.08 &0.79$\pm$0.07&0.81$\pm$0.07\\
    $\checkmark$ & $\checkmark$& 0.81$\pm$0.08&0.83$\pm$0.07&0.84$\pm$0.07&0.86$\pm$0.07\\
    
    \hline
  \end{tabular}}
    \vspace{-5mm}
\end{table}
\noindent\textbf{Evaluation settings.} 
We will employ a 2-way-5-shot-15-query meta-testing on the target domain based on the support set $S_t$ and the query set $Q_t$, which means there are $2\times 5$ samples in the support set for training and $2 \times 15$ samples in the query set for evaluation in each iteration of testing. Since the performance is sensitive to the sampling of $S_t$ and $Q_t$, we report the average accuracy and standard deviation among 200 iterations of testing with randomly sampled $S_t$ and $Q_T$. The performance on the meta-test set sampled from all different anomaly types will be the primary metric for evaluation.
In addition, to explore the performance for a specific anomaly type, 
we build additional meta-test sets sampled from each anomaly type.
We used the same abbreviation of anomaly types as \cite{yao2020dota} for DoTA. 
Note that for Deep SVDD, we did not employ the meta-testing, so the result is among the entire test set; therefore, the standard deviation will not be reported here.
\subsection{Evaluation Result} \label{sec:eval_res}


The evaluation result on DoTA and UCF-Crime is shown in Table \ref{tab:type} and Table \ref{tab:type_ucf}. Among all of the results, 
Deep SVDD\cite{pmlr-v80-ruff18a} only leads to the lowest performance, indicating that simply leveraging normal videos as semi-supervised training is inadequate. 
Liu \etal~\cite{liu2020feature} allows the meta-training in the source domain, which does not show a significant advantage over the supervised classification pretrained on source domain as Xian \etal~\cite{xian2020generalized} on DoTA and get even worse performance on UCF-Crime.
Therefore, our method simply uses the classification pretrained on the source domain and adapts to the normal samples, leading to much better performance. 
In our method, we gain general knowledge from the source domain, and adapt the learned knowledge by the normal samples, and fit to the target domain by leveraging the abnormal samples. 
As a result, anomaly detection is implemented in a cross-domain fashion.

Compared with DoTA, the result of UCF-Crime is less effective. Here are some possible explanations. At the data level, the number of normal samples for the DAM in UCF-Crime is less, but the scenes are more diverse than DoTA, which influenced the efficiency of the DAM. Besides,
in UCF-Crime, 
the anomaly sometimes happens in a small scene region, which will not cause enough global temporal and spatial information disturbance. It will be hard to detect, which might be a limitation of our methods even if it still outperforms the comparison methods.

\subsection{Ablation Study}




\subsubsection{Impact of Different Shots}

The ability of growth with more shots provided is an important metric to evaluate a method for few-shot learning. Therefore, we test the performance of our method in the different stages with a different number of shots provided in the support set. The result is shown in Fig. \ref{tab:shots}. With more shots provided, the performance of our method will increase as expected. 

\subsubsection{Impact of Different Modules}
\label{sec:exp_module_effect}

In order to discover the impact of each module, we test the performance of all anomaly types without specific modules on DoTA. 
A brief result is shown in Table \ref{tab:impact}, and the full results of the ablation study are shown in Appendix~C. 

\noindent\textbf{Pretrain on Source Domain.} Without the pretraining on the source domain, the performance will extremely drop. One possible explanation is that the DAM will be lack corresponding general knowledge learned from the source domain; therefore, even two modules are applied, the performance will drop, which emphasizes the feasibility of cross-domain task.

\noindent\textbf{Domain Adaptation Module.} The accuracy is improved after adaptation. It shows that our DA module managed to transfer the adapted domain knowledge from the source domain to the target domain. 


\noindent\textbf{Meta Context Perception Module.} The performance with only MCPM for all anomaly types is not improved compared with the pretrained model, but the performance for specific anomaly types is improved. (The results are included in Appendix~C.1).  However, if the representation for modeling the meta context is adapted, the learned meta context will benefit the anomaly detection thus get better performance compared with DAM only model. Meanwhile, compared with the video feature computed directly by the maximum of all video clip features, MCPM also gets the better performance, as shown in Appendix C.2, which proves that MCPM can extract better video features under the few-shot setting.



\begin{table}
  \caption{Evaluation on impacts of each module}
  \label{tab:impact}
  \centering
  \begin{tabular}{llllll}
    \hline
    \tabincell{l}{Source\\Domain} & \tabincell{l}{Domain\\Adaptation}& \tabincell{l}{Meta Context\\Perception} & Accuracy\\
    \hline
    \cmark &\xmark &\xmark& 0.65$\pm$0.09\\
    \hline
      \xmark & $\checkmark$ & $\checkmark$& 0.54$\pm$0.11\\
      $\checkmark$ & $\checkmark$ & \xmark & 0.76$\pm$0.09\\
      $\checkmark$ & \xmark & $\checkmark$& 0.65$\pm$0.10\\
      $\checkmark$ & $\checkmark$ & $\checkmark$& \textbf{0.81$\pm$0.08}\\
    
    \hline
  \end{tabular}
\end{table}

\section{Discussion}
\begin{figure}[t]
  \centering
  \includegraphics[width=\linewidth]{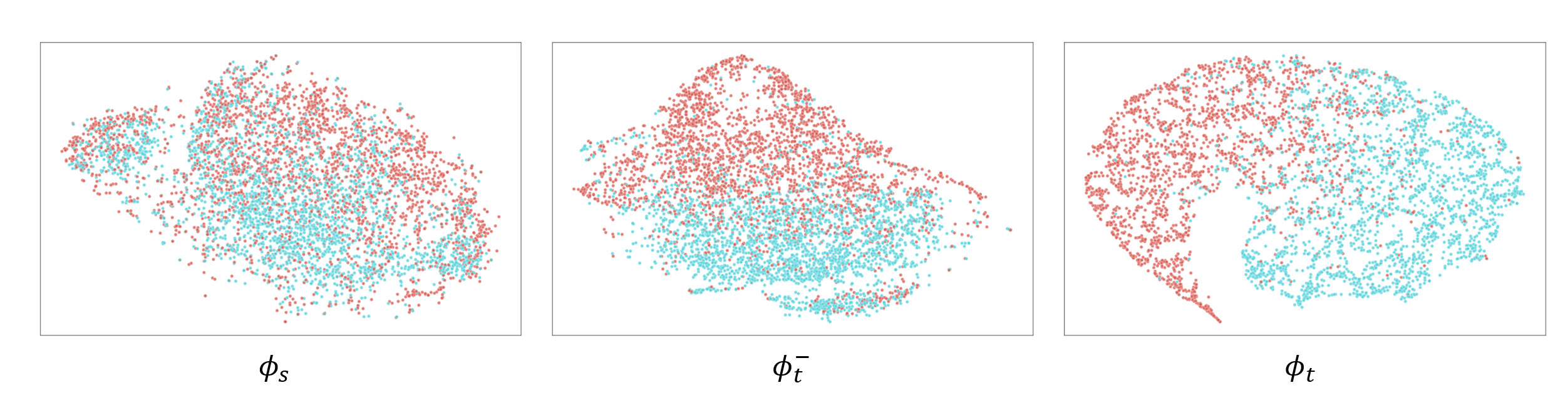}
  \caption{t-SNE plots for features in each stage of Anomaly Crossing under a 5-shot setting on DoTA. Blue and red points refer to the normal and abnormal samples correspondingly.
  }
  \label{fig:tsne}
  \vspace{-4mm}
\end{figure}

\subsection{Why do the modules work?}
The classification accuracy is highly dependent on the grouping of features on the target domain. We hypothesize that our modules can adapt the model into a better representation, i.e., the grouping of features is more separable. To verify that our modules give a better representation of the target domain, we plot the t-SNE~\cite{van2008visualizing} of test data to visualize the features exacted by encoders in each stage. The plot is shown in Fig. \ref{fig:tsne} for the result among all anomaly types in DoTA. The remaining result for each specific anomaly type and the results on UCF-Crime is shown in Appendix D. The features extracted by $\phi_s$ are less separable, which verifies that simply supervised training cannot give a good representation when there is a domain gap. After DAM, the encoder $\phi_t^-$ gains the knowledge from the normal samples, leading to more separable features. However, $\phi_t^-$ does not gain knowledge directly from the abnormal samples. In MCPM, the video encoder $\phi_t$ leverages as few as 5 shots and temporal information on the target domain to further get more separable features, verifying our hypothesis. 

\subsection{Extreme situations for DAM}
In reality, some more extreme situations exist that there will not even be sufficient normal samples (\emph{e.g.}, Medical). A solution is to use a similar dataset without annotations to substitute the normal samples for DAM. We want to evaluate the robustness of our method under such extreme situations, so we test the performance using a different but similar dataset BDD100K~\cite{yu2020bdd100k}, which is a driving video dataset with 100K videos, to substitute the normal samples on DoTA. We will use 10000 samples in BDD100K for DAM.  
The result is shown in Table~\ref{tab:bdd}.
The performance of DAM only model on BDD100K can still yield a pleasing outcome. However, the performance after adding MCPM is worse than directly using the normal samples on the target domain. One possible explanation is that the video context will be more brutal to perceive with the substitute adaption to gain the same performance for video clips, but it will be more complicated to learn the meta context based on domain knowledge learned from a similar dataset.

\subsection{Potential Negative Impacts and Limitation}

\noindent\textbf{Potential Negative Societal Impacts.} If this work is applied to real scenarios, being too confident about the result might cause unpredictable losses (\eg, a malfunction of a crucial machine were not correctly detected). To mitigate such negative impacts, please consider the result rationally and take precautions against inaccurate results.

\noindent\textbf{Limitations.} The performance of our method is sensitive to the sampling of the few shots on the target domain. We simply perform a multi-time sampling to mitigate the impact of different sampling, which will slightly differ between the statistical and the real performance.

\begin{table}
  \caption{Evaluation on an extreme situation that normal samples are insufficient}
  \label{tab:bdd}
  \centering
  \begin{tabular}{llll}
    \hline
    \tabincell{l}{DAM\\Dataset} & \tabincell{l}{Domain\\Adaptation}& \tabincell{l}{Meta Context\\Perception} & Accuracy\\
    \hline
    BDD100K & $\checkmark$ & \xmark & 0.75$\pm$0.04\\
    BDD100K & $\checkmark$ & $\checkmark$& 0.77$\pm$0.09\\
    \hline
      DoTA& $\checkmark$ & \xmark & 0.76$\pm$0.09\\
      DoTA& $\checkmark$ & $\checkmark$& 0.81$\pm$0.08\\
    
    \hline
  \end{tabular}
\end{table}

\section{Conclusion}
We propose the Cross-domain Few-shot Anomaly Detection task to address real-world problems. We devise a new pipeline -- Anomaly Crossing -- to handle video anomaly detection and prove the effectiveness of self-supervised learning and context perception in this task. Anomaly Crossing outperforms existing methods significantly in DoTA and UCF-Crime datasets under two different settings.
This task proves that the knowledge from different domains and tasks, e.g., action recognition, can be transferred to the current task, which builds a bridge from different tasks and enhances the significance of video representation learning by expanding the down-streaming tasks.

{\small
\bibliographystyle{ieee_fullname}
\bibliography{egbib}
}

\vfill

\clearpage

\appendix

\section{Implement Details}
This section provides the implementation details for our experiments, including baselines, Domain Adaptation Module, and Meta Context Perception Module, as a supplementary for Section~4.1. 
\textbf{Our codes are released at \url{https://github.com/likeyhnbm/AnomalyCrossing}.}

\subsection{Experiment Environment}
The experiments are performed on a Linux operation system with 4$\times$Tesla P100 GPUs. If no parameter-updating of the backbone is needed, we will only use one GPU for most cases.
\subsection{Datasets}
\begin{itemize}
    \item DoTA~\cite{yao2020dota}: We capture the frames before the anomaly start time as the normal video and the frames between the anomaly start time and end time as the abnormal video for each video in DoTA. The temporal annotation is gained from the official documents of the dataset. We only select 10 ego-involved anomaly types in DoTA during the evaluation, considering the performance of non-ego types is not as significant as ego-involved ones, and we regard them as our future work. 
    
    \item UCF-Crime~\cite{sultani2019realworld}: We use the videos in ``normal'' category provided in the dataset as the normal videos and capture the abnormal videos based on the temporal annotations in the official documents of the dataset. Considering that the annotated videos in UCF-Crime are not sufficient for evaluation, we annotated more videos to get a more stable evaluation result based on a larger test set.
\end{itemize}

\subsection{Sampling Details}
When extracting frames from a video, the frame sample rate is set as 10. A video clip will contain $L$ frames, where $L$ is set according to the architecture of the video clip encoder.
In our default setting, we will sample a video clip to represent the video where it is sampled. 
Considering the fairness of evaluation when constructing the meta-testing set, for each video in the support set, we will randomly sample a video clip to adapt the model, but for each video in the query set, we will sample the middlemost video clip to give the result.

\begin{table*}[b]
  \caption{\textbf{2-way 5-shot result of different backbones.} The number in the brackets refers to the video clip length.}
  \label{tab:backbones}
  \centering
  \scalebox{0.9}{
  \begin{tabular}{llllll|l}
    \toprule
    \multirow{2}{*}{Backbones} & AH & ST& LA&TC&OC&\multirow{2}{*}{All Types}\\
    &VP&OO-LEFT&OO-RIGHT&VO&UK\\
    
    \midrule
    \midrule
    \multirow{2}{*}{IG65M(8)~\cite{ghadiyaram2019large}}&\res{0.66}{0.10}& \res{0.53}{0.08} &\textbf{\res{0.61}{0.10}} & \textbf{\res{0.65}{0.10}}&\textbf{\res{0.63}{0.10}}
     &\multirow{2}{*}{\textbf{\res{0.65}{0.09}}}\\
    &\textbf{\res{0.65}{0.09}} &\textbf{\res{0.72}{0.09}} & \textbf{\res{0.70}{0.10}} &\res{0.60}{0.09} & \textbf{\res{0.59}{0.10}}\\
     
    \midrule
    \multirow{2}{*}{IG65M(16)~\cite{ghadiyaram2019large}}& \textbf{\res{0.66}{0.09}}& \textbf{\res{0.57}{0.08}} &\res{0.60}{0.09} & \res{0.63}{0.09}&\res{0.61}{0.09}
     &\multirow{2}{*}{\res{0.60}{0.09}}\\
    &\res{0.60}{0.10} &\res{0.71}{0.10} & \res{0.69}{0.10} &\res{0.59}{0.09} & \res{0.56}{0.09}\\
    
    \midrule
     \multirow{2}{*}{IG65M(32)~\cite{ghadiyaram2019large}}& \res{0.63}{0.10}& \res{0.57}{0.09} &\res{0.59}{0.09} & \res{0.61}{0.10}&\res{0.60}{0.10}
     &\multirow{2}{*}{\res{0.61}{0.09}}\\
    &\res{0.55}{0.10} &\res{0.72}{0.10} & \res{0.68}{0.10} &\res{0.57}{0.09} & \res{0.54}{0.09}\\
    \midrule
    \midrule
    \multirow{2}{*}{Kinetics(8)~\cite{kay2017kinetics}}& \res{0.59}{0.10}& \res{0.51}{0.08} &\res{0.57}{0.10} & \res{0.62}{0.10}&\res{0.61}{0.09}
     &\multirow{2}{*}{\res{0.61}{0.10}}\\
    &\res{0.67}{0.08} &\res{0.68}{0.09} & \res{0.68}{0.10} &\textbf{\res{0.63}{0.08}} & \res{0.57}{0.09}\\
    \midrule

    \multirow{2}{*}{Kinetics(16)~\cite{kay2017kinetics}} &\res{0.58}{0.10}& \res{0.54}{0.09} &\res{0.56}{0.08} & \res{0.59}{0.10}&\res{0.57}{0.10}
     &\multirow{2}{*}{\res{0.59}{0.10}}\\
    &\res{0.59}{0.10} &\res{0.68}{0.11} & \res{0.65}{0.11} &\res{0.59}{0.10} & \res{0.56}{0.09}\\
    \midrule

     \multirow{2}{*}{Kinetics(32)~\cite{kay2017kinetics}} & \res{0.61}{0.10}& \res{0.54}{0.09} &\res{0.57}{0.09} & \res{0.59}{0.10}&\res{0.58}{0.09}
     &\multirow{2}{*}{\res{0.58}{0.10}}\\
    &\res{0.54}{0.09} &\res{0.54}{0.13} & \res{0.65}{0.11} &\res{0.59}{0.10} & \res{0.57}{0.09}\\
    \midrule
    \midrule
    \multirow{2}{*}{XDC(8)~\cite{alwassel2019self}} & \res{0.59}{0.10}& \res{0.52}{0.09} &\res{0.55}{0.10} & \res{0.56}{0.09}&\res{0.54}{0.09}
     &\multirow{2}{*}{\res{0.56}{0.09}}\\
    &\res{0.53}{0.09} &\res{0.61}{0.10} & \res{0.60}{0.10} &\res{0.52}{0.09} & \res{0.53}{0.09}\\
    
    \midrule

    \multirow{2}{*}{XDC(16)~\cite{alwassel2019self}} & \res{0.64}{0.11}& \res{0.53}{0.09} &\res{0.60}{0.10} & \res{0.63}{0.10}&\res{0.61}{0.11}
     &\multirow{2}{*}{\res{0.62}{0.11}}\\
    &\res{0.59}{0.09} &\res{0.69}{0.09} & \res{0.65}{0.10} &\res{0.54}{0.10} & \res{0.57}{0.09}\\
    
    \midrule
    \multirow{2}{*}{XDC(32)~\cite{alwassel2019self}} & \res{0.65}{0.10}& \res{0.57}{0.09} &\textbf{\res{0.61}{0.10}} & \res{0.62}{0.10}&\res{0.61}{0.10}
     &\multirow{2}{*}{\res{0.60}{0.10}}\\
    &\res{0.61}{0.10} &\res{0.68}{0.10} & \res{0.63}{0.10} &\res{0.61}{0.09} & \res{0.54}{0.09}\\
    
    \bottomrule
  \end{tabular}}
 
\end{table*}

\subsection{Selection of Backbone}
For the supervised training on the source domain, we evaluate the performance of different backbones with different video clip lengths on the target domain to select the best video clip encoder $\phi_s$. 
We build a model with each backbone followed with a classification head and apply a meta-testing on the model to evaluate the performance~\footnote{Codes are adapted from\\~\href{https://github.com/artest08/LateTemporalModeling3DCNN}{https://github.com/artest08/LateTemporalModeling3DCNN}}, \emph{i.e.}, freeze the backbone, train the classification head on the support set and test the result on the query set. 
We choose cosine distance~\cite{mangla2020charting} as the classification head, and the frame size is set as 224$\times$224. 
The result of 2-way 5-shot 15-query for 200 runs is shown in Table~\ref{tab:backbones}. The detailed architecture of each backbone is shown as follows:
\begin{itemize}
    \item IG65M~\cite{ghadiyaram2019large}: 34-layer R(2+1)D encoder, pretrained on IG-65M.
    \item Kinetics~\cite{kay2017kinetics}: 34-layer R(2+1)D encoder, pretrained on Kinetics.
    \item XDC~\cite{alwassel2019self}: 18-layer R(2+1)D encoder, XDC pretrained on IG-Kinetics~\cite{ghadiyaram2019large,kay2017kinetics}.
\end{itemize}
Therefore, we choose IG65M with a video clip length of 8 as our basic video clip encoder $\phi_s$ for further adaptation in each method. 

\subsection{Baselines}
    \subsubsection{Deep SVDD~\cite{pmlr-v80-ruff18a}}
    We implement Deep SVDD by adapting a PyTorch implementation~\footnote{\href{https://github.com/lukasruff/Deep-SVDD-PyTorch}{https://github.com/lukasruff/Deep-SVDD-PyTorch}} to our datasets. For the training, we will sample the middlemost video clip of each normal video. Then, we will feed these video clips as input and $\phi_s$ as the network to train the Deep SVDD. Then, we will fine-tune the hyperparameter $r$ to get the maximum accuracy on the validation set. We choose a different setting for the validation set based on the amount of annotated samples. 
    For DoTA, we use a copy of the train set as the validation set.
    For UCF-Crime, we split half of the test set as the validation set. Then, normal and abnormal samples are kept as the same number for the testing to maintain the data balance.
    
    This method is not in a few-shot setting, so the evaluation of this method will be slightly different from others. However, the metric we use for evaluation is the classification accuracy on the unseen data, which is consistent with the few-shot setting. Therefore, we believe the evaluation of Deep SVDD can be comparable with others.
    
    \subsubsection{Liu \etal~\cite{liu2020feature}}
    We implement Liu \etal's method based on its official implementation~\footnote{\href{https://github.com/liubingyuu/FTEM_BSR_CDFSL}{https://github.com/liubingyuu/FTEM\_BSR\_CDFSL}}. We apply the meta-training with Batch Spectral Regularization (BSR) to the IG-65M pretrained encoder $\phi_s$ on Kinetics 100~\cite{kay2017kinetics} dataset. We tried support vector machine, fully-connected layer, and cosine distance head as the classification head~\cite{mangla2020charting}. The overall performance of these three heads are \res{0.65}{0.10},\res{0.66}{0.10}, and \res{0.63}{0.09}. Therefore, we choose a fully-connected layer as the classification head on the target domain.
    
    \subsubsection{Xian \etal~\cite{xian2020generalized}}
    We implement Xian \etal's 3DFSV by using $\phi_s$ as the encoder learned from 
    the representation learning stage, and use cosine distance as the classification head on the few-shot learning stage. For the testing stage, we will sample the middlemost video clip feature as the video feature. We also test the performance of use maximum/mean as the feature, as the ``Pretrain+MAX/MEAN'' row shown in Table~\ref{tab:dota}, but there are only trivial changes on the performance.

\subsection{Domain Adaptation Module (DAM)}
We apply the Decoupling the Scene and the Motion (DSM)~\cite{wang2020enhancing} on the normal samples as the DAM. The implementation is adapted from the official implementation \footnote{\href{https://github.com/FingerRec/DSM-decoupling-scene-motion}{https://github.com/FingerRec/DSM-decoupling-scene-motion}}. The hyperparameters we use is mentioned in the Section~4.1 of the main paper.

\subsection{Meta Context Perception Module (MCPM)}
We construct the Semantic-Temporal Graph Convolution Network (STGCN) by a 1D convolution layer activated by ReLU followed by a GCNeXt\cite{xu2020g} Block. The implementation of GCNeXt is from the official implementation of Xu \etal \footnote{\href{https://github.com/frostinassiky/gtad}{https://github.com/frostinassiky/gtad}}. We will sample video clips with fixed lengths from the beginning of the video with a stride. The minimum and maximum frame lengths are set to ensure learning a meaningful context under memory limitation. If the video length is less than the minimum frame length, we will sample from the beginning again until we sample enough frames.   
The hyperparameters we use are mentioned in Section~4.1 of the main paper.

\section{Introduction for each anomaly type}
In this section, we introduce each anomaly type in Detection of Traffic Anomaly (DoTA)~\cite{yao2020dota} and UCF-Crime~\cite{sultani2019realworld} datasets as well as the abbreviation in Table.~1 of the main paper.

We construct $N+1$ different meta-testing sets to test the robustness of our method, where $N$ is the number of anomaly types. $N$ of the meta-testing sets contains only one specific anomaly type, annotated as the anomaly name, and one contains all 10 different types of anomaly, annotated as ``All Types''. Considering the difference between different anomaly types, ``All Types'' is a more complicated task than other meta-testing sets. The introduction of each anomaly type in DoTA is shown as follows:

\begin{itemize}
    \item AH: Moving ahead or waiting. Collision with another vehicle moving ahead or waiting, 592 videos in total.
    \item ST: Start-stop or stationary. Collision with another vehicle that starts stops or is stationary, 66 videos in total.
    \item LA: Lateral. Collision with another vehicle moving laterally in the same direction, 643 videos in total.
    \item TC: Turning. Collision with another vehicle that turns into or crosses a road, 1330 videos in total.
    \item OC: Oncoming. Collision with another oncoming vehicle, 528 videos in total.
    \item VP: Pedestrian. A collision between vehicle and pedestrian, 52 videos in total.
    \item OO-left, OO-right: Leaving to left and leaving to right. Out-of-control and leaving the roadway to the left or right, 266 videos in total for left, and 203 videos in total for right.
    \item VO: Collision with an obstacle in the roadway, 64 videos in total.
    \item UK: Unknown type, 56 videos in total.
    \item All Types: Samples containing all types, 3800 videos in total.
\end{itemize}

The introduction of each anomaly type in UCF-Crime is shown as follows:
\begin{itemize}
    \item Arson: Burning of property (such as a building), 61 videos in total.
    \item Burglary: Breaking and entering a dwelling, 142 videos in total.
    \item Explosion: Bursting forth with sudden violence or noise, 61 videos in total.
    \item Road Accidents: 164 videos in total.
    \item Shooting: Gun violence directed toward people, 71 videos in total.
    \item Shoplifting: Stealing displayed goods from a store, 71 videos in total.
    \item Abuse: 91 videos in total.
    \item Arrest: 66 videos in total.
    \item Assault: 63 videos in total.
    \item Fighting: 72 videos in total.
    \item Robbery: 152 videos in total.
    \item Stealing: 104 videos in total.
    \item Vandalism: 74 videos in total.
    \item All Types: Samples containing all types, 1342 videos in total.
\end{itemize}

The full result for UCF-Crime is shown in Table~\ref{tab:type_ucf_a}

\begin{table*}
  \caption{2-way 5-shot result of different anomaly types in UCF-Crime. 
  }
  \label{tab:type_ucf_a}
  \centering
  \resizebox{\linewidth}{!}{
  \begin{tabular}{llllllll|l}
    \toprule
    \multirow{2}{*}{Methods} & Arson & Burglary& Explosion&Abuse&Arrest&Assault&Fighting&\multirow{2}{*}{All Types}\\
    &Road Accidents&Shooting&Shoplifting&Robbery&Stealing&Vandalism&&\\
    \midrule
    \multirow{2}{*}{Deep SVDD~\cite{pmlr-v80-ruff18a}} &0.59 & 0.55& 0.59&0.54&0.65&0.59&0.47 &\multirow{2}{*}{0.55}\\
    &0.57 &0.52& 0.63&0.51&0.47&0.55&&\\
    \midrule
    \multirow{2}{*}{Liu \etal~\cite{liu2020feature}}&0.70$\pm$0.09 & 0.59$\pm$0.10& 0.65$\pm$0.10&0.66$\pm$0.10&0.59$\pm$0.10&0.66$\pm$0.10&0.66$\pm$0.10 &\multirow{2}{*}{0.57$\pm$0.09}\\
    &0.61$\pm$0.11 &0.56$\pm$0.10& 0.65$\pm$0.10&0.59$\pm$0.09&0.67$\pm$0.10&0.64$\pm$0.10&\\
    \midrule
    \multirow{2}{*}{Xian \etal~\cite{xian2020generalized}} & 0.80$\pm$0.08 & 0.62$\pm$0.08& 0.73$\pm$0.09&0.75$\pm$0.09&0.65$\pm$0.10&0.72$\pm$0.09&0.72$\pm$0.10 &\multirow{2}{*}{0.61$\pm$0.11}\\ &0.68$\pm$0.09 & 0.60$\pm$0.09 &0.70$\pm$0.09&0.60$\pm$0.10&0.71$\pm$0.10&\textbf{0.71$\pm$0.09}& \\
    \midrule
    
    \multirow{2}{*}{Ours} & \textbf{0.86$\pm$0.07} & \textbf{0.69$\pm$0.09}& \textbf{0.78$\pm$0.09}&\textbf{0.80$\pm$0.08}&\textbf{0.67$\pm$0.10}&\textbf{0.73$\pm$0.09}&\textbf{0.76$\pm$0.09}&\multirow{2}{*}{\textbf{0.67$\pm$0.09}}\\ &\textbf{0.75$\pm$0.09} & \textbf{0.72$\pm$0.09} &\textbf{0.77$\pm$0.09}&\textbf{0.70$\pm$0.10}&\textbf{0.76$\pm$0.08}&0.69$\pm$0.10 &\\
    \bottomrule
  \end{tabular}}  
\end{table*}

\section{Additional Ablation study}
This section describes the ablation studies in more detail by removing or replacing some components of our pipeline. We provide type-level results and more combinations to support the analysis in Section~4.3.2 of the main paper. 

The result is shown in Table~\ref{tab:dota} and Table~\ref{tab:ucf}. Pretrain refers to the supervised training on the source domain, the dataset name following the DAM indicates where the normal samples are from for DAM, and MAX/MEAN means directly using the maximum/mean of video clip features as the video feature to replace the MCPM.

\subsection{Impact of MCPM for different anomaly types}
In the Meta Context Perception Module part of Section 4.3.2 in the main paper, we propose that the performance of MCPM is dependent on the length of the event. The event length in UCF-Crime is longer than events in DoTA. Therefore, as a result, shown in Table~\ref{tab:ucf}, the impact of MCPM is larger compared with the performance on DoTA. Meanwhile, the performance for each anomaly type is improved, which proves that MCPM learns the meta context, which can be adapted to different anomaly types and domains under the few-shot setting.  
\begin{table*}
  \caption{2-way 5-shot result of different anomaly types in DoTA.}
  \label{tab:dota}
  \centering
  \resizebox{\linewidth}{!}{
  \begin{tabular}{llllll|l}
    \toprule
    \multirow{2}{*}{Modules} & AH & ST& LA&TC&OC&\multirow{2}{*}{All Types}\\
    &VP&OO-LEFT&OO-RIGHT&VO&UK\\
    
    \midrule
    \midrule
    \multirow{2}{*}{DAM(BDD100K)}& \res{0.60}{0.10}& \res{0.55}{0.09} &\res{0.57}{0.11} & \res{0.61}{0.10}&\res{0.61}{0.10}
     &\multirow{2}{*}{\res{0.60}{0.10}}\\
    &\res{0.51}{0.09} &\res{0.71}{0.09} & \res{0.68}{0.10} &\res{0.55}{0.10} & \res{0.53}{0.09}\\
     
    \midrule
    \multirow{2}{*}{DAM(DoTA)}& \res{0.53}{0.10}& \res{0.47}{0.08} &\res{0.52}{0.08} & \res{0.53}{0.10}&\res{0.53}{0.09}
     &\multirow{2}{*}{\res{0.53}{0.09}}\\
    &\res{0.49}{0.08} &\res{0.56}{0.09} & \res{0.58}{0.10} &\res{0.49}{0.08} & \res{0.48}{0.09}\\
    
    \midrule
     \multirow{2}{*}{DAM(BDD100K)+MCPM}& \res{0.61}{0.09}& \res{0.57}{0.09} &\res{0.56}{0.10} & \res{0.61}{0.10}&\res{0.63}{0.10}
     &\multirow{2}{*}{\res{0.61}{0.09}}\\
    &\res{0.52}{0.09} &\res{0.71}{0.09} & \res{0.70}{0.10} &\res{0.62}{0.09} & \res{0.49}{0.08}\\
    \midrule
    \multirow{2}{*}{DAM(DoTA)+MCPM}& \res{0.53}{0.09}& \res{0.49}{0.09} &\res{0.52}{0.10} & \res{0.54}{0.09}&\res{0.55}{0.09}
     &\multirow{2}{*}{\res{0.54}{0.11}}\\
    &\res{0.49}{0.08} &\res{0.59}{0.10} & \res{0.60}{0.10} &\res{0.53}{0.09} & \res{0.46}{0.07}\\
    \midrule
    \midrule
    \multirow{2}{*}{Pretrain} &\res{0.66}{0.10}& \res{0.53}{0.08} &\res{0.61}{0.10} & \res{0.65}{0.10}&\res{0.63}{0.10}
     &\multirow{2}{*}{\res{0.65}{0.09}}\\
    &\res{0.65}{0.09} &\res{0.72}{0.09} & \res{0.70}{0.10} &\res{0.60}{0.09} & \res{0.59}{0.10}\\
    \midrule
    \midrule
     \multirow{2}{*}{Pretrain+DAM(BDD100K)} & \res{0.80}{0.03}& \res{0.70}{0.06} &\res{0.73}{0.04} & \res{0.76}{0.03}&\res{0.75}{0.04}
     &\multirow{2}{*}{\res{0.75}{0.04}}\\
    &\res{0.72}{0.05} &\res{0.85}{0.01} & \res{0.83}{0.03} &\res{0.66}{0.06} & \res{0.70}{0.07}\\
    \midrule
    
    \multirow{2}{*}{Pretrain+DAM(DoTA)} & \res{0.81}{0.04}& \res{0.69}{0.06} &\res{0.73}{0.07} & \res{0.77}{0.05}&\res{0.75}{0.05}
     &\multirow{2}{*}{\res{0.75}{0.05}}\\
    &\res{0.73}{0.06} &\res{0.81}{0.04} & \res{0.78}{0.06} &\res{0.67}{0.05} & \res{0.66}{0.06}\\
    
    \midrule
    \midrule
    \multirow{2}{*}{Pretrain+MAX} & \res{0.67}{0.09}& \res{0.53}{0.08} &\res{0.62}{0.10} & \res{0.67}{0.10}&\res{0.66}{0.10}
     &\multirow{2}{*}{\res{0.64}{0.10}}\\
    &\res{0.64}{0.10} &\res{0.71}{0.10} & \res{0.68}{0.10} &\res{0.61}{0.10} & \res{0.58}{0.08}\\
    
    \midrule
    \multirow{2}{*}{Pretrain+MEAN} & \res{0.66}{0.09}& \res{0.53}{0.09} &\res{0.61}{0.10} & \res{0.67}{0.10}&\res{0.66}{0.10}
     &\multirow{2}{*}{\res{0.64}{0.10}}\\
    &\res{0.64}{0.10} &\res{0.71}{0.10} & \res{0.70}{0.10} &\res{0.62}{0.09} & \res{0.59}{0.10}\\
    \midrule
    
    \multirow{2}{*}{Pretrain+MCPM} & \res{0.67}{0.10}& \res{0.55}{0.09} &\res{0.62}{0.10} & \res{0.68}{0.10}&\res{0.67}{0.10}
     &\multirow{2}{*}{\res{0.65}{0.10}}\\
    &\res{0.71}{0.09} &\res{0.70}{0.10} & \res{0.70}{0.10} &\res{0.64}{0.11} & \res{0.58}{0.08}\\
    
    \midrule
    \multirow{2}{*}{Pretrain+DAM(DoTA)+MAX} & \res{0.82}{0.09}& \res{0.71}{0.09} &\res{0.77}{0.09} & \res{0.80}{0.09}&\res{0.79}{0.09}
     &\multirow{2}{*}{\res{0.78}{0.09}}\\
    &\res{0.72}{0.09} &\res{0.80}{0.08} & \res{0.76}{0.09} &\res{0.77}{0.10} & \res{0.66}{0.11}\\
    
    \midrule
    \multirow{2}{*}{Pretrain+DAM(DoTA)+MEAN} & \res{0.81}{0.09}& \res{0.70}{0.09} &\res{0.75}{0.10} & \res{0.78}{0.09}&\res{0.77}{0.09}
     &\multirow{2}{*}{\res{0.76}{0.10}}\\
    &\res{0.74}{0.10} &\res{0.81}{0.08} & \res{0.78}{0.08} &\res{0.74}{0.09} & \res{0.66}{0.10}\\

    \midrule
    \midrule
    \multirow{2}{*}{Pretrain+DAM(BDD100K)+MCPM} & \res{0.83}{0.08}& \res{0.71}{0.10} &\res{0.74}{0.10} & \res{0.79}{0.08}&\res{0.79}{0.07}
     &\multirow{2}{*}{\res{0.77}{0.09}}\\
    &\res{0.77}{0.09} &\textbf{\res{0.84}{0.07}} & \textbf{\res{0.83}{0.07}} &\res{0.73}{0.09} & \res{0.72}{0.10}\\
    \midrule
    \multirow{2}{*}{Pretrain+DAM(DoTA)+MCPM} & \textbf{0.84$\pm$0.07} & \textbf{0.75$\pm$0.09}& \textbf{0.80$\pm$0.08}& \textbf{0.82$\pm$0.08} & \textbf{0.83$\pm$0.07} &\multirow{2}{*}{\textbf{0.81$\pm$0.08}}\\
    &\textbf{0.79$\pm$0.07} &\textbf{0.84$\pm$0.07}& 0.80$\pm$0.09  & \textbf{0.82$\pm$0.09}& \textbf{0.69$\pm$0.10} \\
    \bottomrule
  \end{tabular}}
  \vspace{-5mm}
\end{table*}

\subsection{Comparison between MCPM and MAX/MEAN}
In order to prove that the STGCN updated video clip features are better than the original video clip features, we remove the STGCN in MCPM and extract the video feature by directly using the maximum/mean of video  clip features. As the result shown in Table~\ref{tab:dota} and Table~\ref{tab:ucf}, the performance of MCPM is better than 
MAX/MEAN in most types of anomaly without DAM and in each anomaly type with DAM, which proves the effectiveness of MCPM.

\begin{table*}
  \caption{2-way 5-shot result of different anomaly types in UCF-Crime. 
  }
  \label{tab:ucf}
  \centering
  \resizebox{\linewidth}{!}{
  \begin{tabular}{llllllll|l}
    \toprule
    \multirow{2}{*}{Modules} & Arson & Burglary& Explosion&Abuse&Arrest&Assault&Fighting&\multirow{2}{*}{All Types}\\
    &Road Accidents&Shooting&Shoplifting&Robbery&Stealing&Vandalism&&\\
    \midrule
    \multirow{2}{*}{Pretrain} &\res{0.80}{0.08}&\res{0.62}{0.09}&\res{0.73}{0.09} &\res{0.75}{0.09}&\res{0.65}{0.09}&\res{0.72}{0.09}&\res{0.72}{0.10}&\multirow{2}{*}{\res{0.62}{0.10}}\\
    &\res{0.68}{0.09}&\res{0.60}{0.09}&\res{0.70}{0.09}&\res{0.60}{0.10}&\res{0.71}{0.10}&\res{0.71}{0.09}&\\
    
    \midrule
    \multirow{2}{*}{Pretrain+MAX} &\res{0.83}{0.08}&\res{0.64}{0.11}&\res{0.76}{0.08} &\res{0.75}{0.10}&\res{0.64}{0.11}&\res{0.70}{0.12}&\res{0.75}{0.10}&\multirow{2}{*}{\res{0.66}{0.11}}\\
    &\res{0.67}{0.11}&\res{0.64}{0.11}&\res{0.73}{0.10}&\res{0.60}{0.11}&\res{0.76}{0.10}&\res{0.69}{0.10}&\\
    
    \midrule
    
    \multirow{2}{*}{Pretrain+MEAN} &\res{0.82}{0.08}&\res{0.64}{0.10}&\res{0.76}{0.09} &\res{0.73}{0.09}&\res{0.66}{0.10}&\res{0.69}{0.10}&\res{0.74}{0.08}&\multirow{2}{*}{\res{0.64}{0.10}}\\
    &\res{0.68}{0.10}&\res{0.63}{0.10}&\res{0.72}{0.09}&\res{0.60}{0.10}&\res{0.75}{0.08}&\res{0.68}{0.10}&\\

    \midrule
    \multirow{2}{*}{Pretrain+MCPM} &\res{0.82}{0.09}&\res{0.66}{0.10}&\res{0.76}{0.09} 
    &\res{0.76}{0.09}&\res{0.65}{0.09}&\res{0.71}{0.09}&\res{0.76}{0.09}&\multirow{2}{*}{\res{0.64}{0.10}}\\
    &\res{0.72}{0.09}&\res{0.65}{0.10}&\res{0.73}{0.09}&\res{0.61}{0.10}&\res{0.75}{0.09}&\res{0.69}{0.08}&\\
    
    \midrule
    
    \multirow{2}{*}{Pretrain+DAM}&\res{0.84}{0.07}&\res{0.64}{0.09}&\res{0.76}{0.08}&\res{0.75}{0.09}&\res{0.62}{0.10}&\res{0.70}{0.11}&\res{0.75}{0.09}
    &\multirow{2}{*}{\res{0.64}{0.10}}\\
    &\res{0.68}{0.09}&\res{0.64}{0.09}&\res{0.72}{0.08}&\res{0.64}{0.10}&\res{0.71}{0.10}&\res{0.67}{0.10}&\\
    \midrule
    
    \multirow{2}{*}{Pretrain+DAM+MAX} &\res{0.84}{0.08}&\res{0.69}{0.11}&\res{0.75}{0.09}&\res{0.78}{0.10}&\res{0.65}{0.11}&\res{0.72}{0.11}&\res{0.76}{0.10}
    &\multirow{2}{*}{\res{0.65}{0.12}}\\
    &\res{0.70}{0.10}&\res{0.70}{0.10}&\res{0.75}{0.09}&\res{0.68}{0.12}&\textbf{\res{0.78}{0.09}}&\textbf{\res{0.71}{0.12}}&\\
    \midrule
    
    \multirow{2}{*}{Pretrain+DAM+MEAN} &\res{0.86}{0.07}&\res{0.66}{0.10}&\res{0.77}{0.08} &\res{0.78}{0.09}&\res{0.65}{0.10}&\res{0.71}{0.09}&\textbf{\res{0.76}{0.09}}
    &\multirow{2}{*}{\res{0.66}{0.10}}\\
    &\res{0.71}{0.10}&\res{0.70}{0.10}&\res{0.76}{0.09}&\res{0.69}{0.10}&\res{0.77}{0.09}&\res{0.68}{0.11}&\\
    \midrule

    \multirow{2}{*}{Pretrain+DAM+MCPM} &\textbf{\res{0.86}{0.07}}&\textbf{\res{0.69}{0.09}}&\textbf{\res{0.78}{0.09} }&\textbf{\res{0.80}{0.08}}&\textbf{\res{0.67}{0.10}}&\textbf{\res{0.73}{0.09}}&\textbf{\res{0.76}{0.09}}
    &\multirow{2}{*}{\textbf{\res{0.67}{0.09}}}\\
    &\textbf{\res{0.75}{0.09}}&\textbf{\res{0.72}{0.09}}&\textbf{\res{0.77}{0.08}}&\textbf{\res{0.70}{0.10}}&\res{0.76}{0.08}&\res{0.69}{0.10}&\\
    \bottomrule
  \end{tabular}}
\end{table*}

\subsection{Impact of DAM for different anomaly types}
The goal of DAM is to obtain a better video clip encoder $\phi_t^-$. As the result shown in Table~\ref{tab:dota} and Table~\ref{tab:ucf},
the performance will be significantly better with DAM regardless of the setting of other modules. This result suggests the robustness of DAM as expected. Considering different sources of normal samples for DAM, we found that the effectiveness of DAM is related to the quality of normal samples. Such quality can be evaluated by the distance to the target domain, the resolution of the video, and the amount of videos. DoTA has no distance to the target domain, and 4136 videos are used for DAM, which leads to the best performance. BDD100K~\cite{yu2020bdd100k} is close to the target domain, and 10,000 high-resolution videos are used for DAM; therefore, the result can be comparable to the best performance. For UCF-Crime, most of the videos are captured by surveillance cameras, whose resolution is not high. Meanwhile, only 950 normal videos are used for DAM; therefore, the effectiveness of DAM is not as significant as one on DoTA, although it has no distance to the target domain.


\section{More t-SNE~\cite{van2008visualizing} plots in DoTA and UCF-Crime}
\begin{figure}[htbp]
\centering
\begin{subfigure}[t]{0.32\textwidth}
\centering
\includegraphics[width=\textwidth]{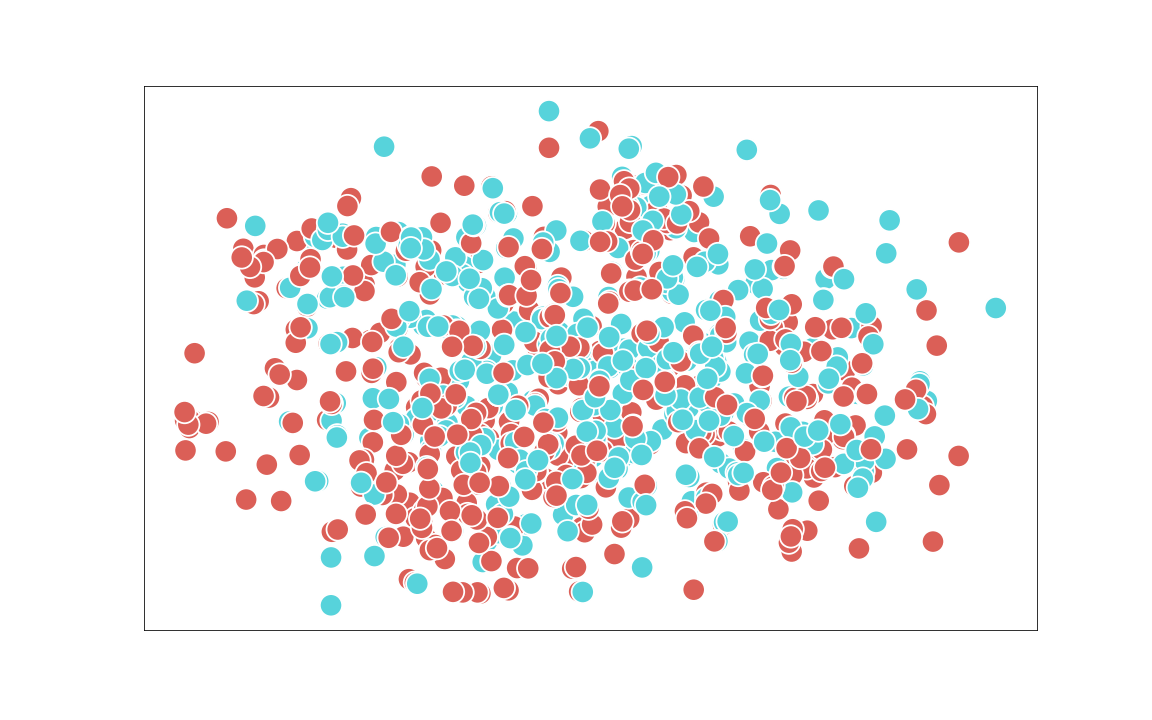}
\caption{$\phi_s$}
\end{subfigure}
\begin{subfigure}[t]{0.32\textwidth}
\centering
\includegraphics[width=\textwidth]{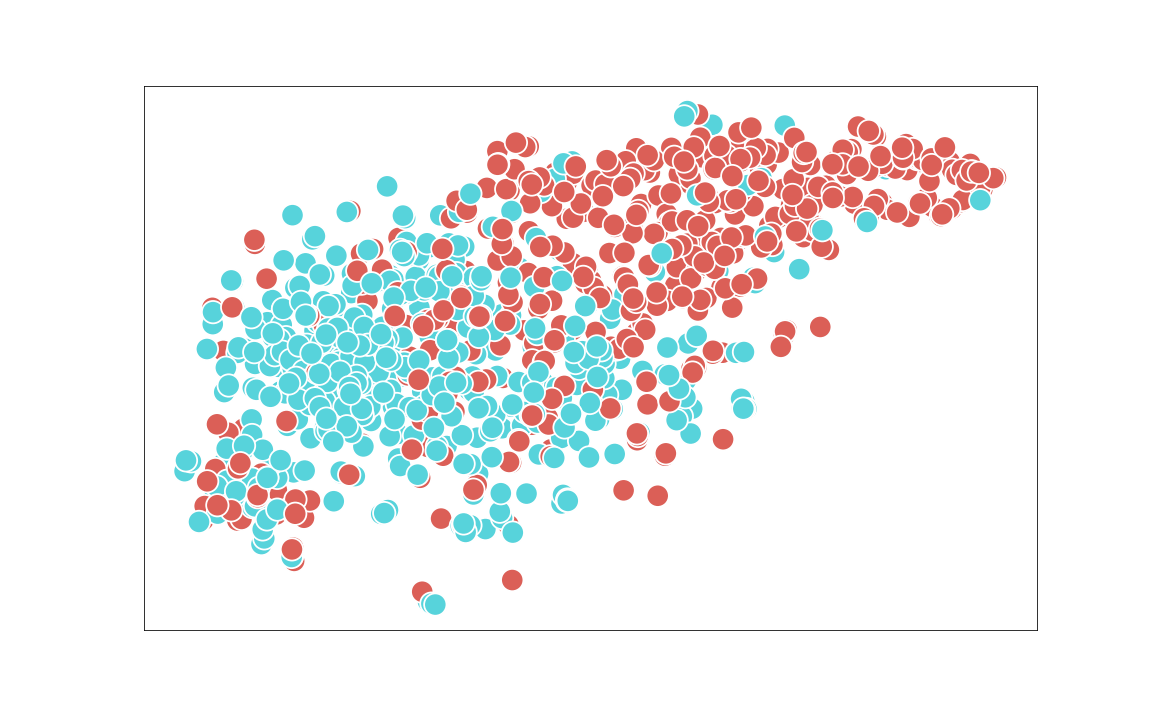}
\caption{$\phi_t^-$}
\end{subfigure}
\begin{subfigure}[t]{0.32\textwidth}
\centering
\includegraphics[width=\textwidth]{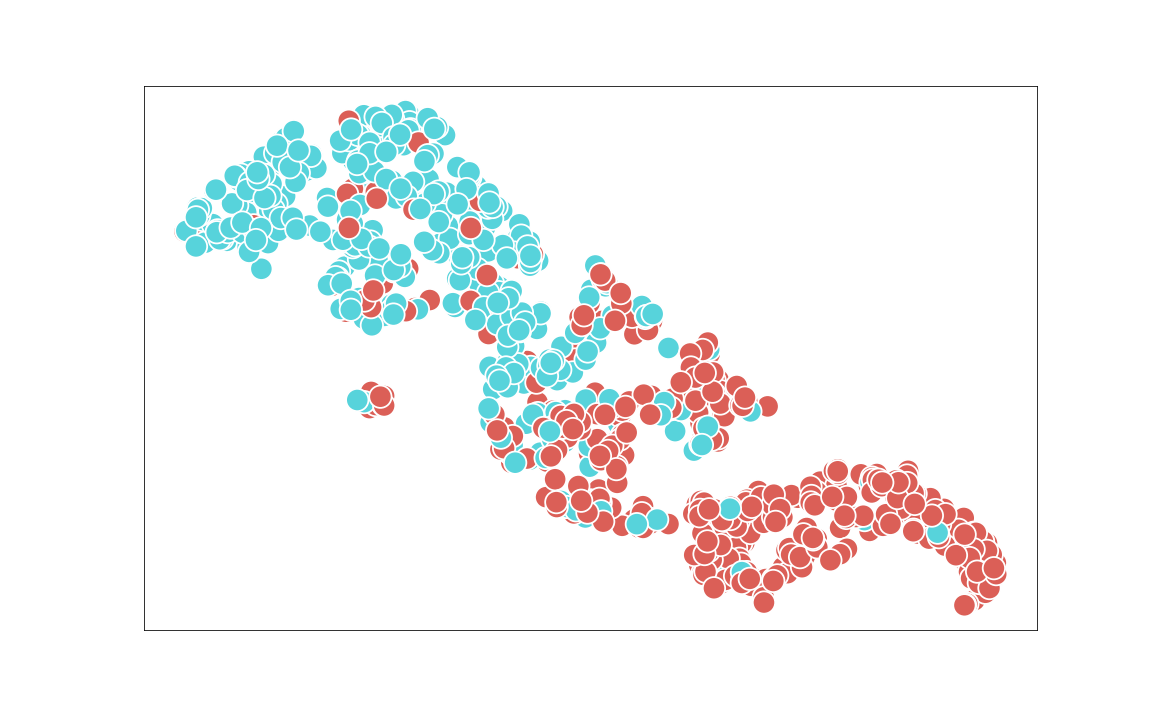}
\caption{$\phi_t$}
\end{subfigure}
\caption{t-SNE plots for ``Lateral'' in DoTA}
\label{fig:start}
\end{figure}

\begin{figure}[htbp]
\centering
\begin{subfigure}[t]{0.32\textwidth}
\centering
\includegraphics[width=\textwidth]{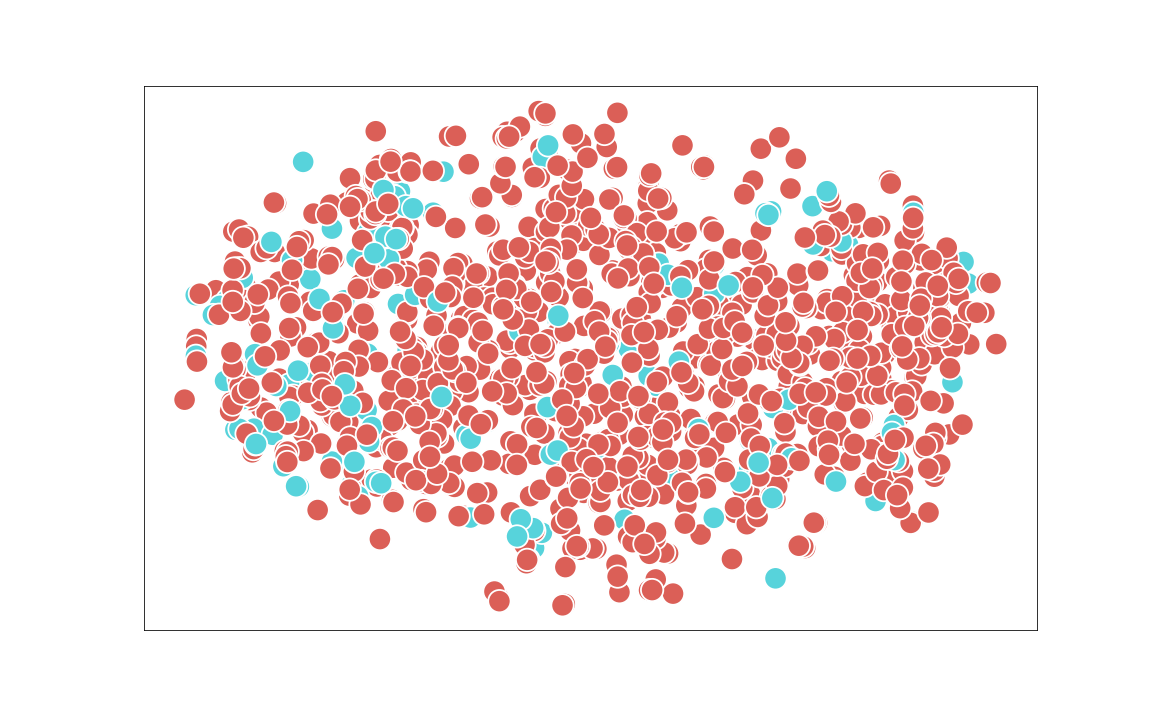}
\caption{$\phi_s$}
\end{subfigure}
\begin{subfigure}[t]{0.32\textwidth}
\centering
\includegraphics[width=\textwidth]{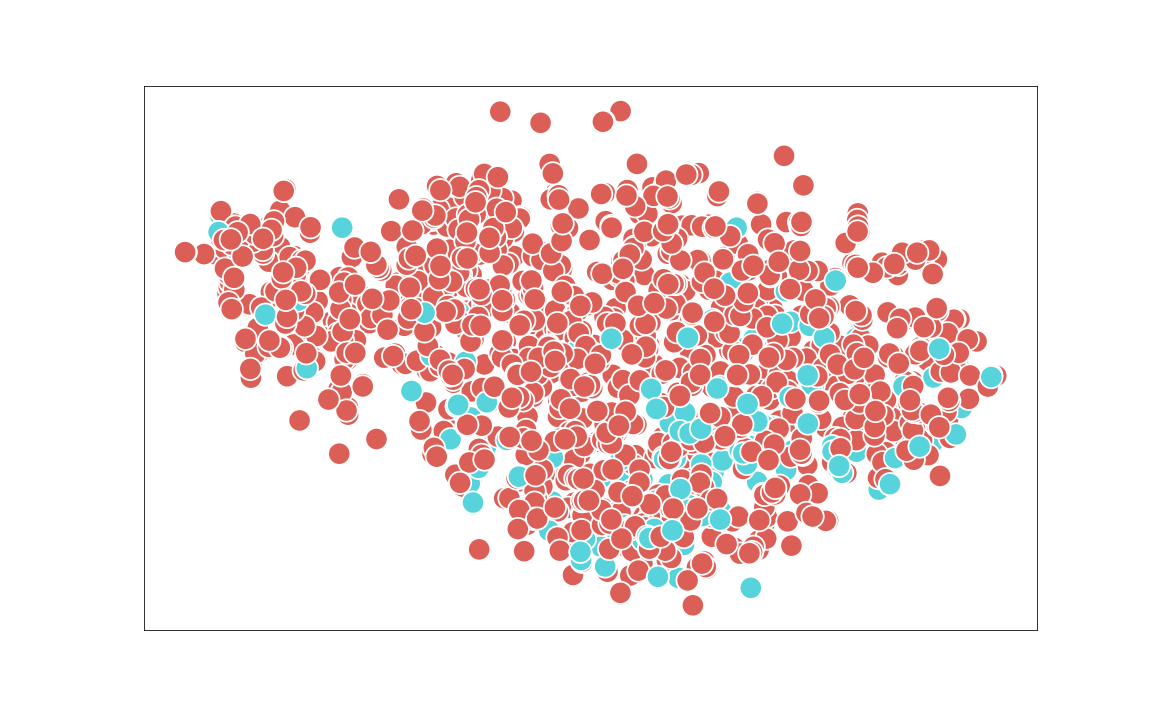}
\caption{$\phi_t^-$}
\end{subfigure}
\begin{subfigure}[t]{0.32\textwidth}
\centering
\includegraphics[width=\textwidth]{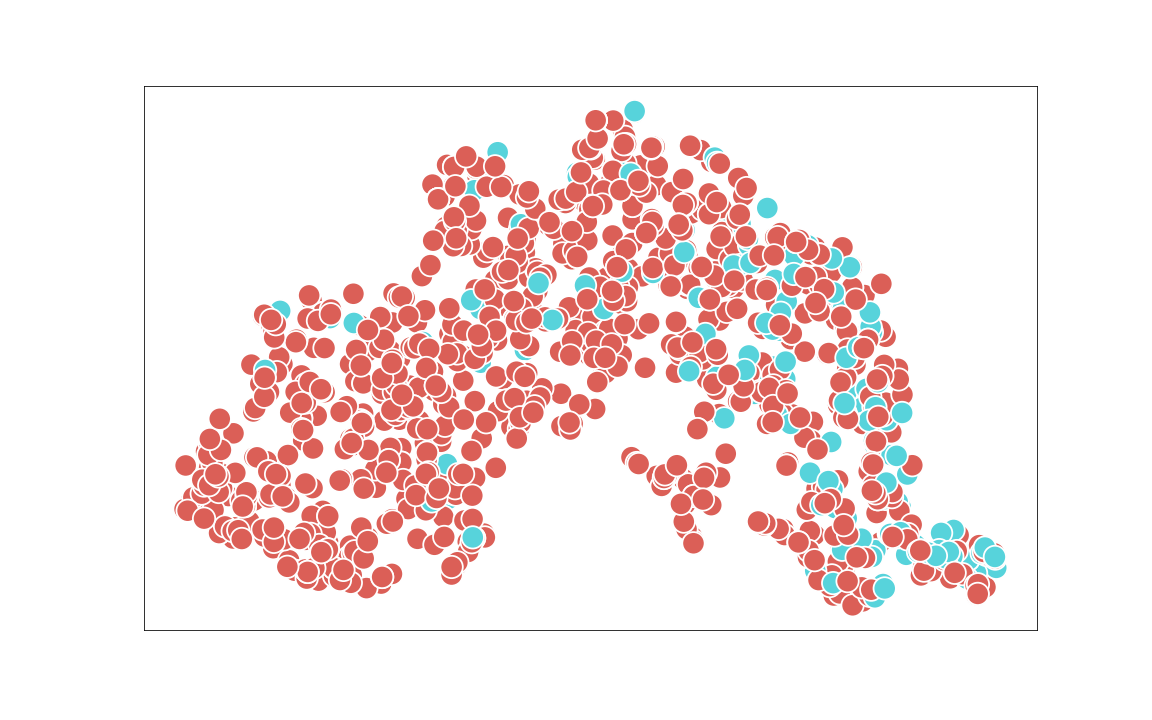}
\caption{$\phi_t$}
\end{subfigure}
\caption{t-SNE plots for ``All'' in UCF Crime}
\label{fig:end}
\end{figure}

In this section, we provide t-SNE plots for each anomaly type as well as the ``All types'' in DoTA and UCF-Crime to prove our modules improve the feature space to be more separable under various types of anomaly. The results are shown from Fig.~\ref{fig:start} to Fig.~\ref{fig:end}. Note that the t-SNE plot might be slightly different for different runs.
These plots show the evolution of features from $\phi_s$ to $\phi_t$, which proves the effectiveness and robustness of Anomaly Crossing. 

These plots support our analysis of ablation studies. The performance is related to the grouping of the features. For most anomaly types, with the help of DAM, the feature extracted by $\phi_t^-$ will be more separable compared with those extracted by $\phi_s$.
For some long-time events (\emph{i.e.}, events in UCF-Crime), the performance of DAM is not as significant as others, leading to less separable features. With the help of learning meta context from MCPM, the video feature extracted by $\phi_t$ will be more separable, proving that MCPM supplements DAM. As a result, $\phi_s$ will be adapted to $\phi_t$ by DAM and MCPM as a better representation of the target domain.

\begin{figure}[htbp]
\centering
\begin{subfigure}[t]{0.32\textwidth}
\centering
\includegraphics[width=\textwidth]{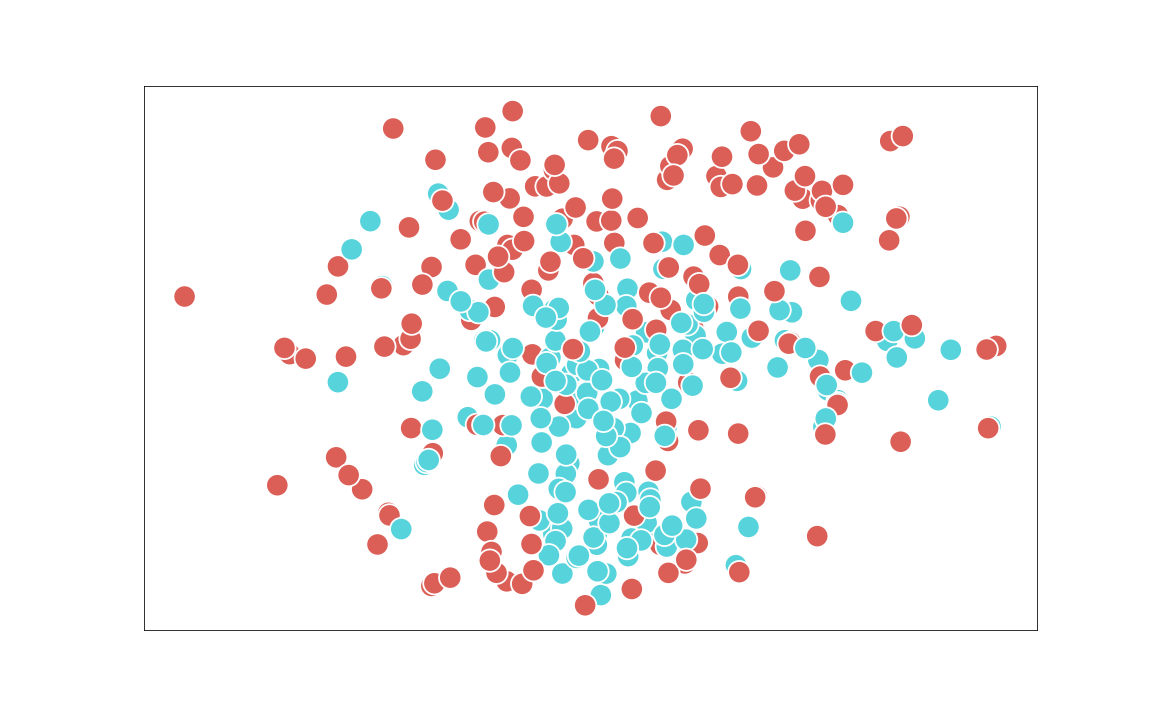}
\caption{$\phi_s$}
\end{subfigure}
\begin{subfigure}[t]{0.32\textwidth}
\centering
\includegraphics[width=\textwidth]{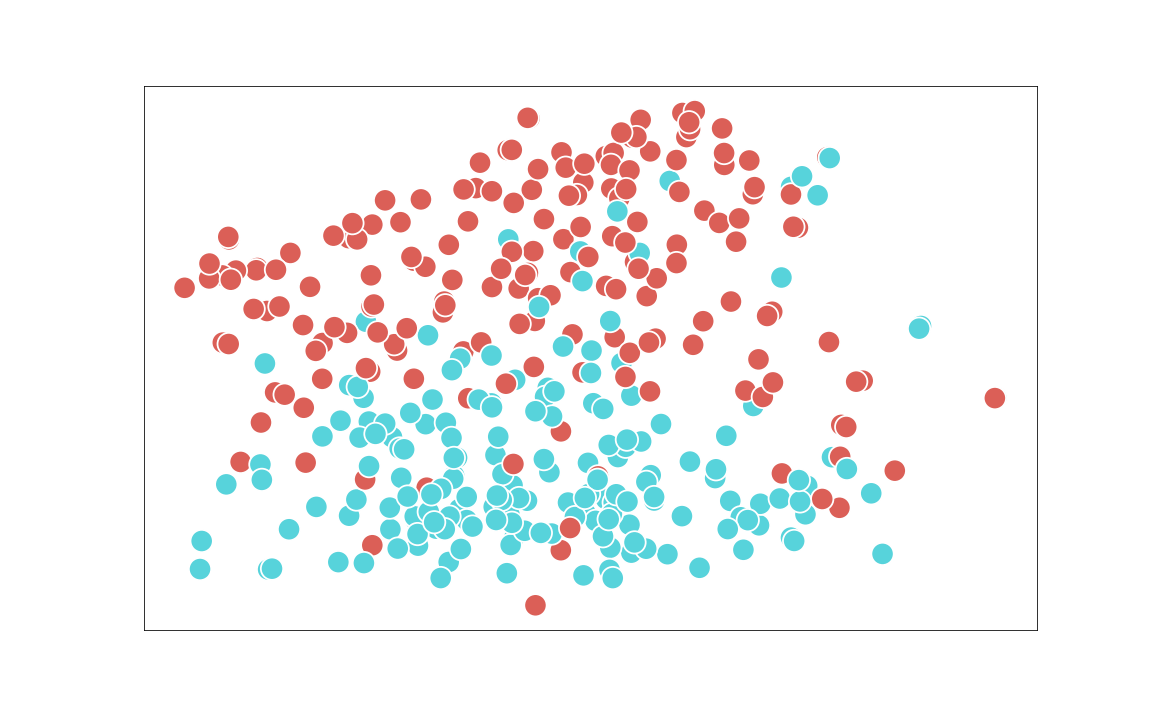}
\caption{$\phi_t^-$}
\end{subfigure}
\begin{subfigure}[t]{0.32\textwidth}
\centering
\includegraphics[width=\textwidth]{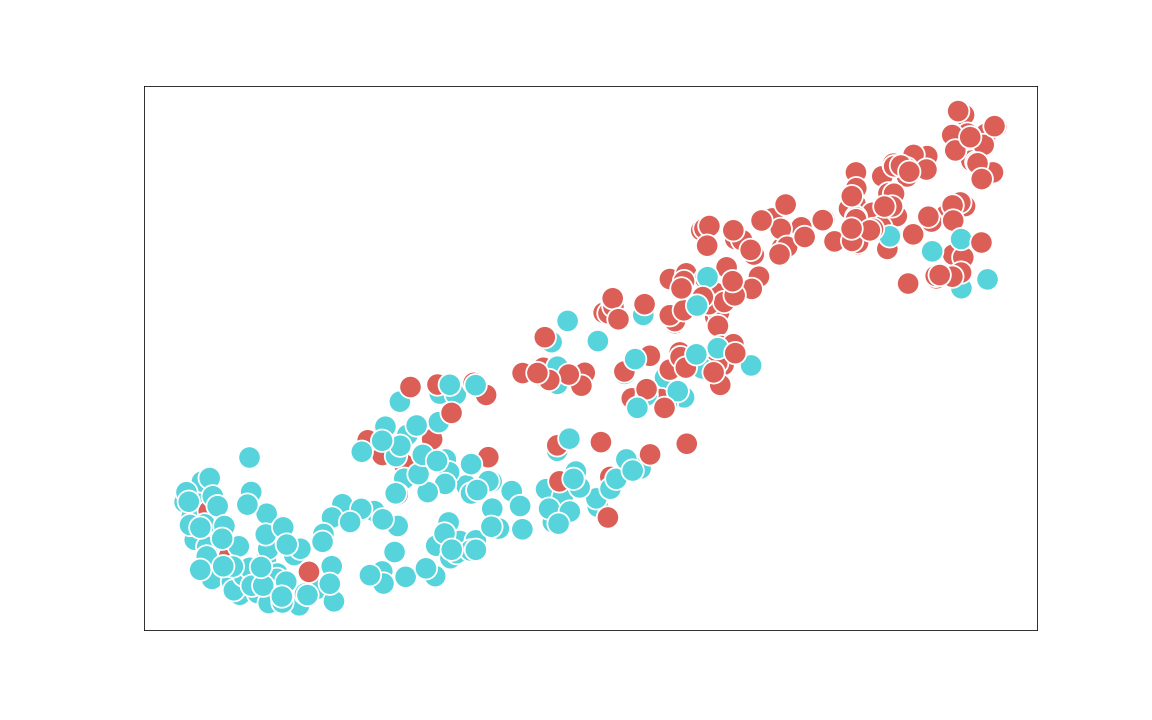}
\caption{$\phi_t$}
\end{subfigure}
\caption{t-SNE plots for ``Leave to left'' in DoTA}
\end{figure}
\begin{figure}[htbp]
\centering
\begin{subfigure}[t]{0.32\textwidth}
\centering
\includegraphics[width=\textwidth]{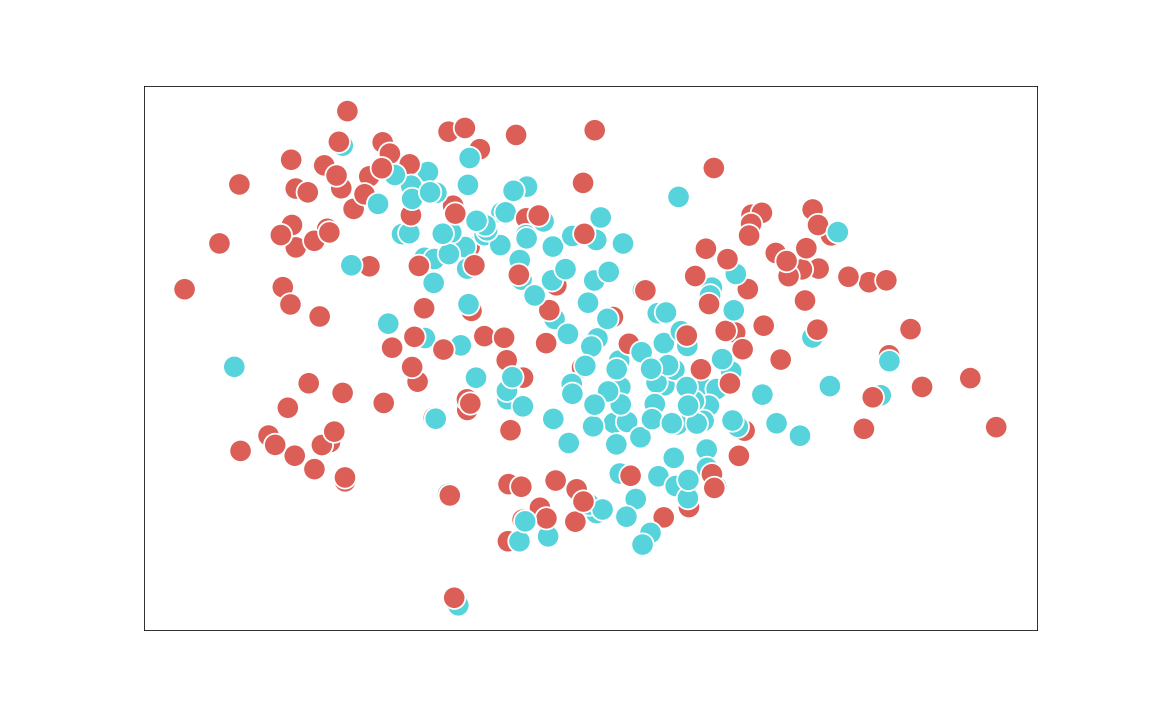}
\caption{$\phi_s$}
\end{subfigure}
\begin{subfigure}[t]{0.32\textwidth}
\centering
\includegraphics[width=\textwidth]{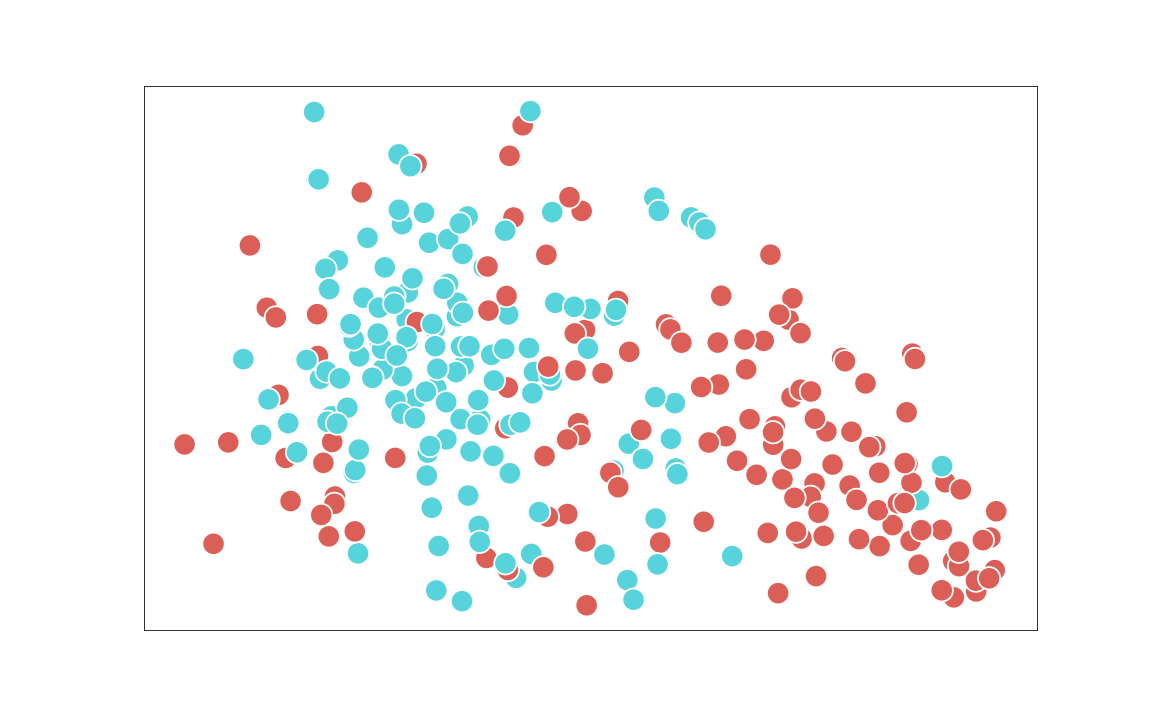}
\caption{$\phi_t^-$}
\end{subfigure}
\begin{subfigure}[t]{0.32\textwidth}
\centering
\includegraphics[width=\textwidth]{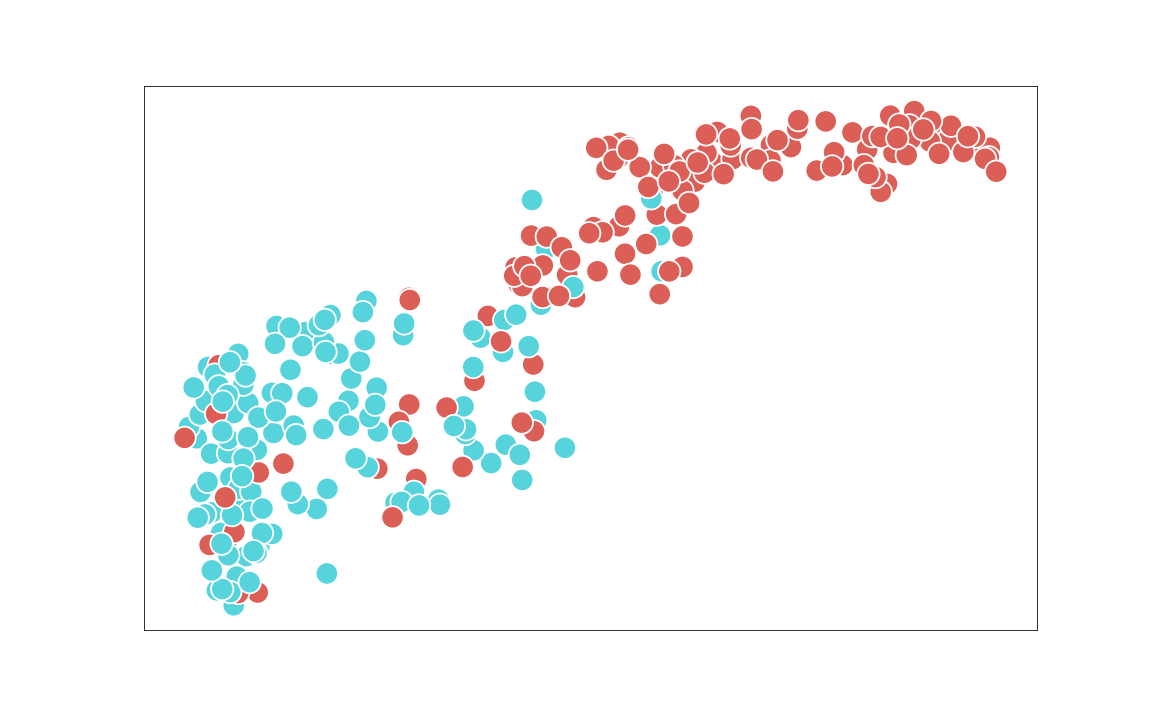}
\caption{$\phi_t$}
\end{subfigure}
\caption{t-SNE plots for ``Leave to right'' in DoTA}
\end{figure}
\begin{figure}[htbp]
\centering
\begin{subfigure}[t]{0.32\textwidth}
\centering
\includegraphics[width=\textwidth]{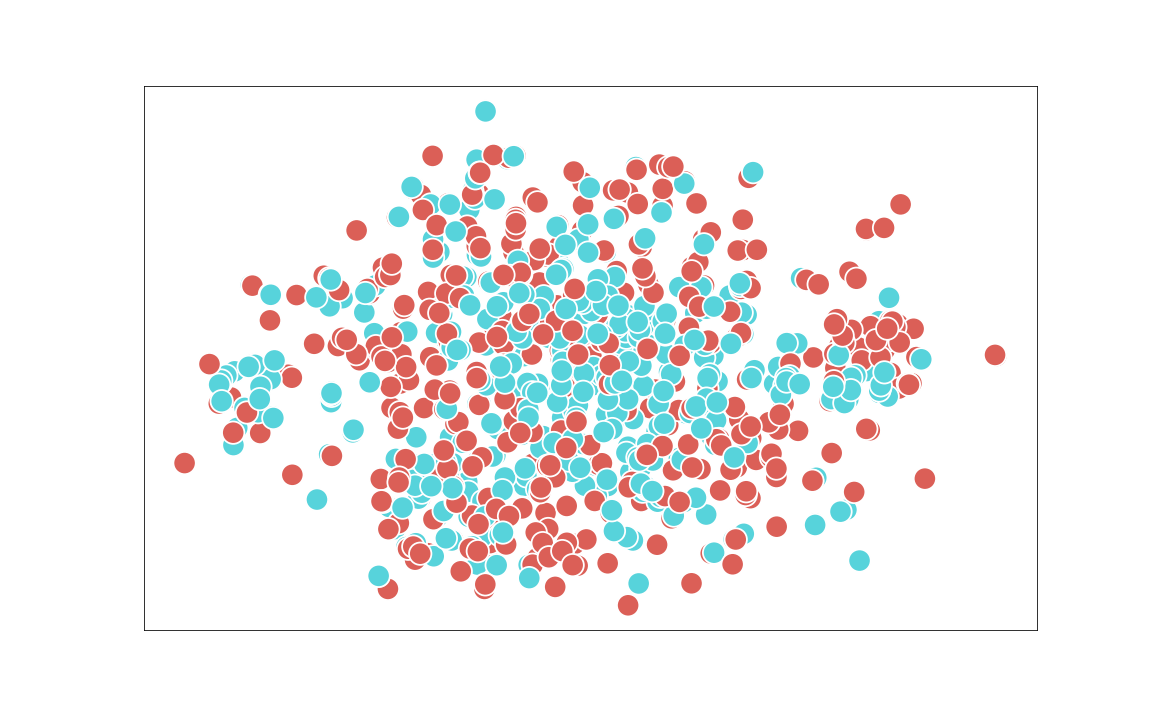}
\caption{$\phi_s$}
\end{subfigure}
\begin{subfigure}[t]{0.32\textwidth}
\centering
\includegraphics[width=\textwidth]{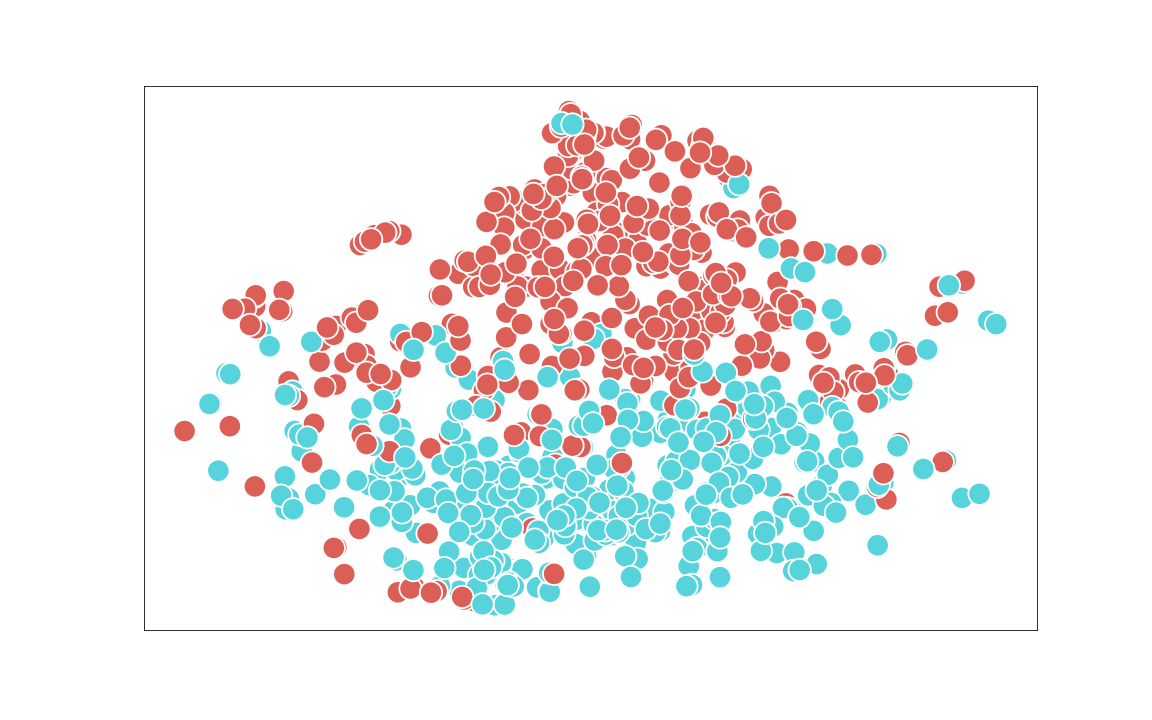}
\caption{$\phi_t^-$}
\end{subfigure}
\begin{subfigure}[t]{0.32\textwidth}
\centering
\includegraphics[width=\textwidth]{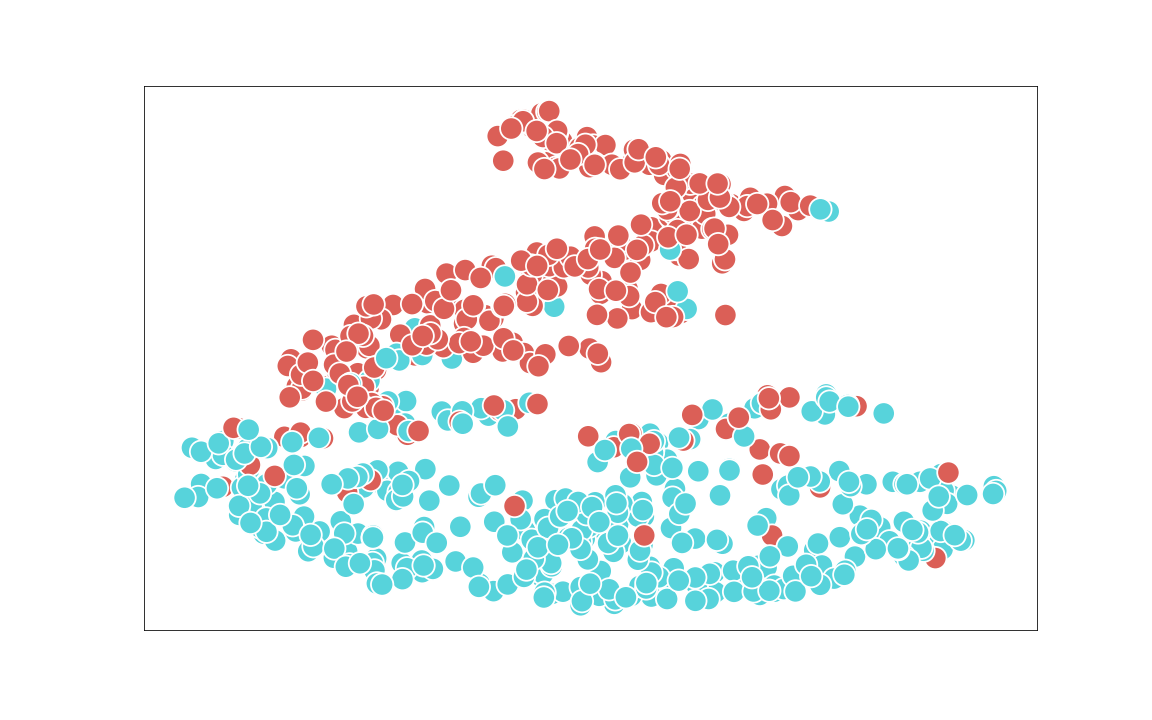}
\caption{$\phi_t$}
\end{subfigure}
\caption{t-SNE plots for ``Moving ahead or waiting'' in DoTA}
\end{figure}
\begin{figure}[htbp]
\centering
\begin{subfigure}[t]{0.32\textwidth}
\centering
\includegraphics[width=\textwidth]{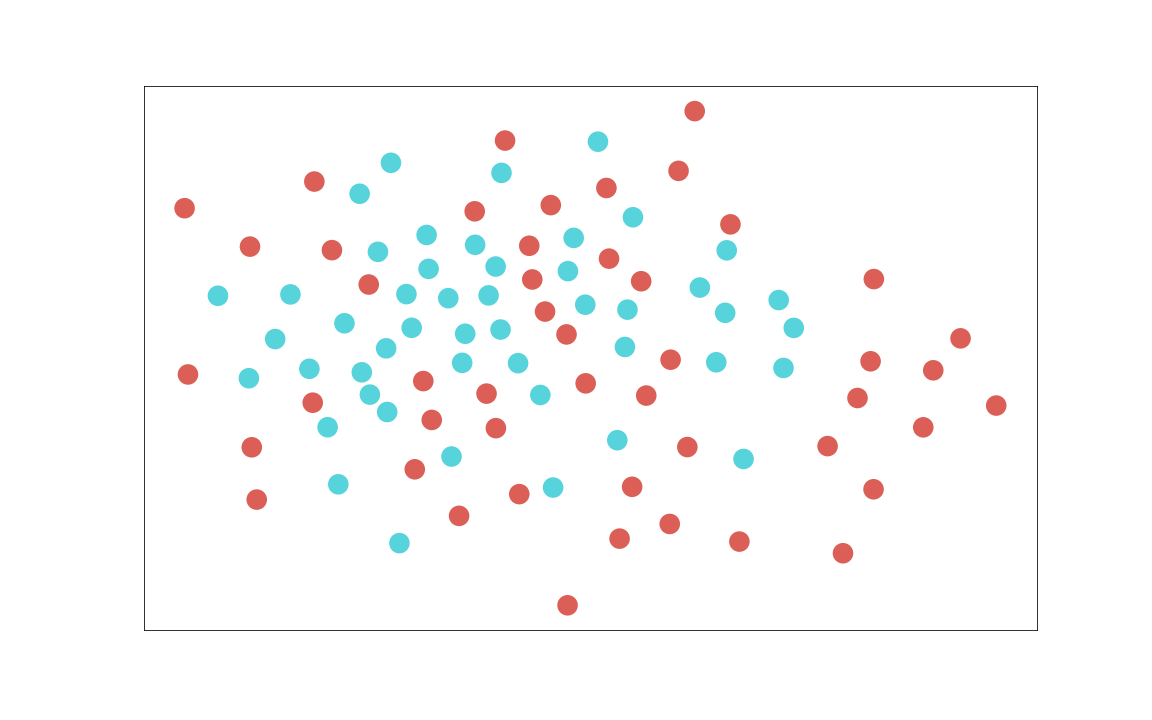}
\caption{$\phi_s$}
\end{subfigure}
\begin{subfigure}[t]{0.32\textwidth}
\centering
\includegraphics[width=\textwidth]{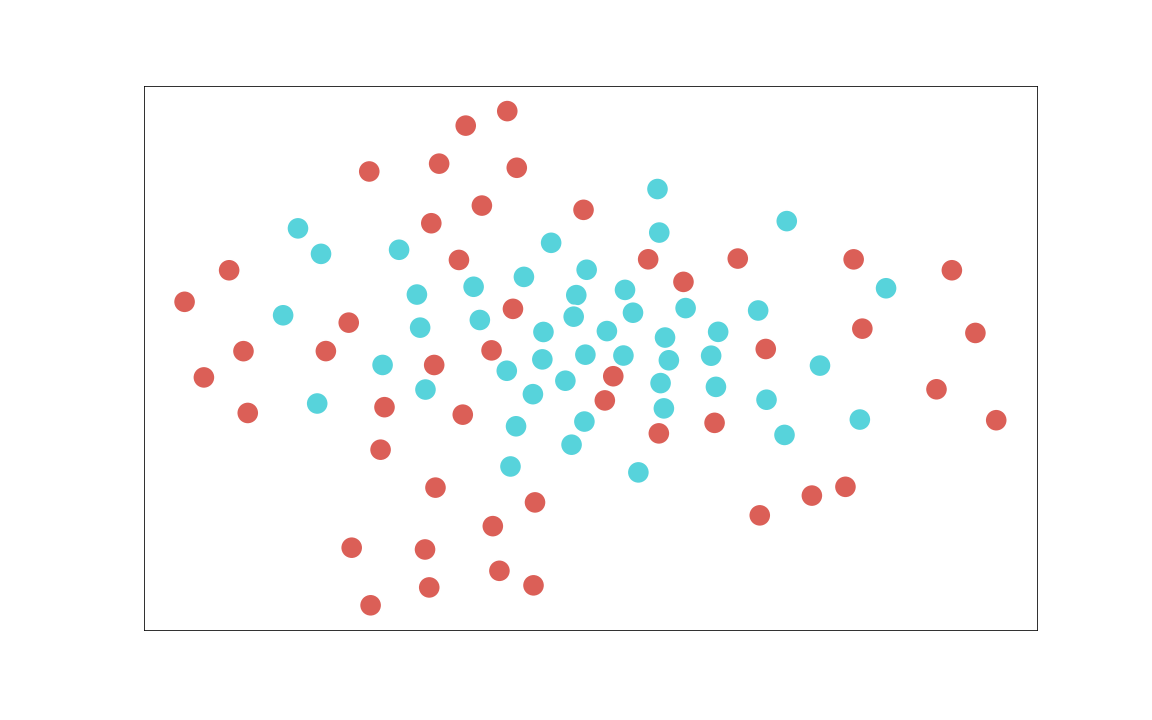}
\caption{$\phi_t^-$}
\end{subfigure}
\begin{subfigure}[t]{0.32\textwidth}
\centering
\includegraphics[width=\textwidth]{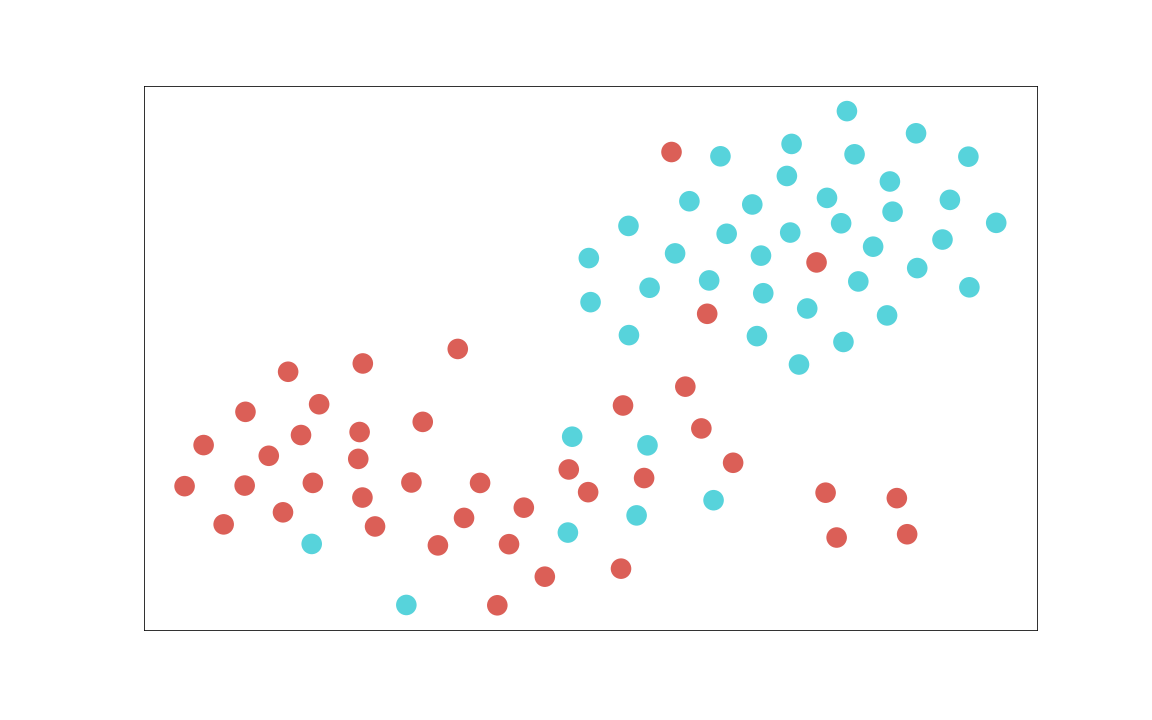}
\caption{$\phi_t$}
\end{subfigure}
\caption{t-SNE plots for ``Obstacle'' in DoTA}
\end{figure}
\begin{figure}[htbp]
\centering
\begin{subfigure}[t]{0.32\textwidth}
\centering
\includegraphics[width=\textwidth]{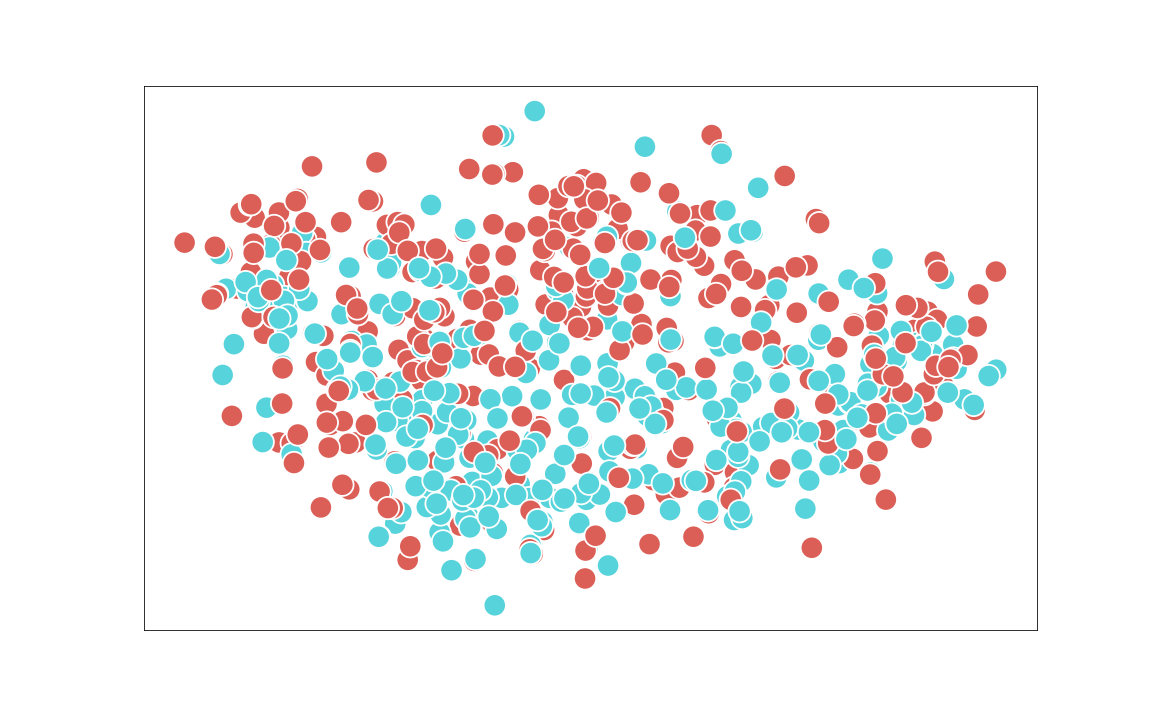}
\caption{$\phi_s$}
\end{subfigure}
\begin{subfigure}[t]{0.32\textwidth}
\centering
\includegraphics[width=\textwidth]{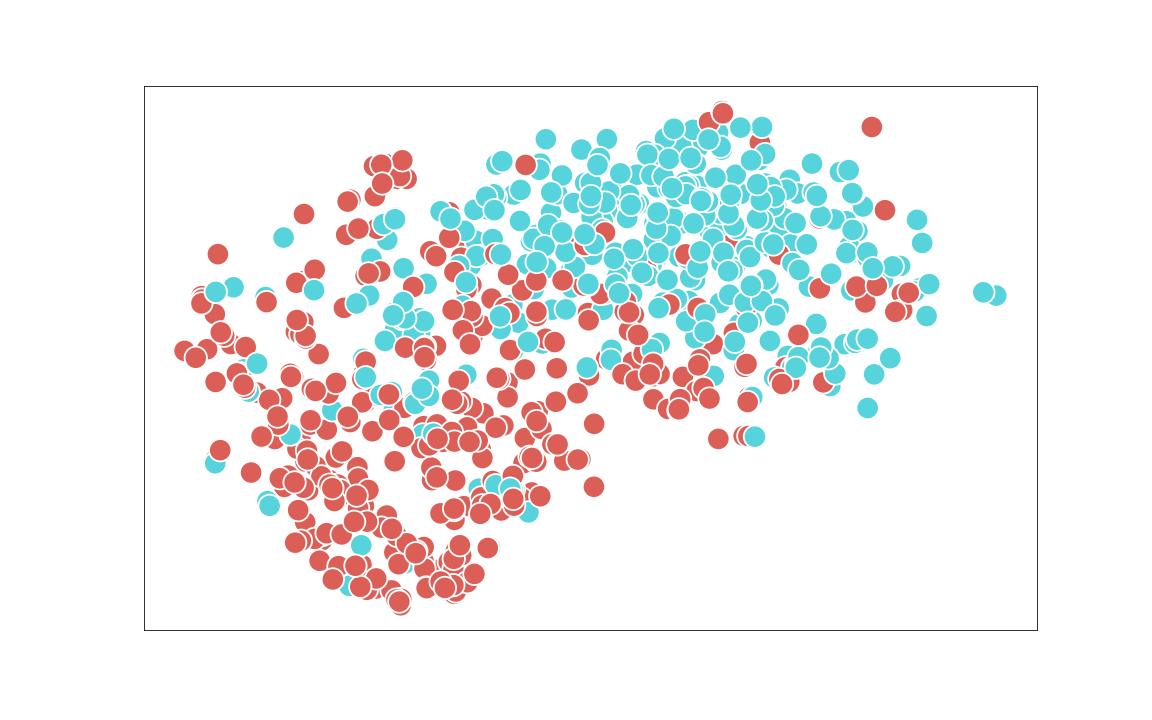}
\caption{$\phi_t^-$}
\end{subfigure}
\begin{subfigure}[t]{0.32\textwidth}
\centering
\includegraphics[width=\textwidth]{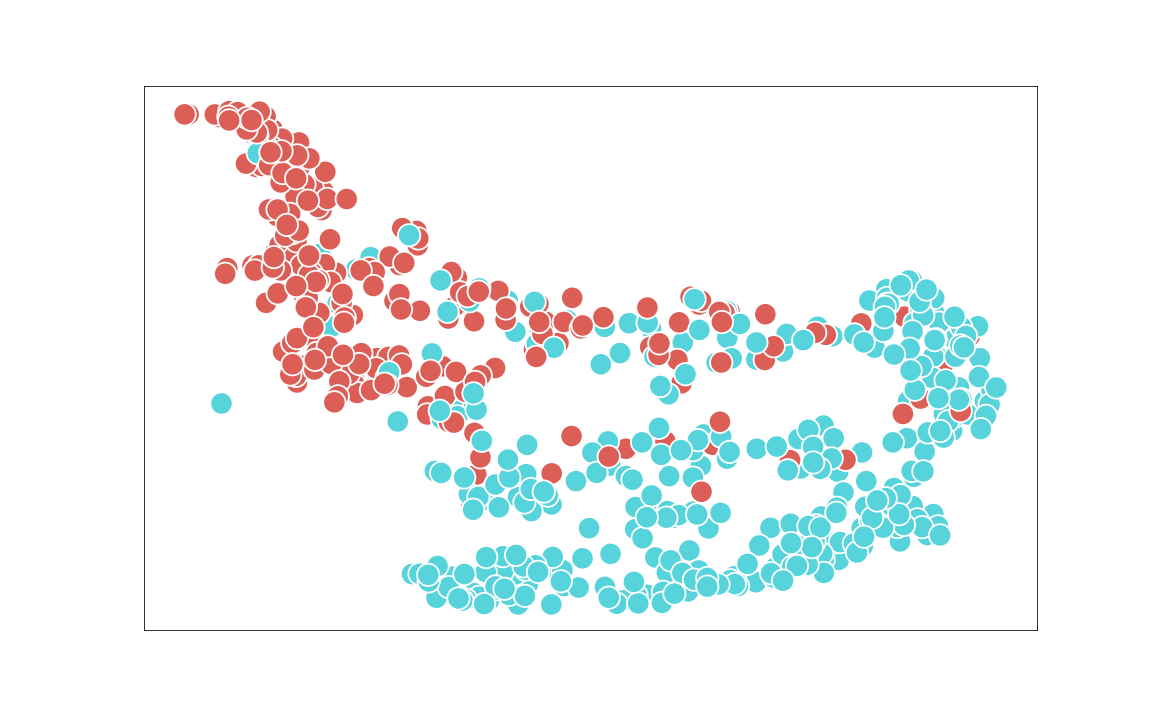}
\caption{$\phi_t$}
\end{subfigure}
\caption{t-SNE plots for ``Oncoming'' in DoTA}
\end{figure}
\begin{figure}[htbp]
\centering
\begin{subfigure}[t]{0.32\textwidth}
\centering
\includegraphics[width=\textwidth]{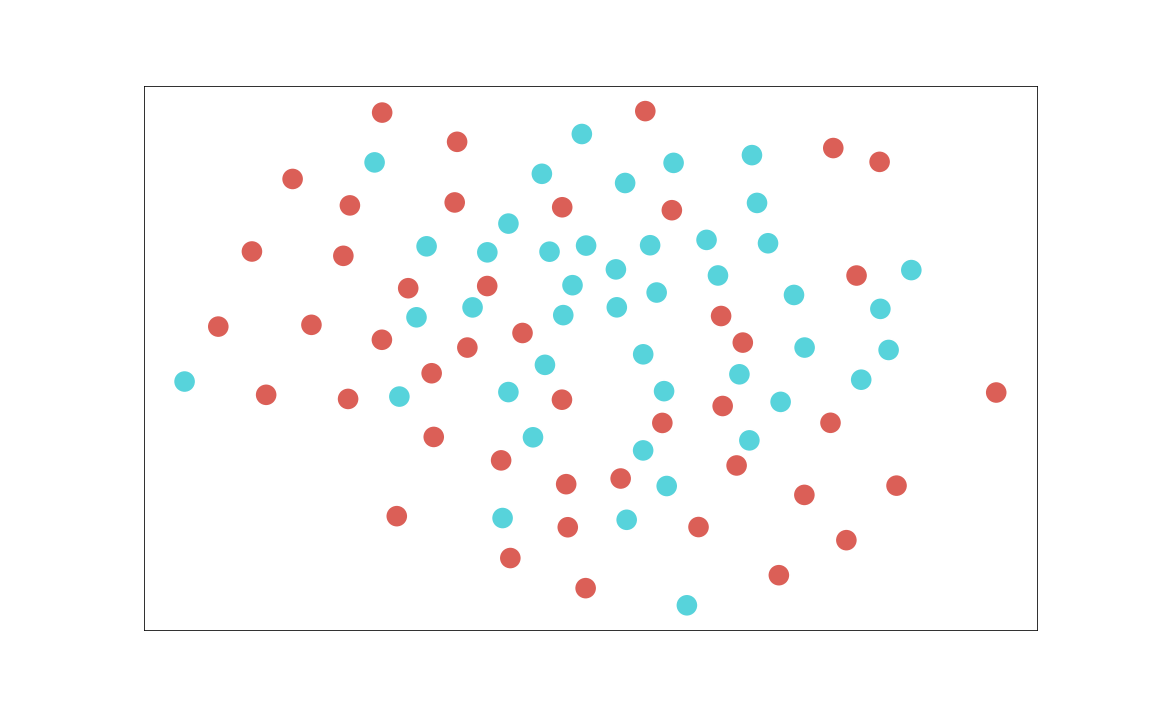}
\caption{$\phi_s$}
\end{subfigure}
\begin{subfigure}[t]{0.32\textwidth}
\centering
\includegraphics[width=\textwidth]{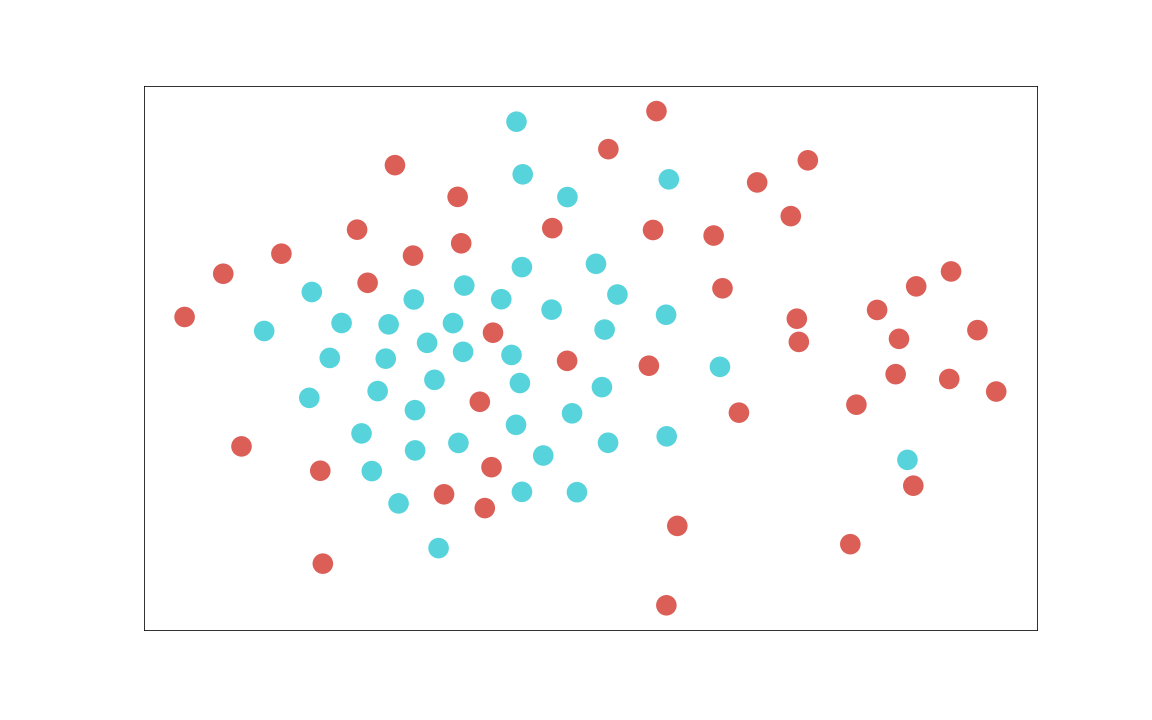}
\caption{$\phi_t^-$}
\end{subfigure}
\begin{subfigure}[t]{0.32\textwidth}
\centering
\includegraphics[width=\textwidth]{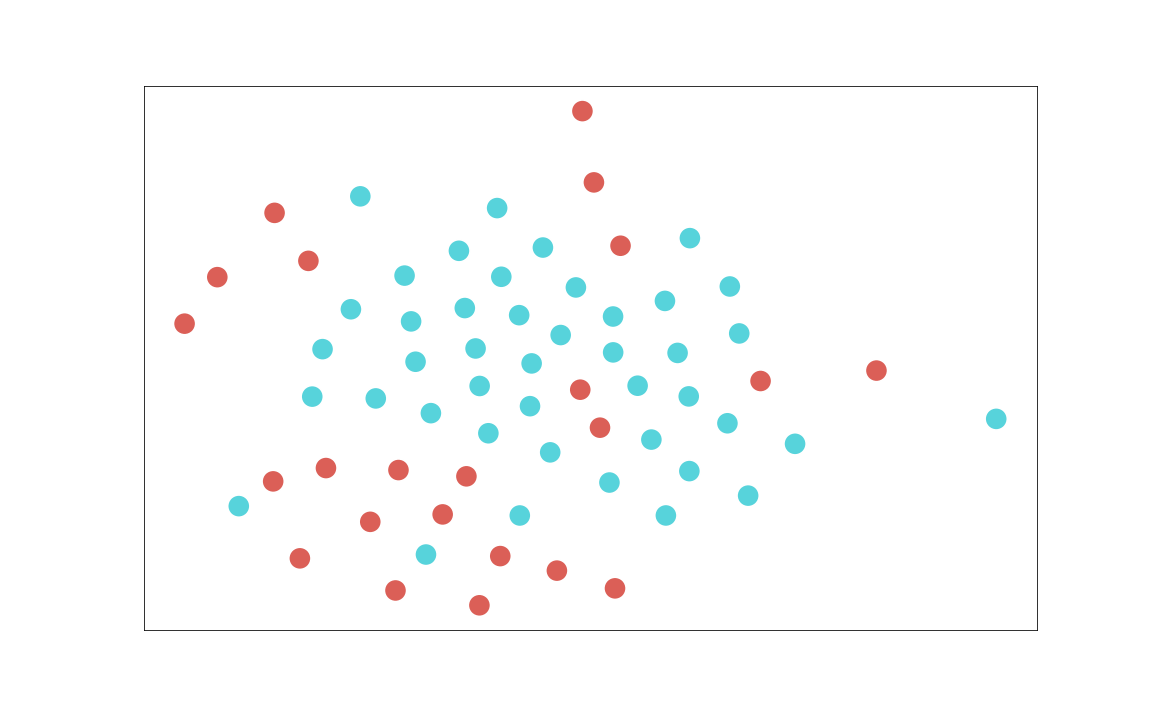}
\caption{$\phi_t$}
\end{subfigure}
\caption{t-SNE plots for ``Pedestrian'' in DoTA}
\end{figure}
\begin{figure}[htbp]
\centering
\begin{subfigure}[t]{0.32\textwidth}
\centering
\includegraphics[width=\textwidth]{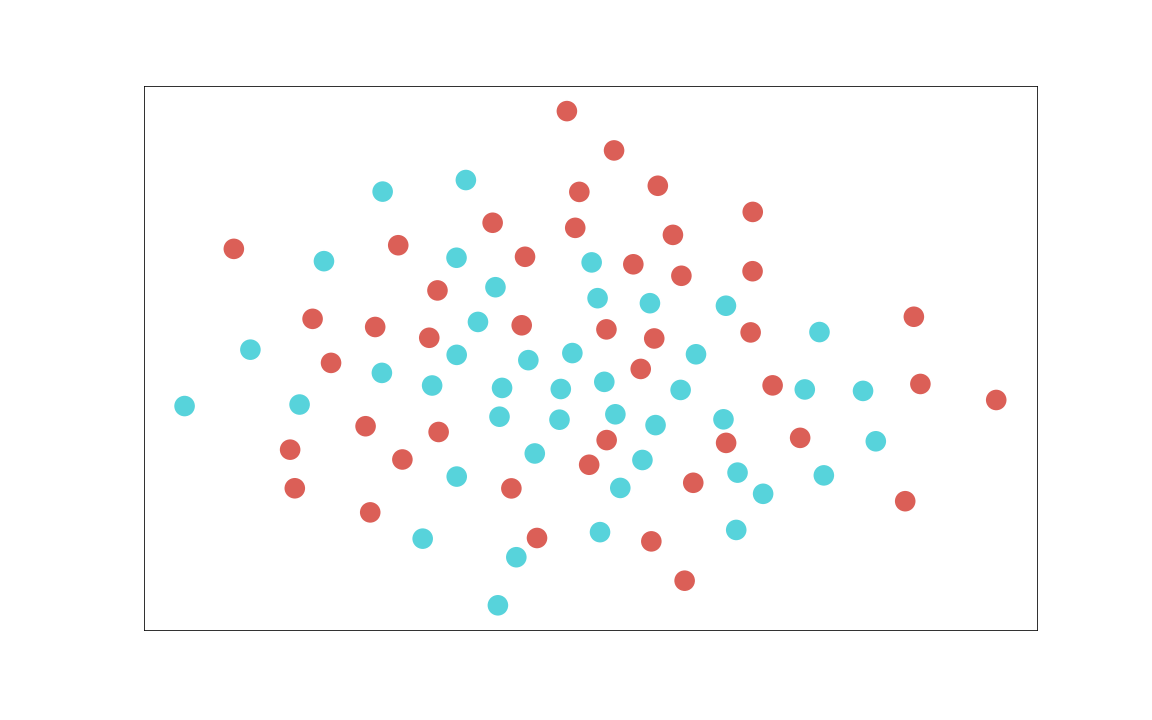}
\caption{$\phi_s$}
\end{subfigure}
\begin{subfigure}[t]{0.32\textwidth}
\centering
\includegraphics[width=\textwidth]{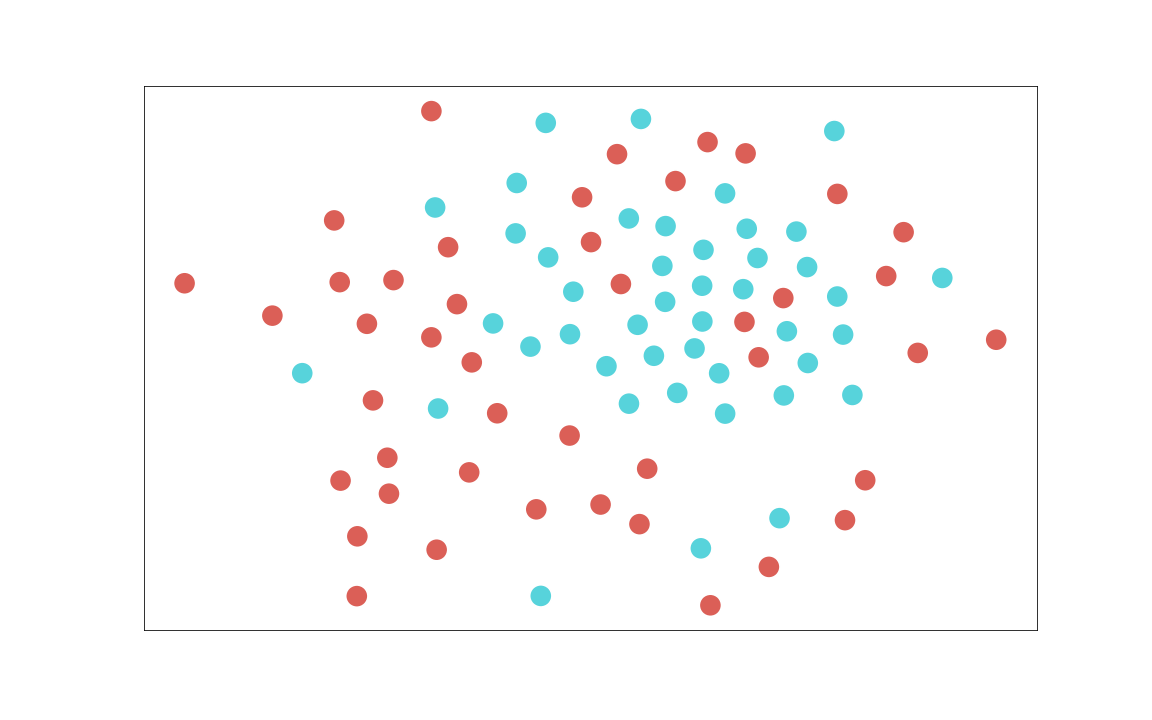}
\caption{$\phi_t^-$}
\end{subfigure}
\begin{subfigure}[t]{0.32\textwidth}
\centering
\includegraphics[width=\textwidth]{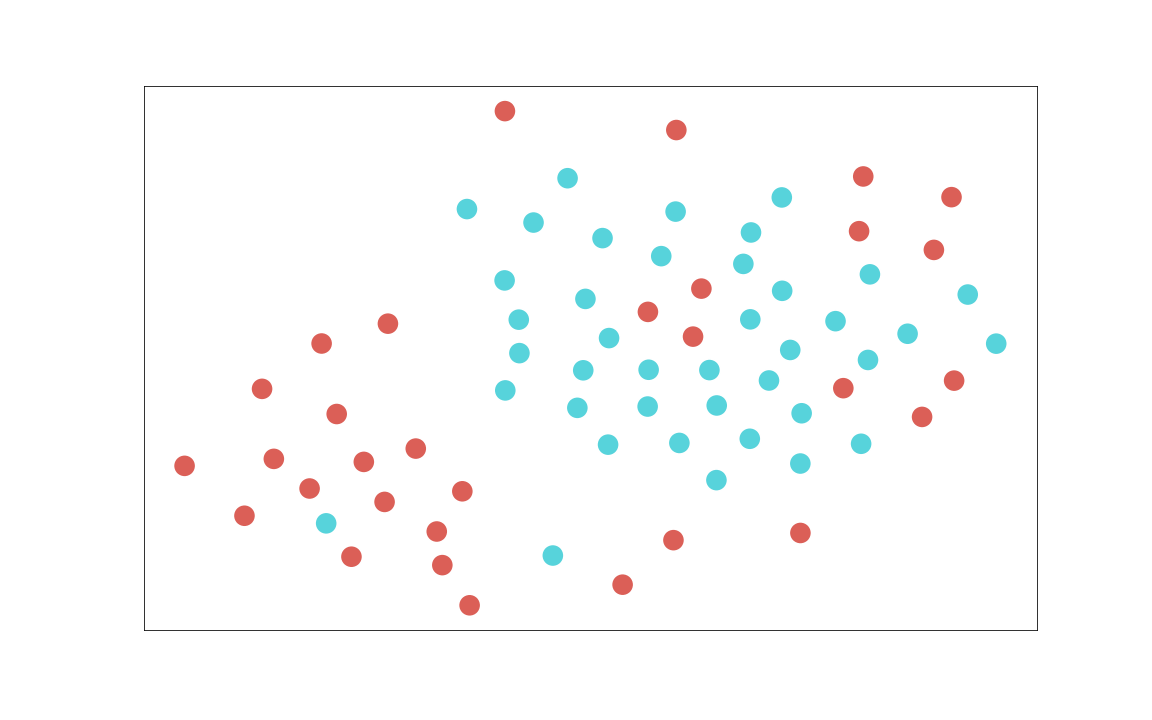}
\caption{$\phi_t$}
\end{subfigure}
\caption{t-SNE plots for ``Start Stop or Stationary'' in DoTA}
\end{figure}
\begin{figure}[htbp]
\centering
\begin{subfigure}[t]{0.32\textwidth}
\centering
\includegraphics[width=\textwidth]{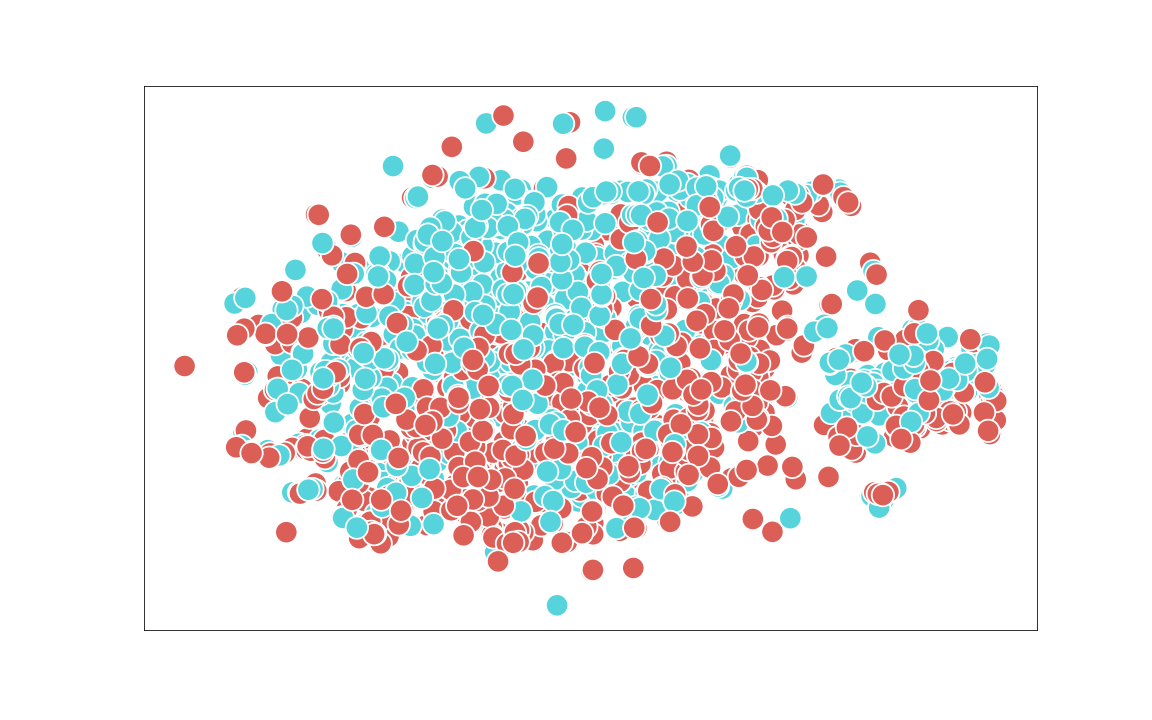}
\caption{$\phi_s$}
\end{subfigure}
\begin{subfigure}[t]{0.32\textwidth}
\centering
\includegraphics[width=\textwidth]{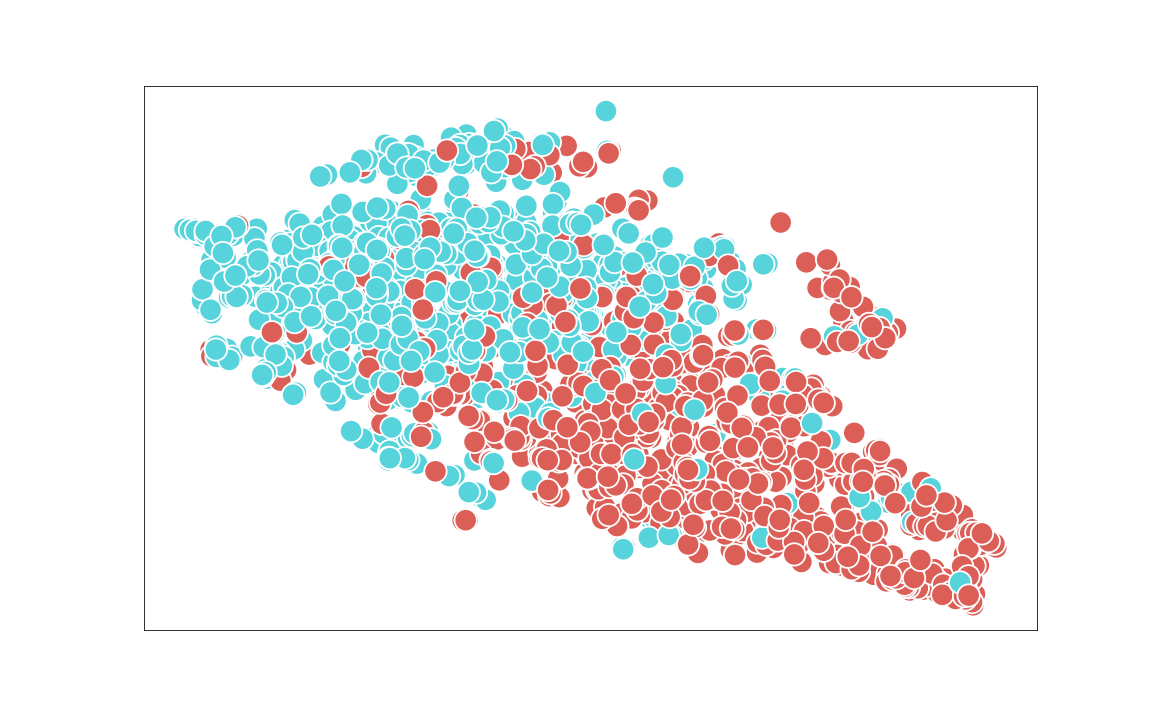}
\caption{$\phi_t^-$}
\end{subfigure}
\begin{subfigure}[t]{0.32\textwidth}
\centering
\includegraphics[width=\textwidth]{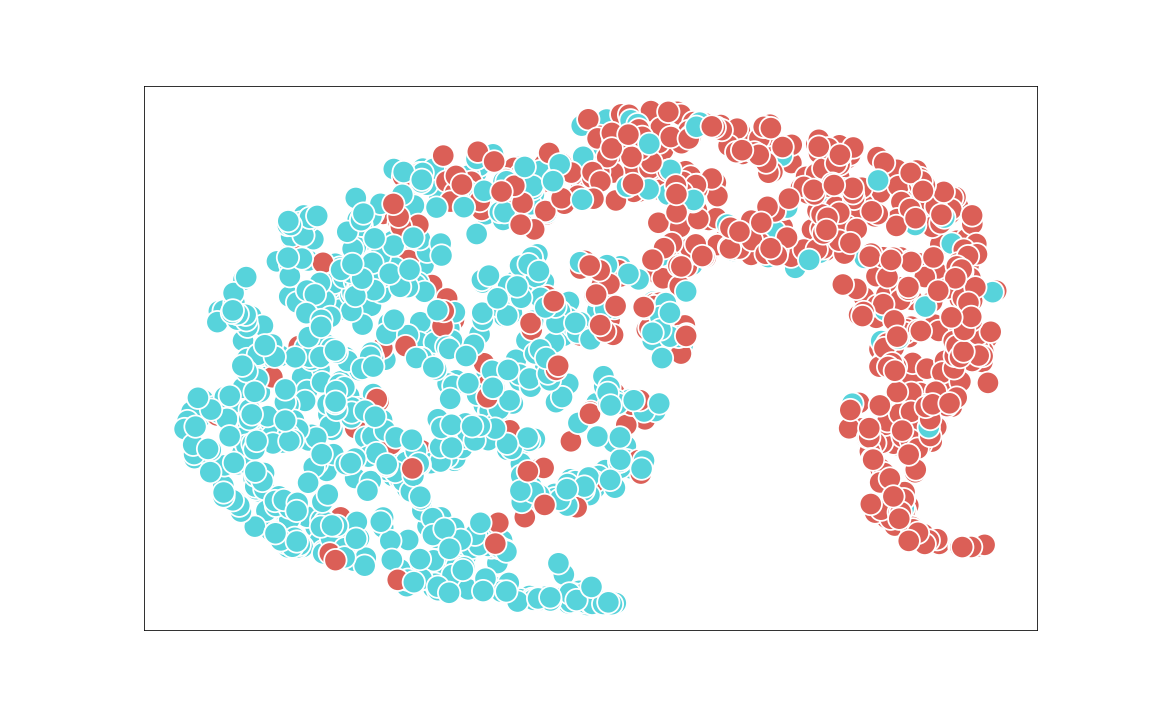}
\caption{$\phi_t$}
\end{subfigure}
\caption{t-SNE plots for ``Turning'' in DoTA}
\end{figure}
\begin{figure}[htbp]
\centering
\begin{subfigure}[t]{0.32\textwidth}
\centering
\includegraphics[width=\textwidth]{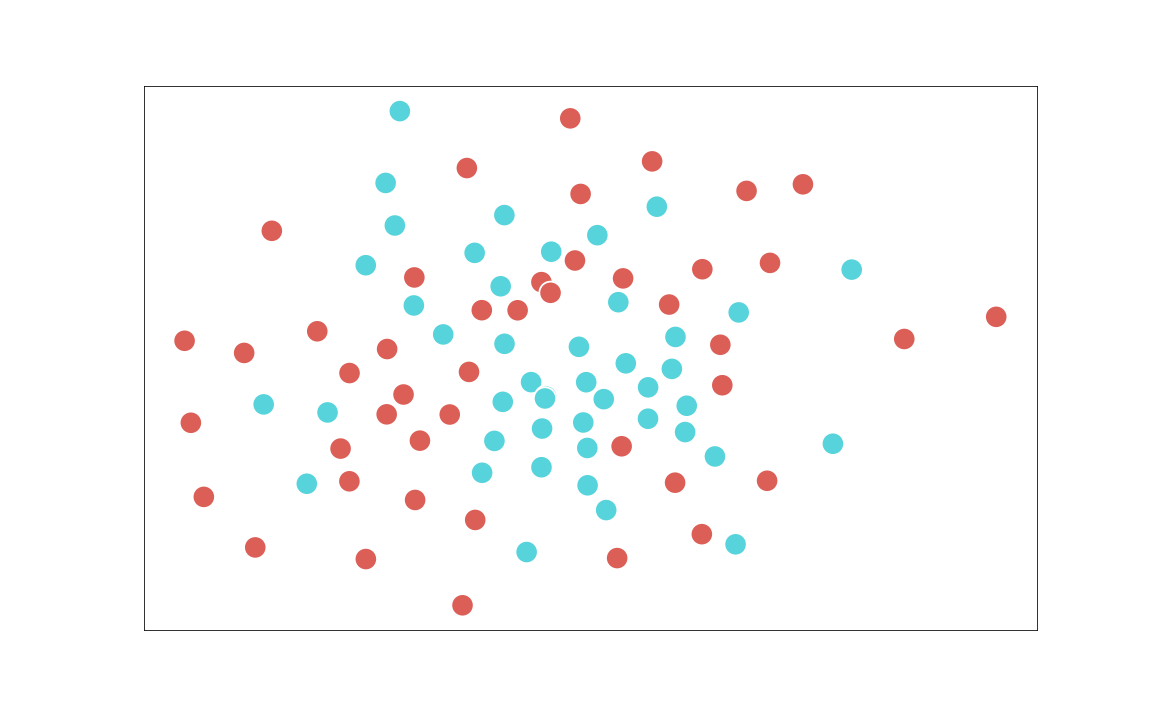}
\caption{$\phi_s$}
\end{subfigure}
\begin{subfigure}[t]{0.32\textwidth}
\centering
\includegraphics[width=\textwidth]{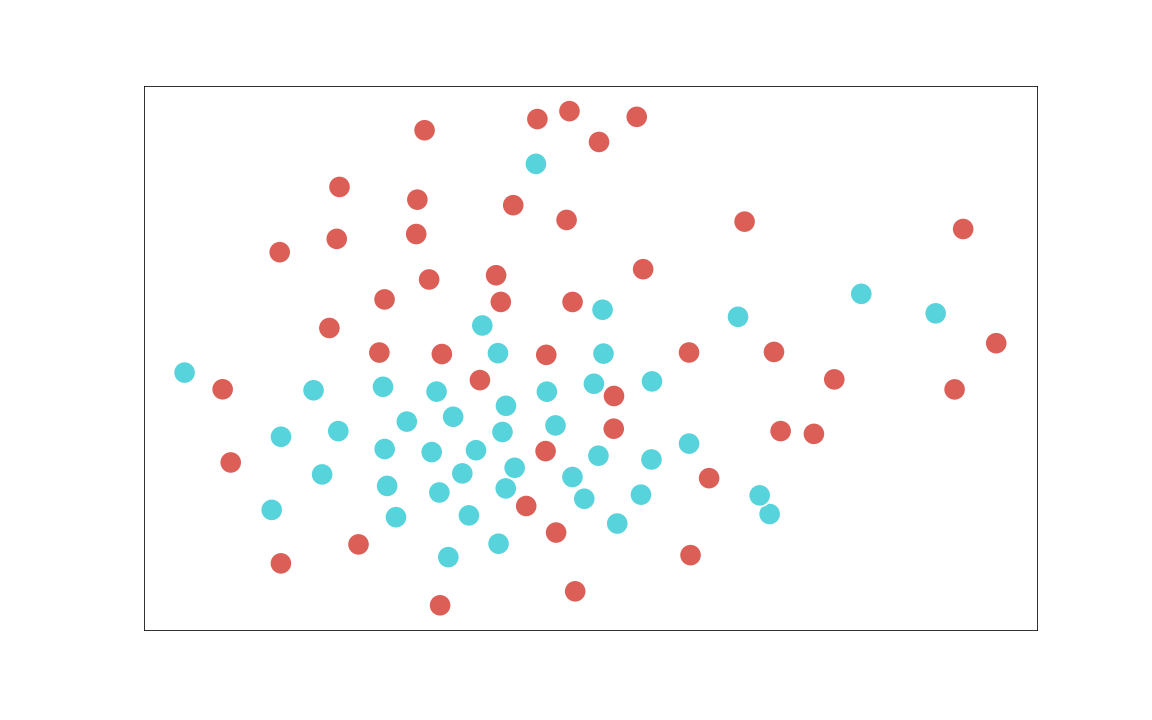}
\caption{$\phi_t^-$}
\end{subfigure}
\begin{subfigure}[t]{0.32\textwidth}
\centering
\includegraphics[width=\textwidth]{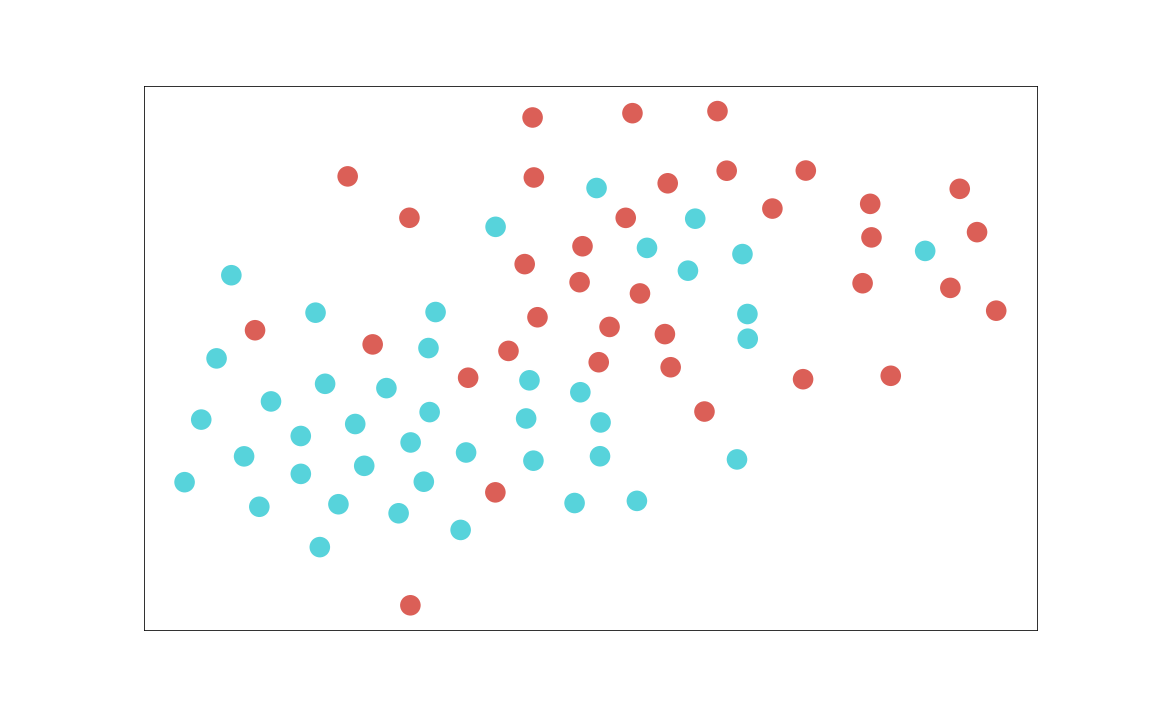}
\caption{$\phi_t$}
\end{subfigure}
\caption{t-SNE plots for ``Unknown'' in DoTA}
\end{figure}
\begin{figure}[htbp]
\centering
\begin{subfigure}[t]{0.32\textwidth}
\centering
\includegraphics[width=\textwidth]{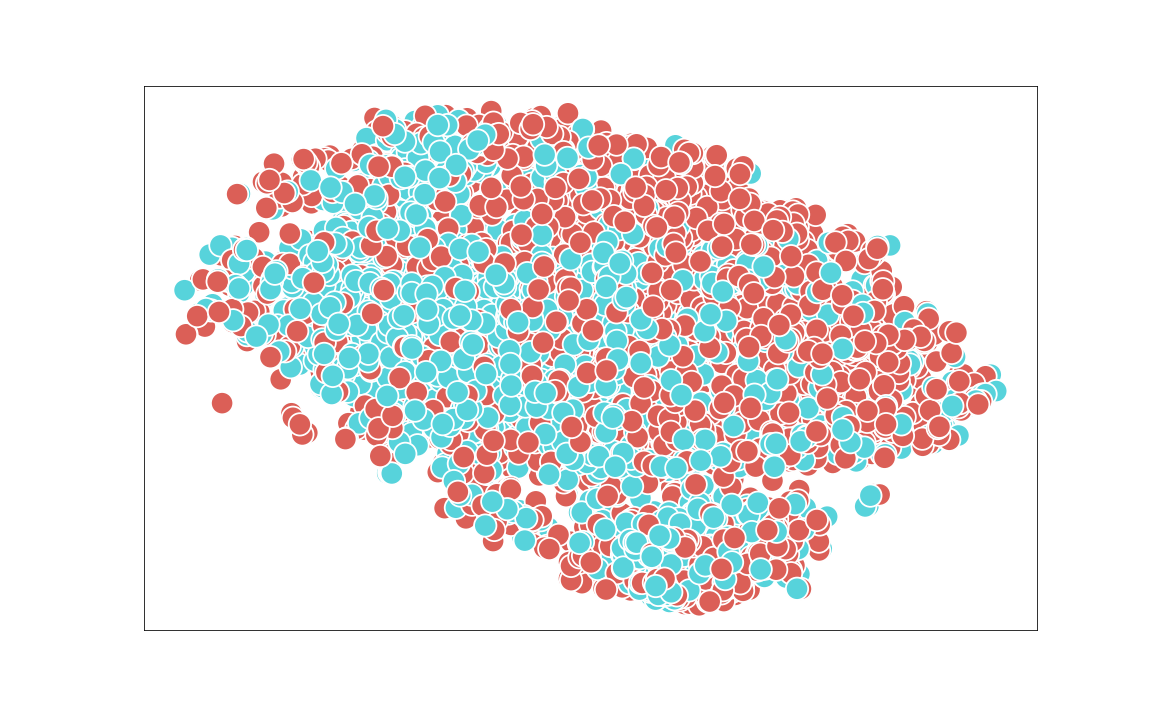}
\caption{$\phi_s$}
\end{subfigure}
\begin{subfigure}[t]{0.32\textwidth}
\centering
\includegraphics[width=\textwidth]{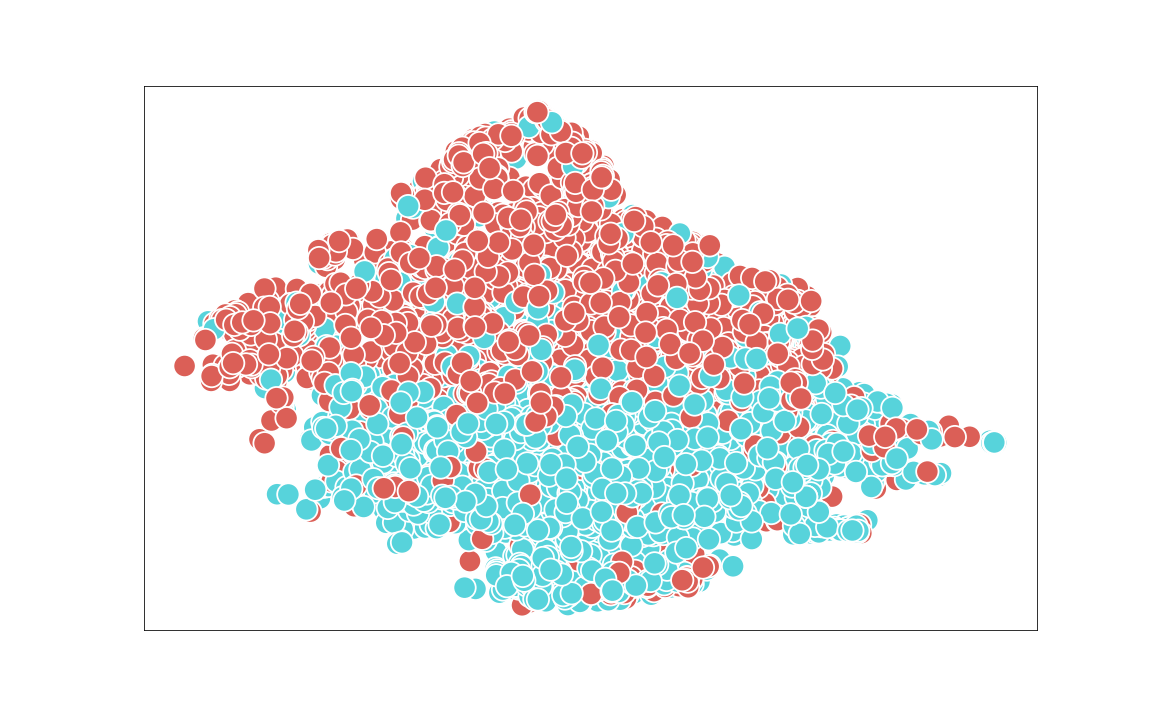}
\caption{$\phi_t^-$}
\end{subfigure}
\begin{subfigure}[t]{0.32\textwidth}
\centering
\includegraphics[width=\textwidth]{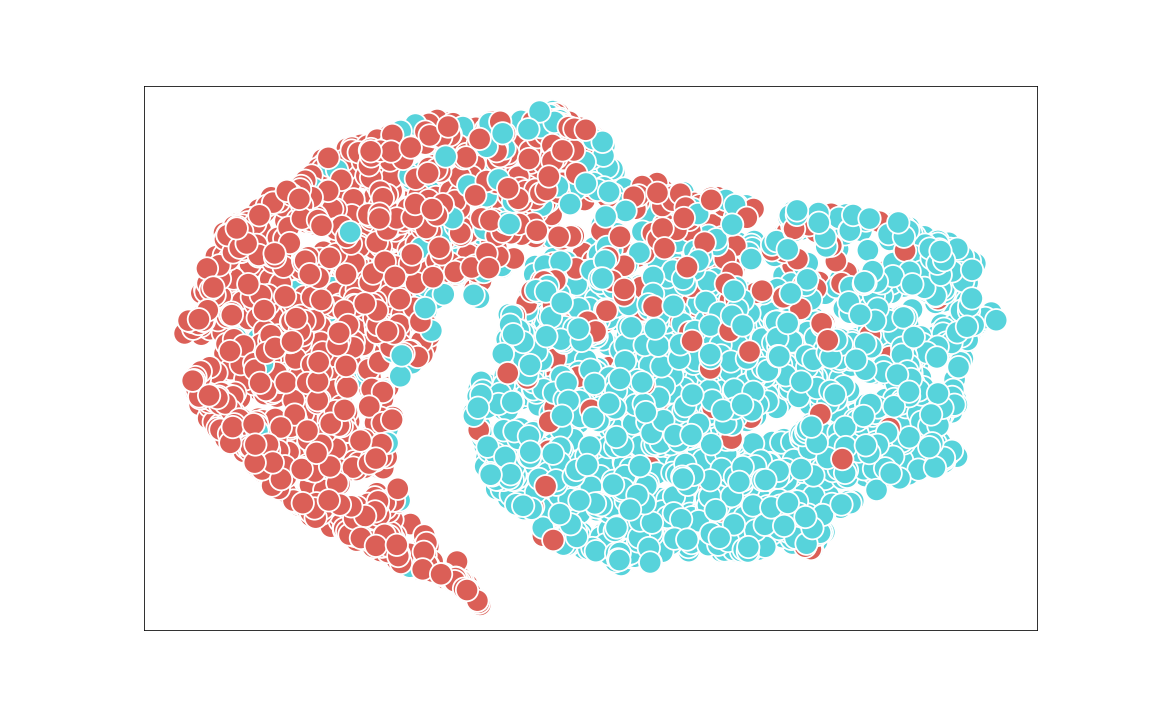}
\caption{$\phi_t$}
\end{subfigure}
\caption{t-SNE plots for ``All Types'' in DoTA}
\end{figure}

\begin{figure}[htbp]
\centering
\begin{subfigure}[t]{0.32\textwidth}
\centering
\includegraphics[width=\textwidth]{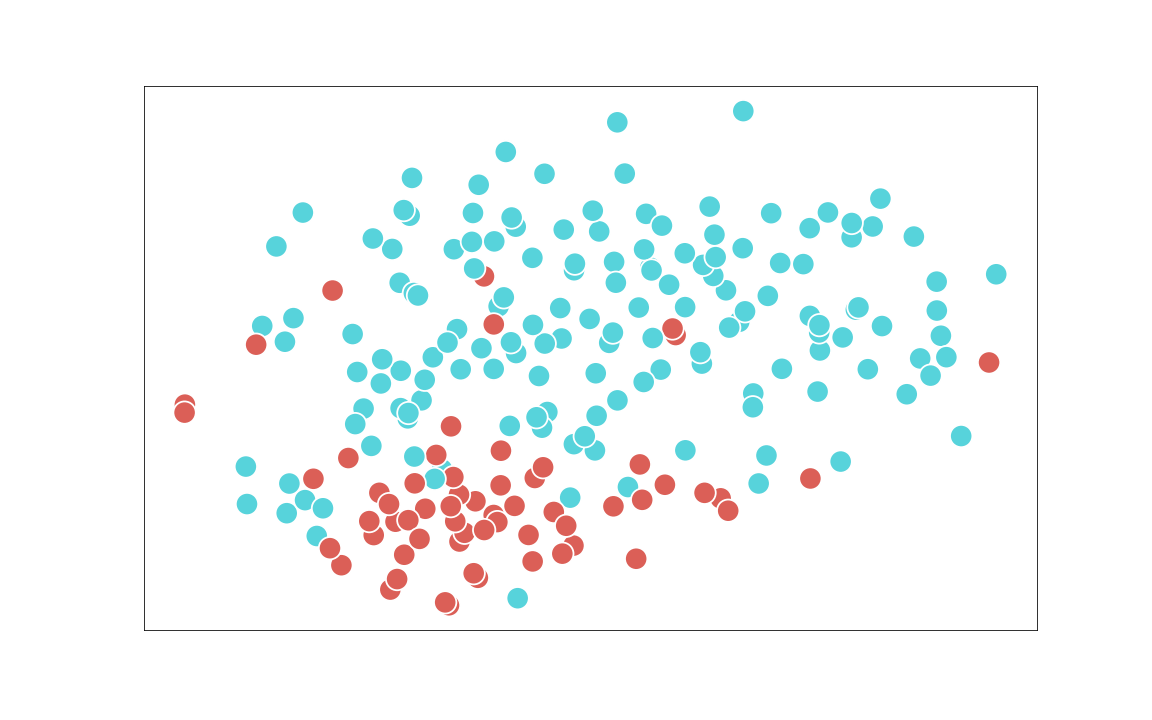}
\caption{$\phi_s$}
\end{subfigure}
\begin{subfigure}[t]{0.32\textwidth}
\centering
\includegraphics[width=\textwidth]{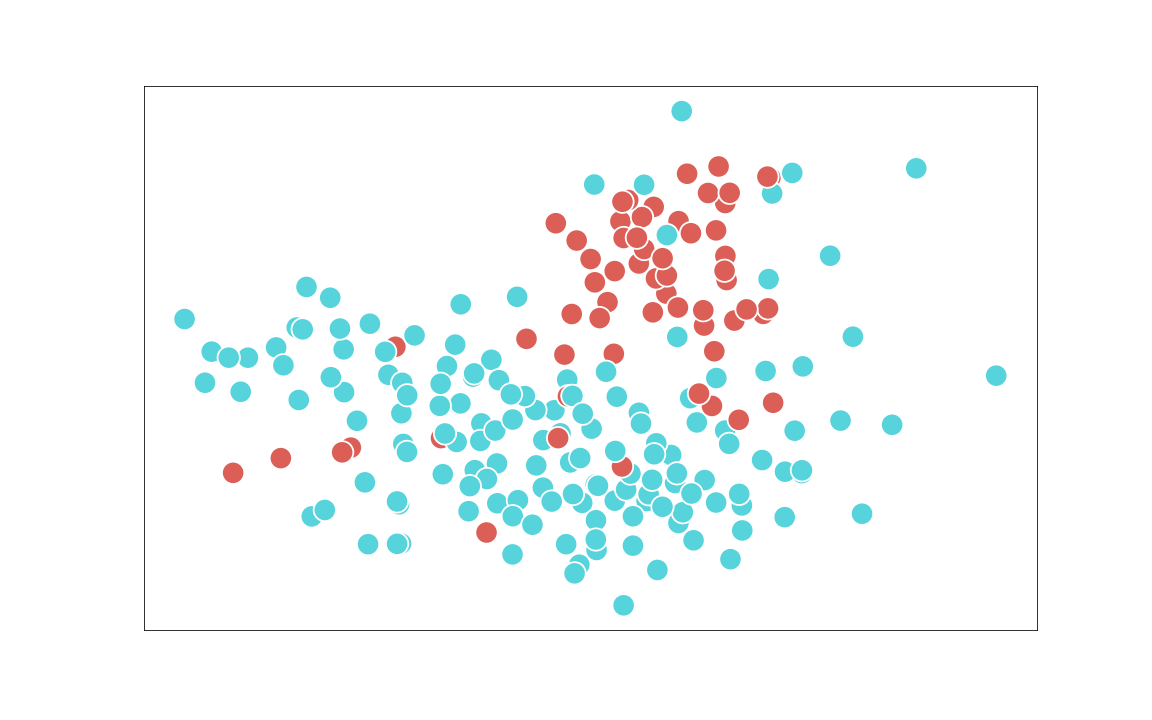}
\caption{$\phi_t^-$}
\end{subfigure}
\begin{subfigure}[t]{0.32\textwidth}
\centering
\includegraphics[width=\textwidth]{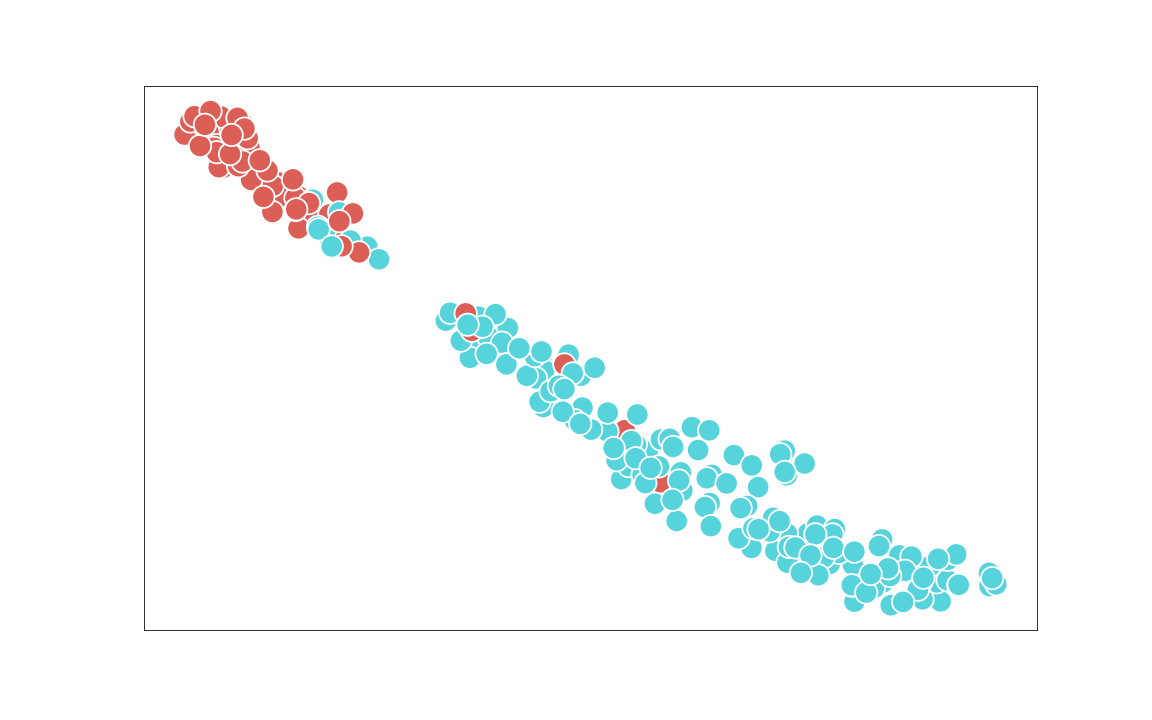}
\caption{$\phi_t$}
\end{subfigure}
\caption{t-SNE plots for ``Arson'' in UCF Crime}
\end{figure}
\begin{figure}[htbp]
\centering
\begin{subfigure}[t]{0.32\textwidth}
\centering
\includegraphics[width=\textwidth]{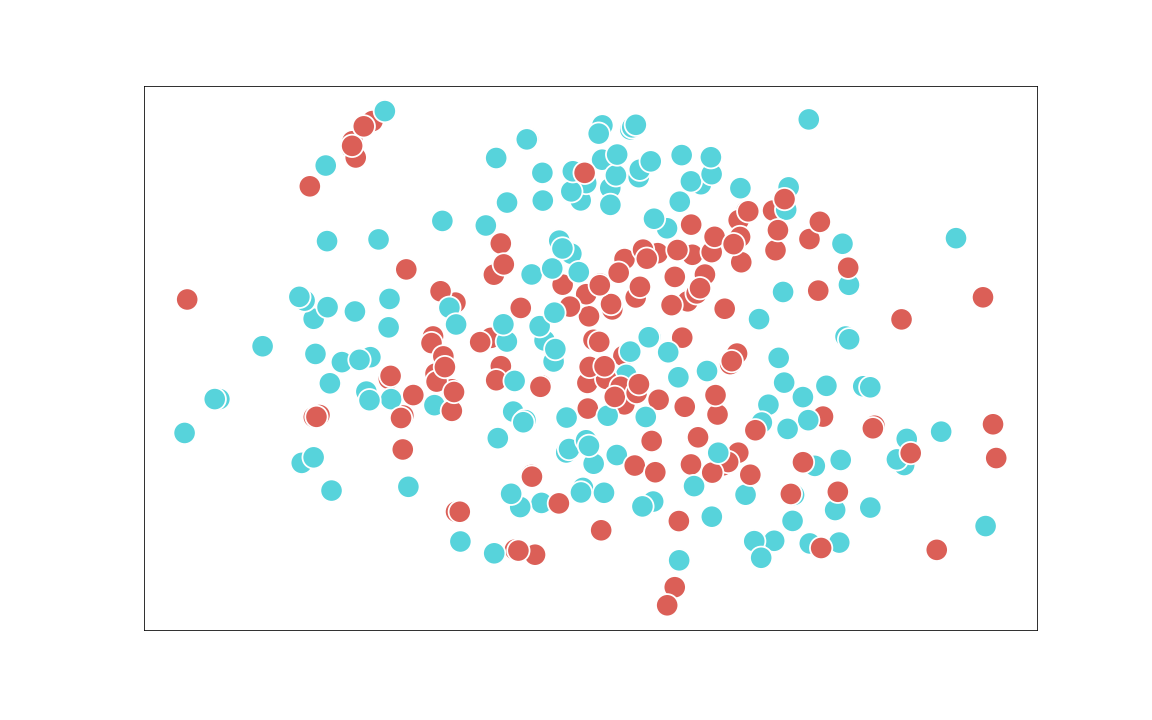}
\caption{$\phi_s$}
\end{subfigure}
\begin{subfigure}[t]{0.32\textwidth}
\centering
\includegraphics[width=\textwidth]{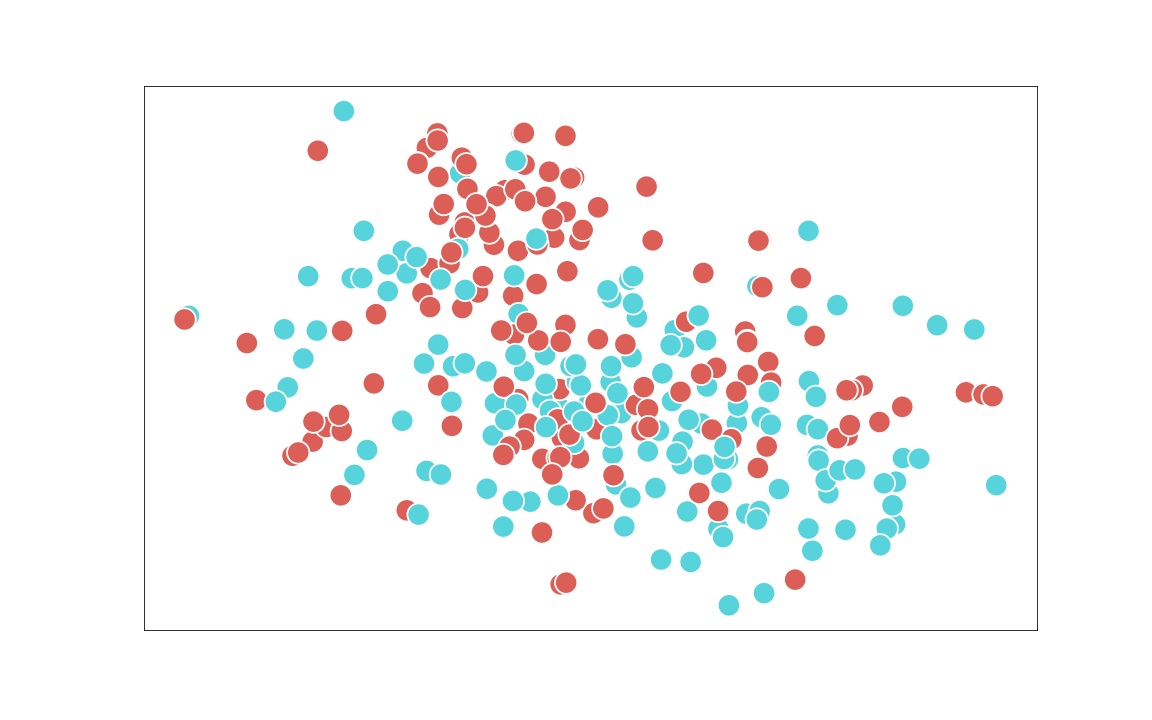}
\caption{$\phi_t^-$}
\end{subfigure}
\begin{subfigure}[t]{0.32\textwidth}
\centering
\includegraphics[width=\textwidth]{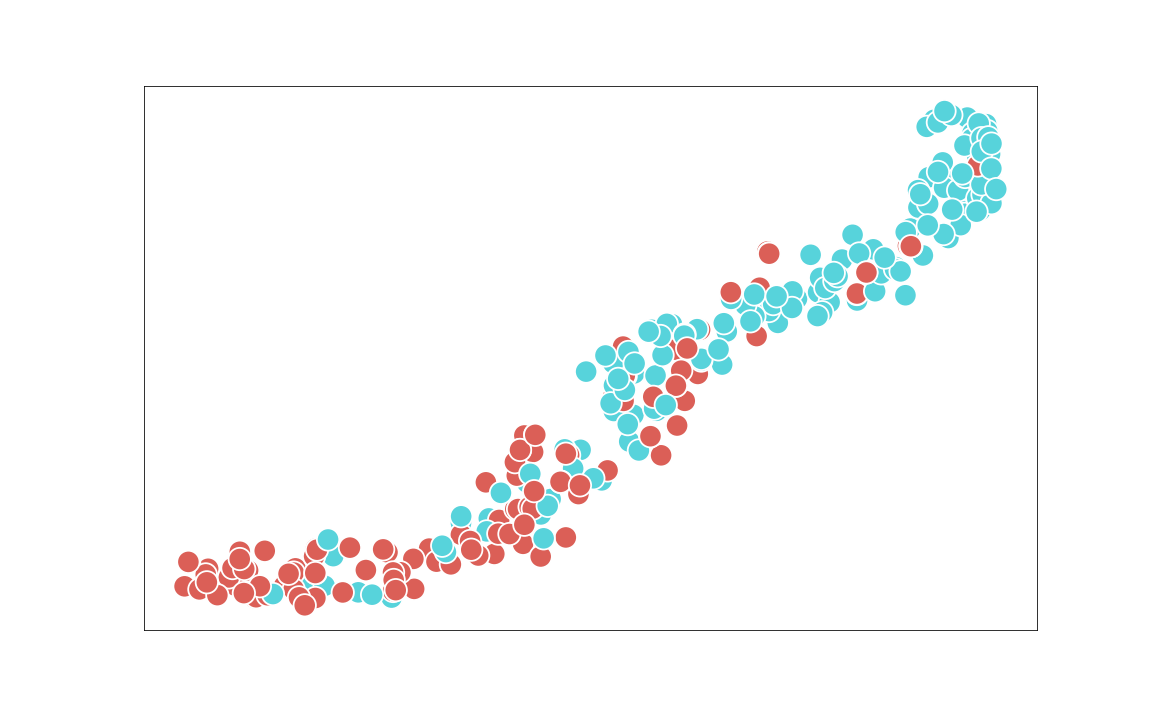}
\caption{$\phi_t$}
\end{subfigure}
\caption{t-SNE plots for ``Burglary'' in UCF Crime}
\end{figure}
\begin{figure}[htbp]
\centering
\begin{subfigure}[t]{0.32\textwidth}
\centering
\includegraphics[width=\textwidth]{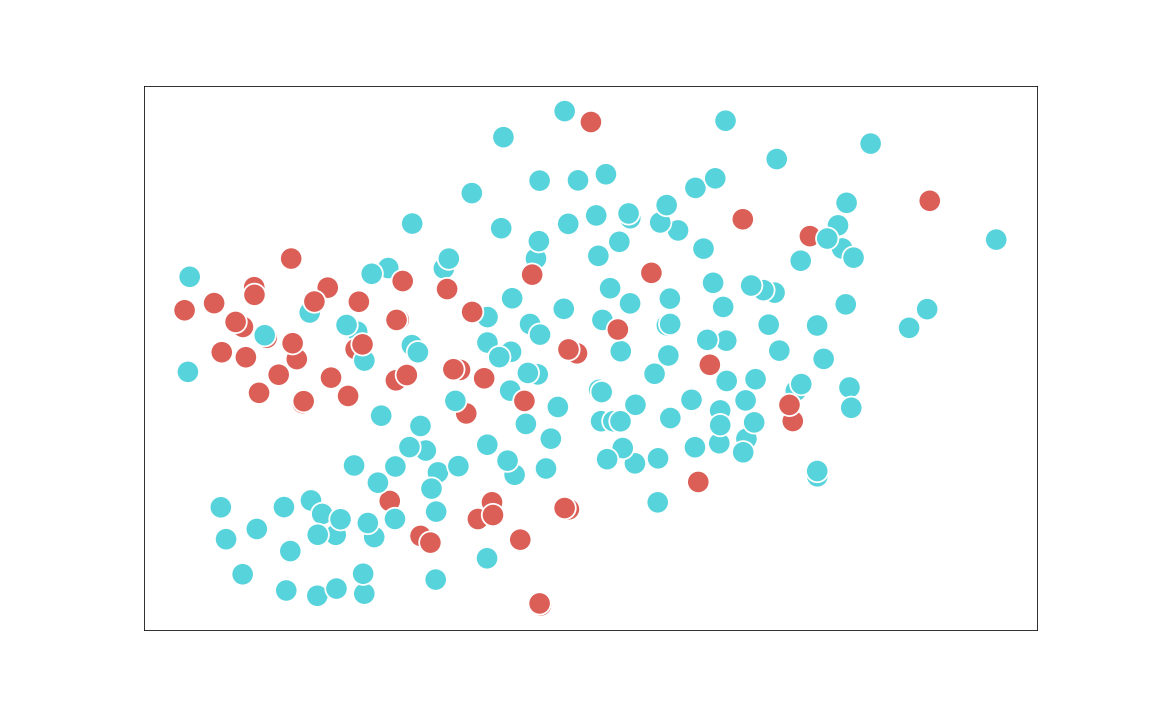}
\caption{$\phi_s$}
\end{subfigure}
\begin{subfigure}[t]{0.32\textwidth}
\centering
\includegraphics[width=\textwidth]{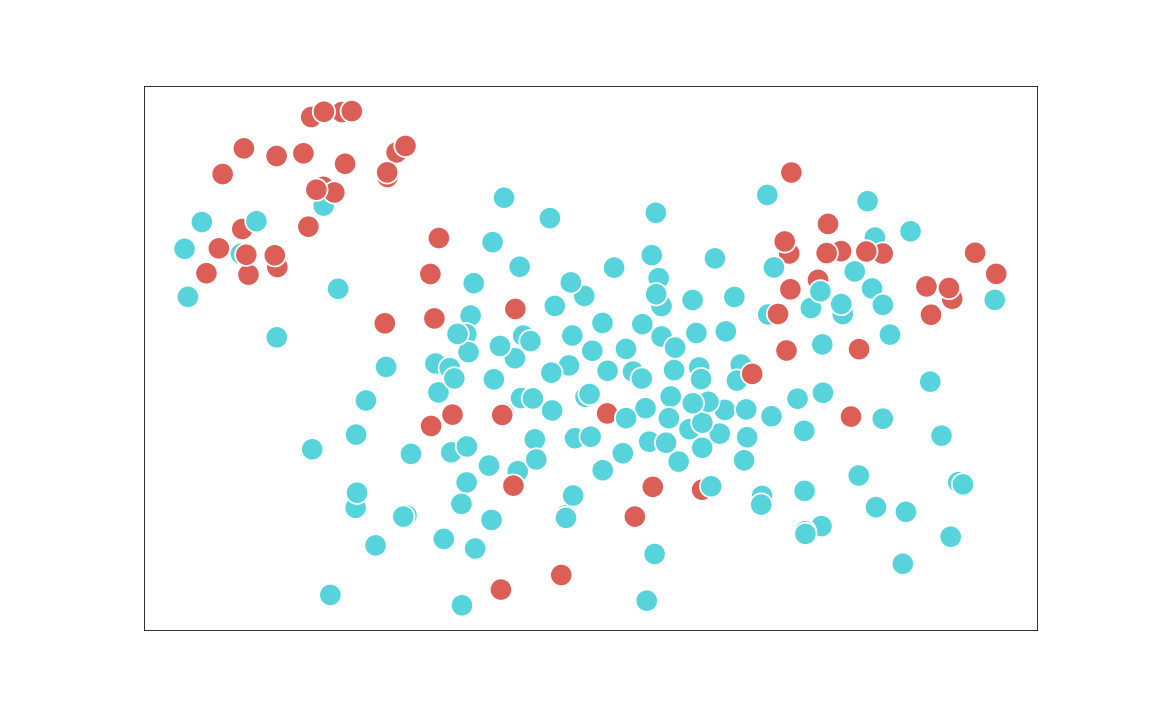}
\caption{$\phi_t^-$}
\end{subfigure}
\begin{subfigure}[t]{0.32\textwidth}
\centering
\includegraphics[width=\textwidth]{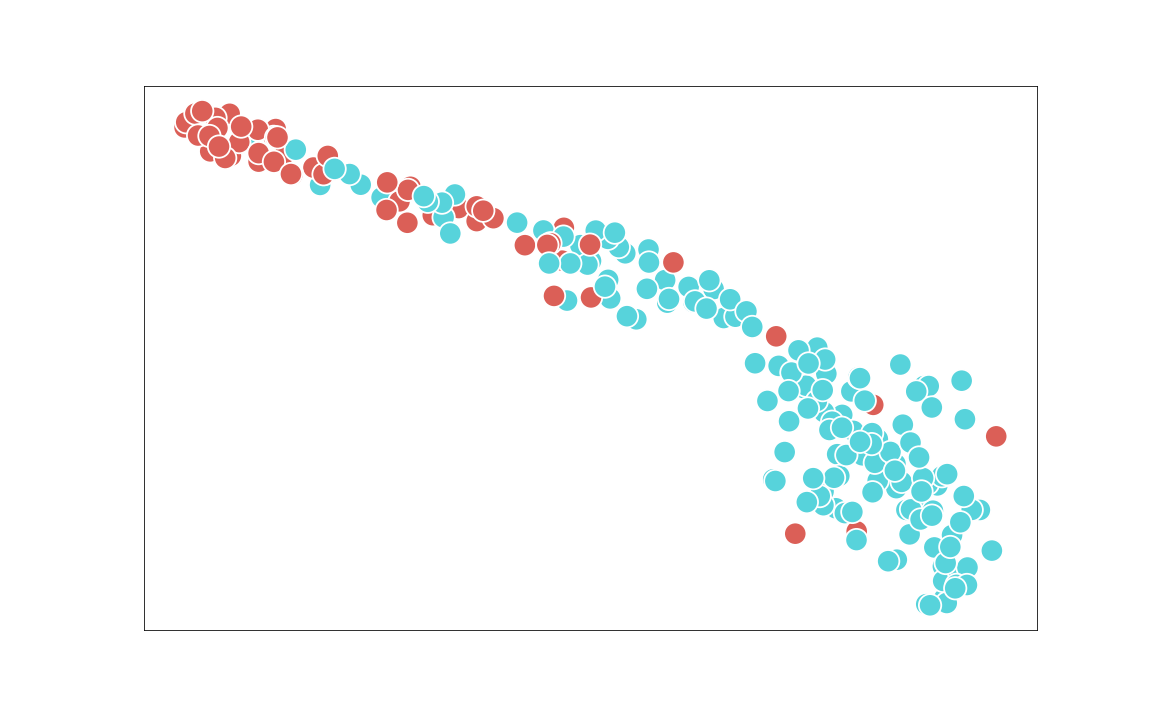}
\caption{$\phi_t$}
\end{subfigure}
\caption{t-SNE plots for ``Explosion'' in UCF Crime}
\end{figure}
\begin{figure}[htbp]
\centering
\begin{subfigure}[t]{0.32\textwidth}
\centering
\includegraphics[width=\textwidth]{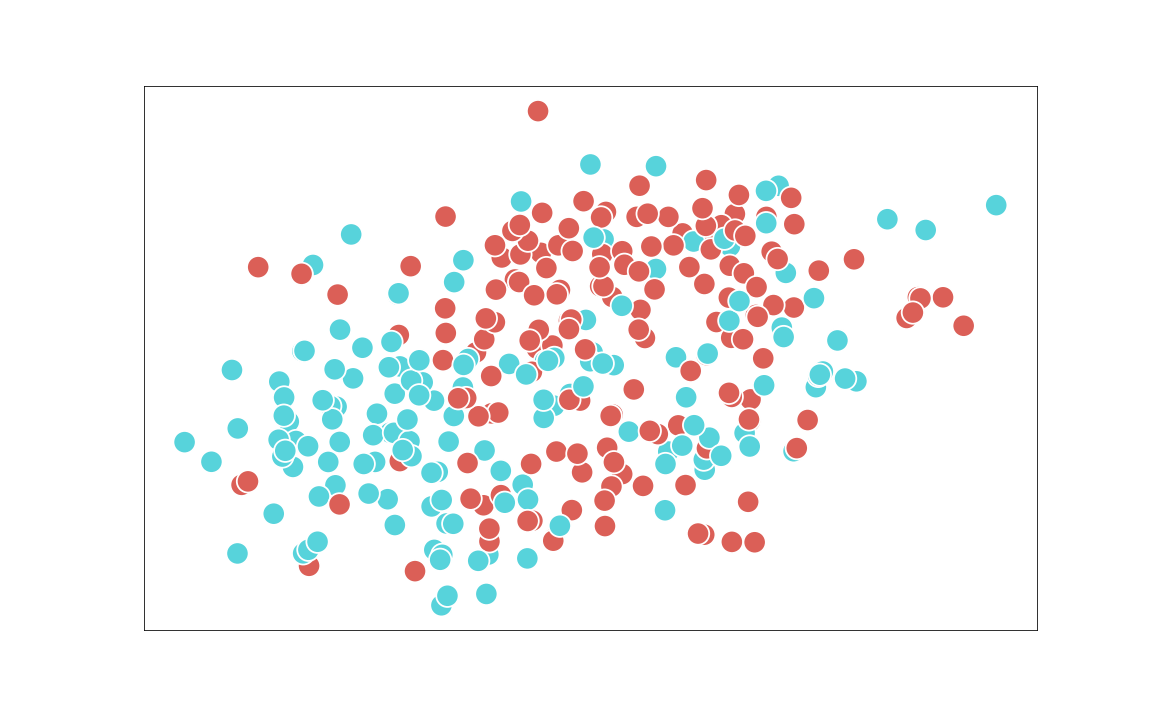}
\caption{$\phi_s$}
\end{subfigure}
\begin{subfigure}[t]{0.32\textwidth}
\centering
\includegraphics[width=\textwidth]{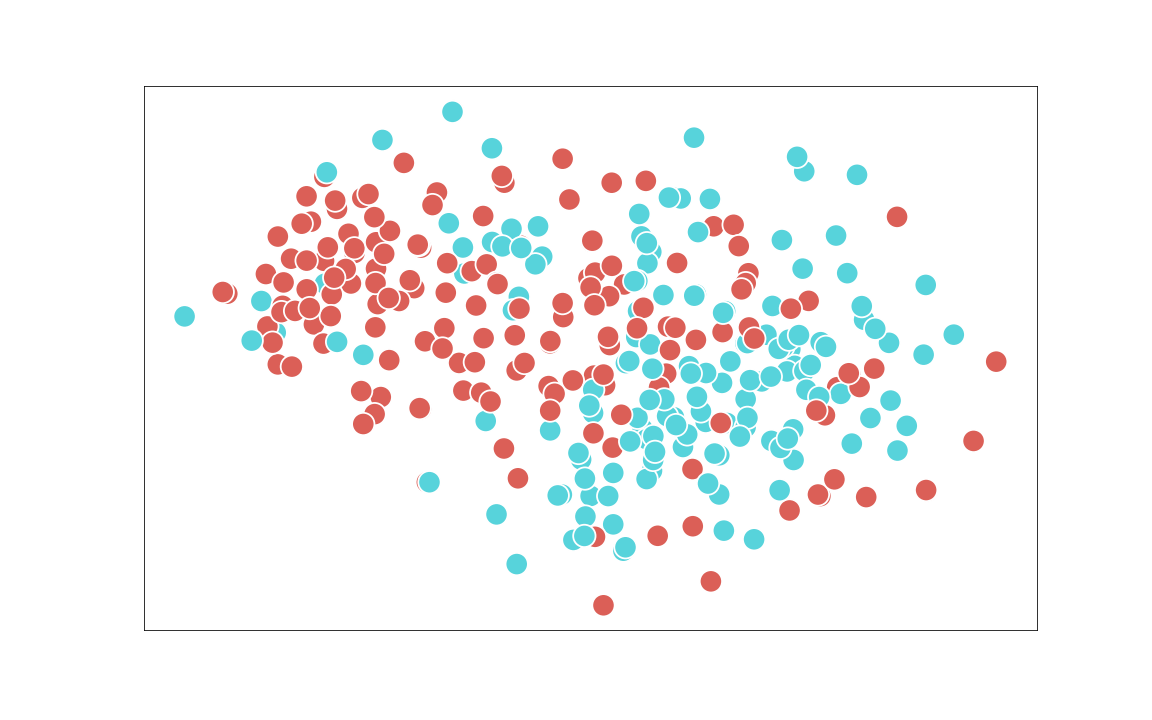}
\caption{$\phi_t^-$}
\end{subfigure}
\begin{subfigure}[t]{0.32\textwidth}
\centering
\includegraphics[width=\textwidth]{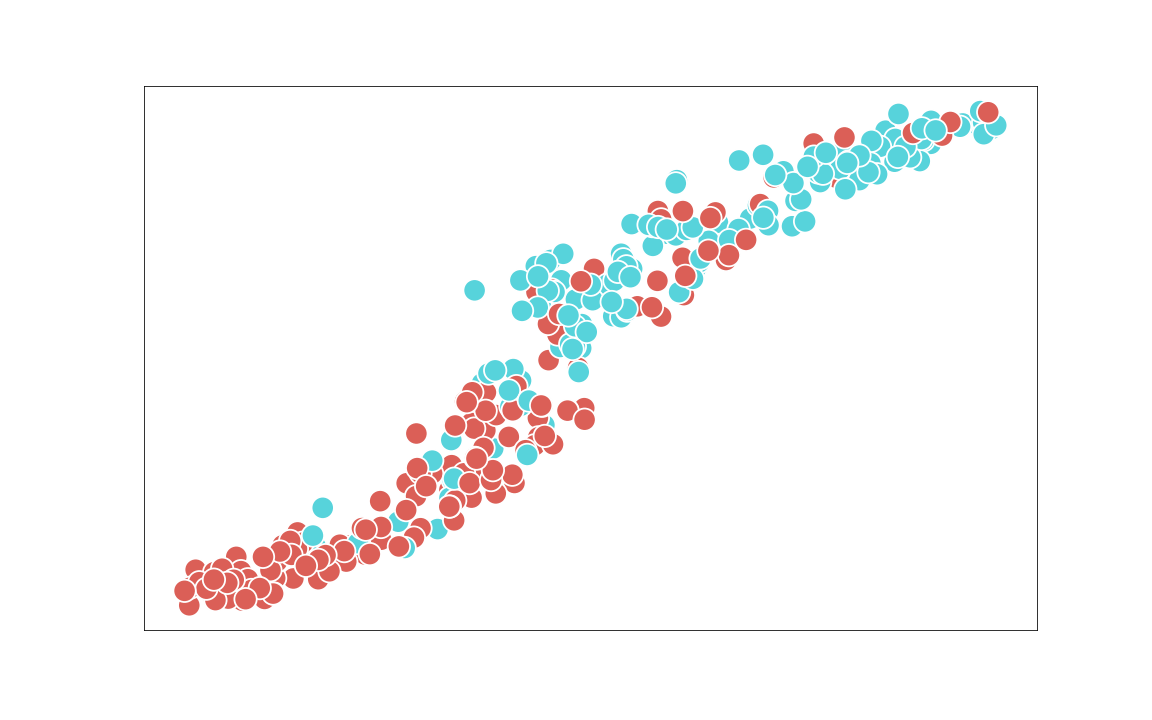}
\caption{$\phi_t$}
\end{subfigure}
\caption{t-SNE plots for ``RoadAccidents'' in UCF Crime}
\end{figure}
\begin{figure}[htbp]
\centering
\begin{subfigure}[t]{0.32\textwidth}
\centering
\includegraphics[width=\textwidth]{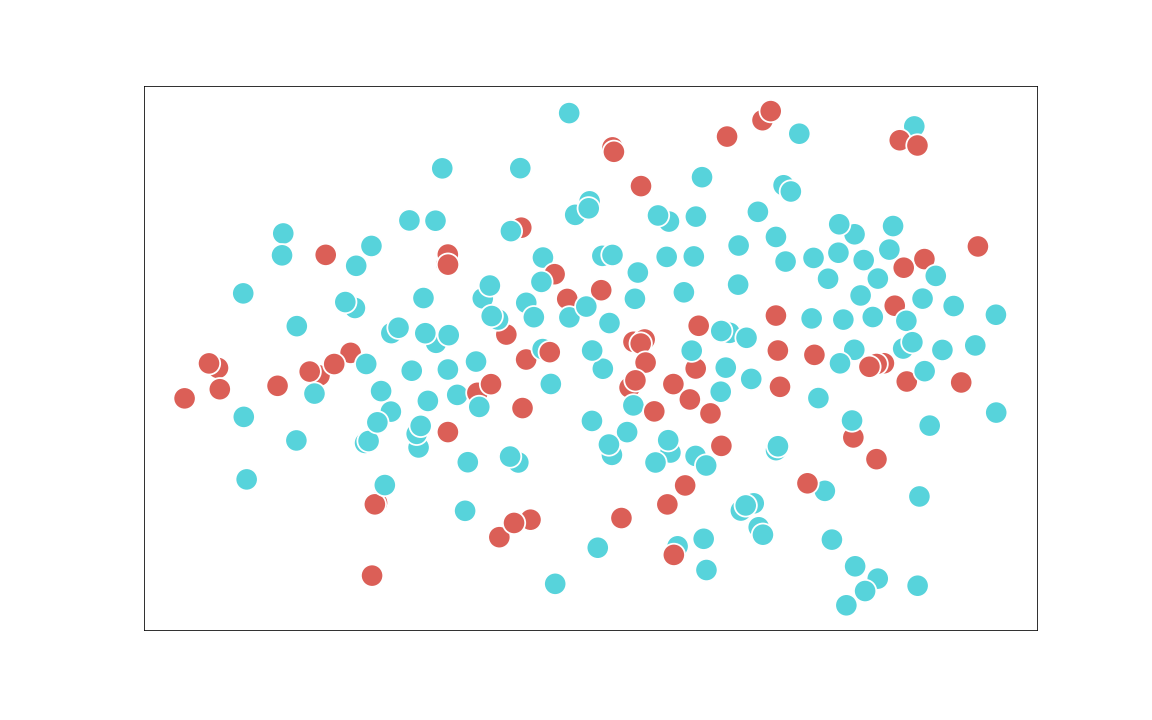}
\caption{$\phi_s$}
\end{subfigure}
\begin{subfigure}[t]{0.32\textwidth}
\centering
\includegraphics[width=\textwidth]{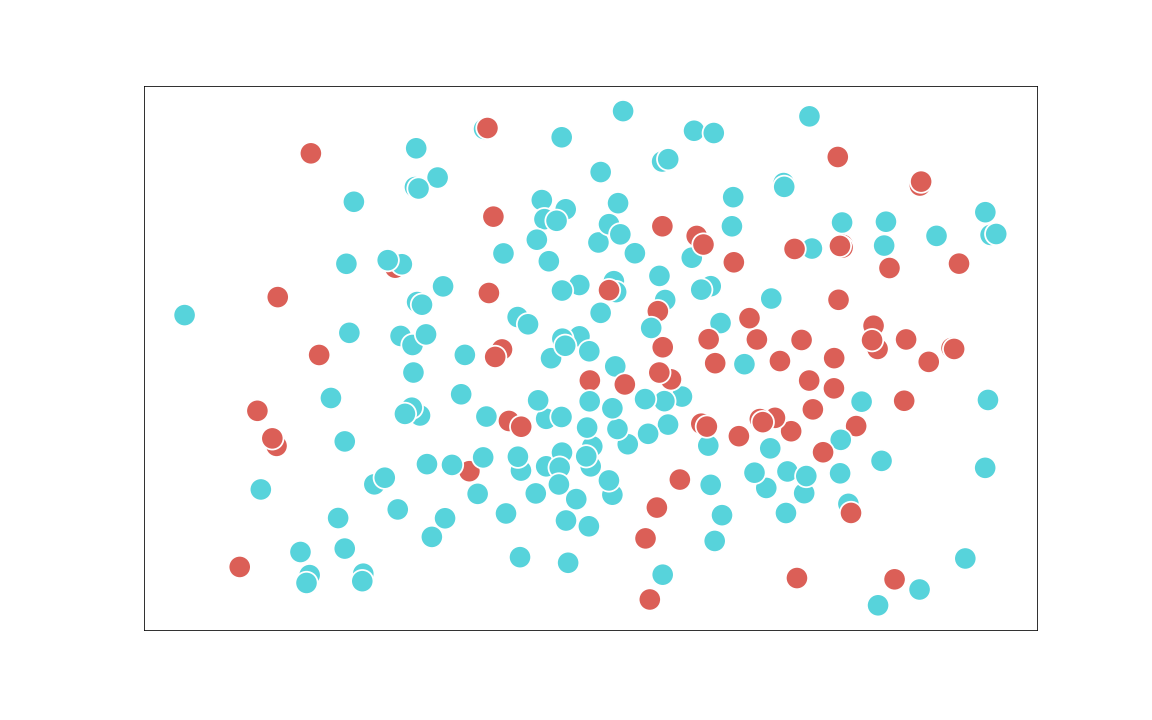}
\caption{$\phi_t^-$}
\end{subfigure}
\begin{subfigure}[t]{0.32\textwidth}
\centering
\includegraphics[width=\textwidth]{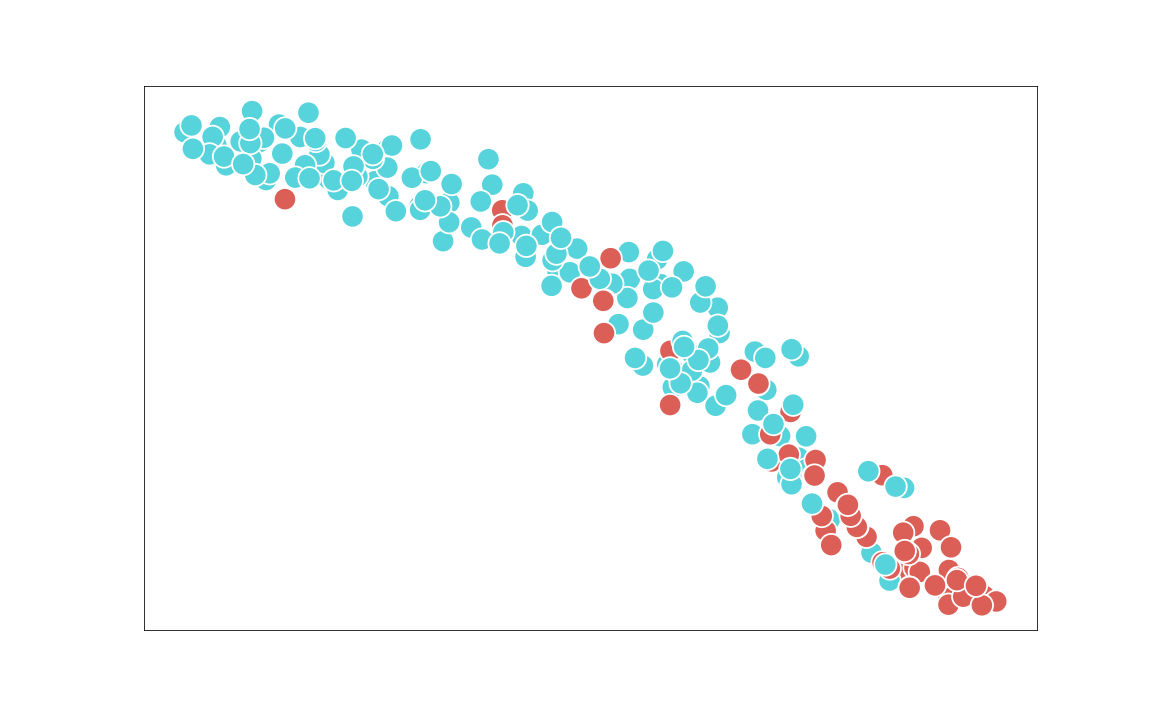}
\caption{$\phi_t$}
\end{subfigure}
\caption{t-SNE plots for ``Shooting'' in UCF Crime}
\end{figure}
\begin{figure}[htbp]
\centering
\begin{subfigure}[t]{0.32\textwidth}
\centering
\includegraphics[width=\textwidth]{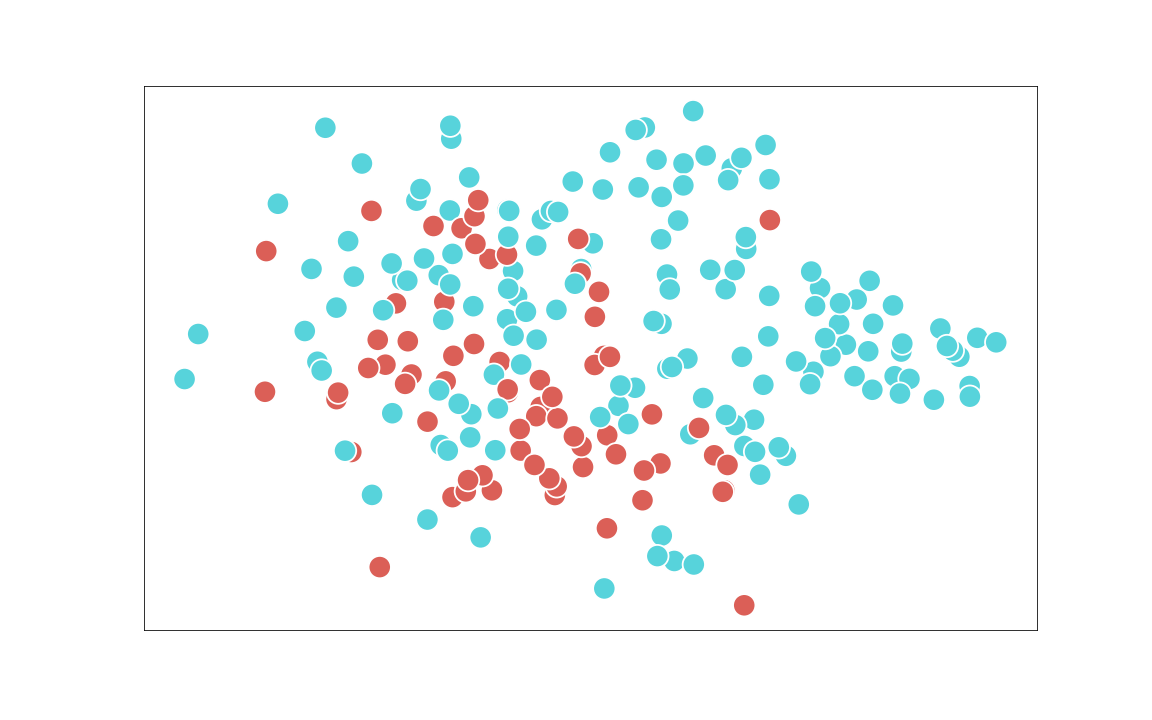}
\caption{$\phi_s$}
\end{subfigure}
\begin{subfigure}[t]{0.32\textwidth}
\centering
\includegraphics[width=\textwidth]{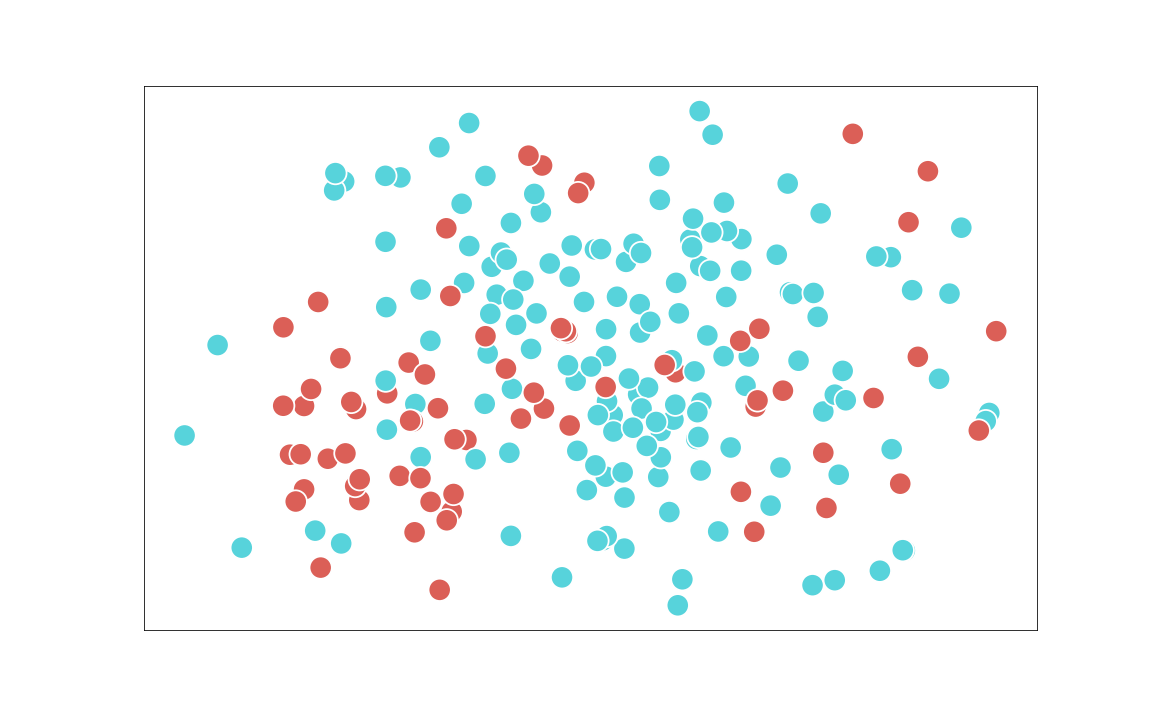}
\caption{$\phi_t^-$}
\end{subfigure}
\begin{subfigure}[t]{0.32\textwidth}
\centering
\includegraphics[width=\textwidth]{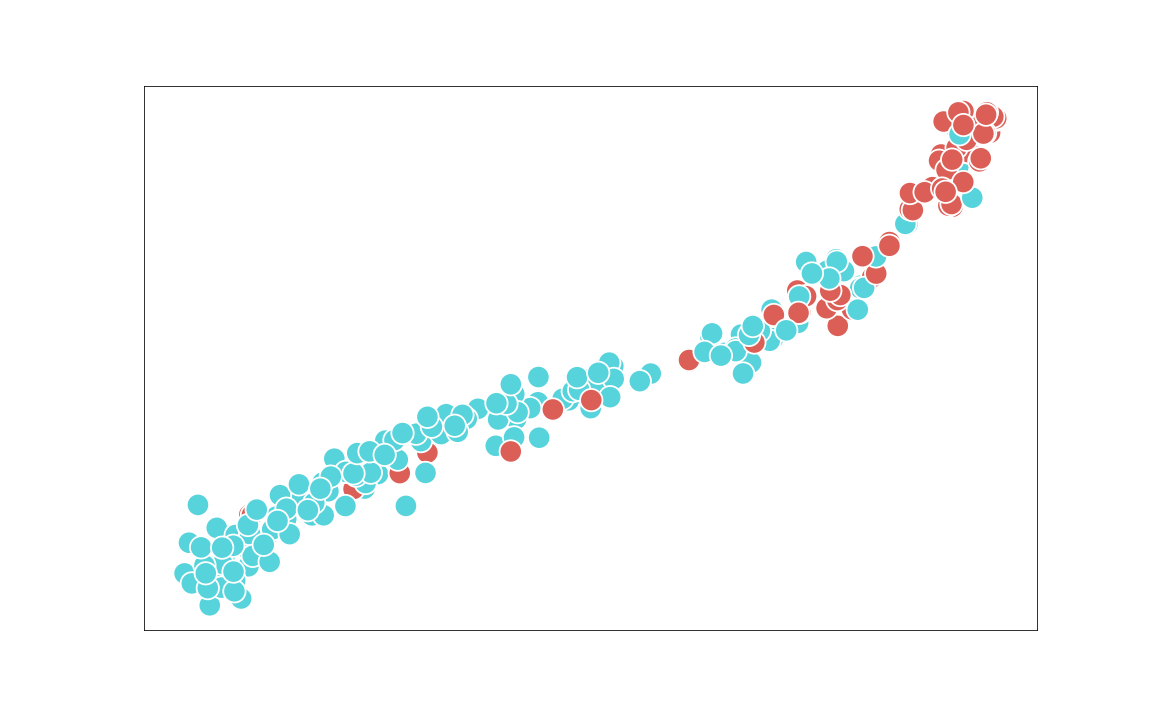}
\caption{$\phi_t$}
\end{subfigure}
\caption{t-SNE plots for ``Shoplifting'' in UCF Crime}
\end{figure}
\begin{figure}[htbp]
\centering
\begin{subfigure}[t]{0.32\textwidth}
\centering
\includegraphics[width=\textwidth]{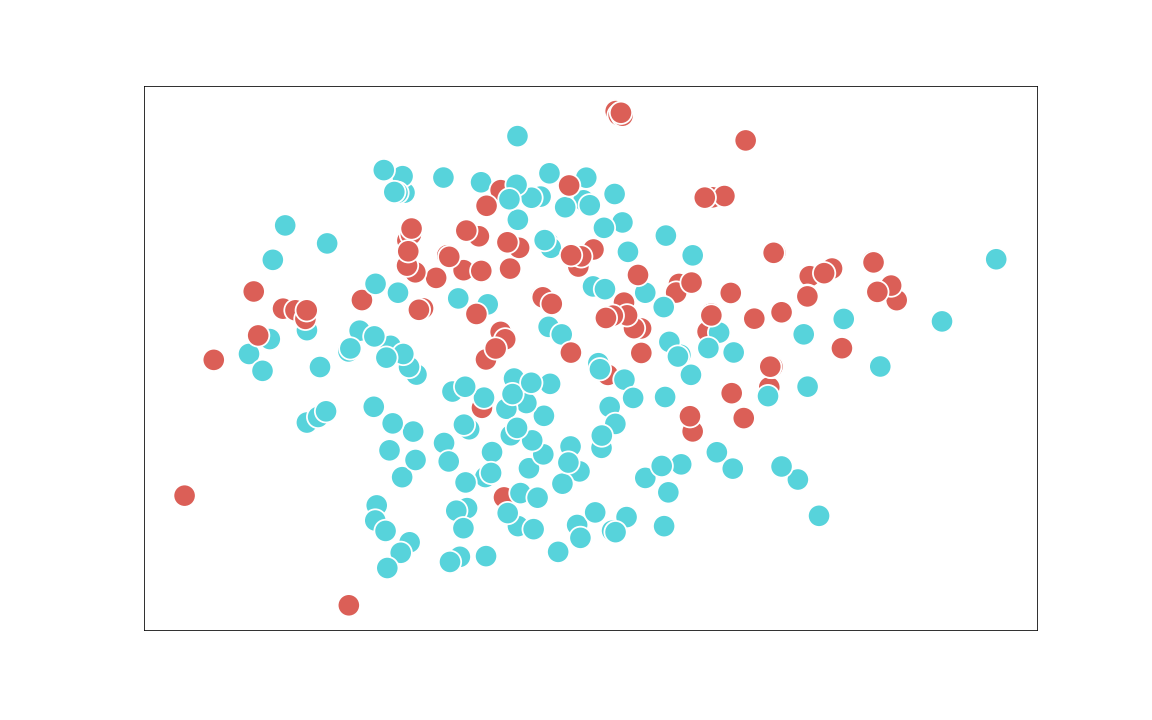}
\caption{$\phi_s$}
\end{subfigure}
\begin{subfigure}[t]{0.32\textwidth}
\centering
\includegraphics[width=\textwidth]{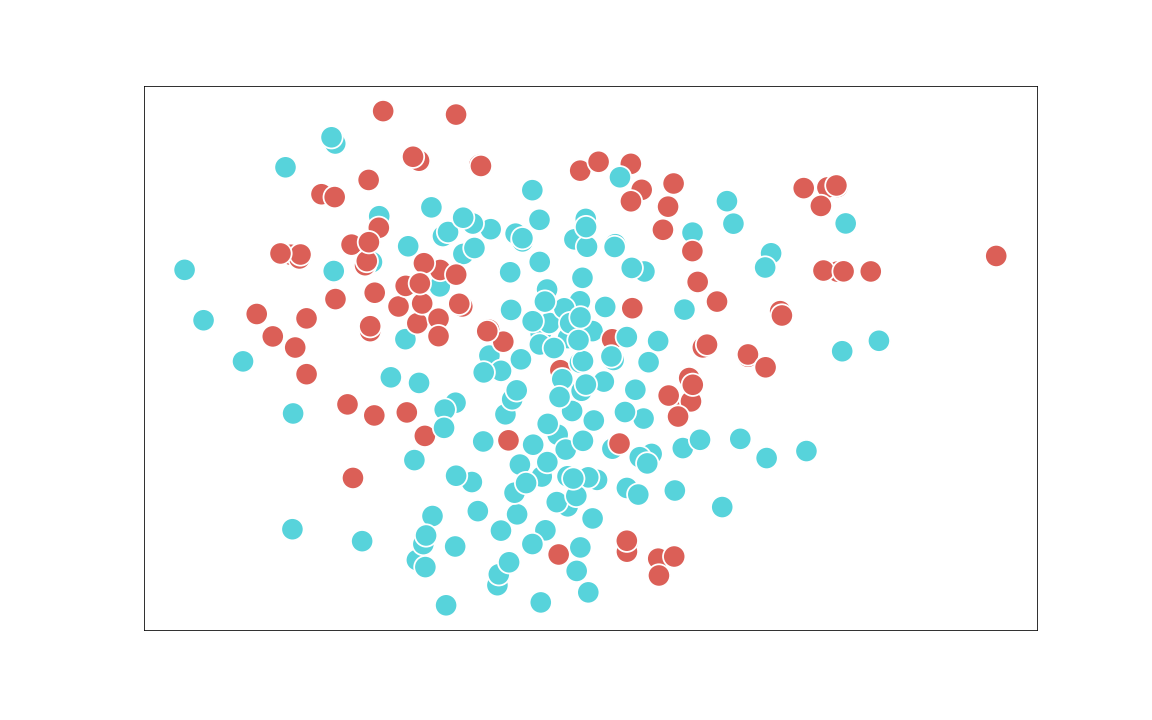}
\caption{$\phi_t^-$}
\end{subfigure}
\begin{subfigure}[t]{0.32\textwidth}
\centering
\includegraphics[width=\textwidth]{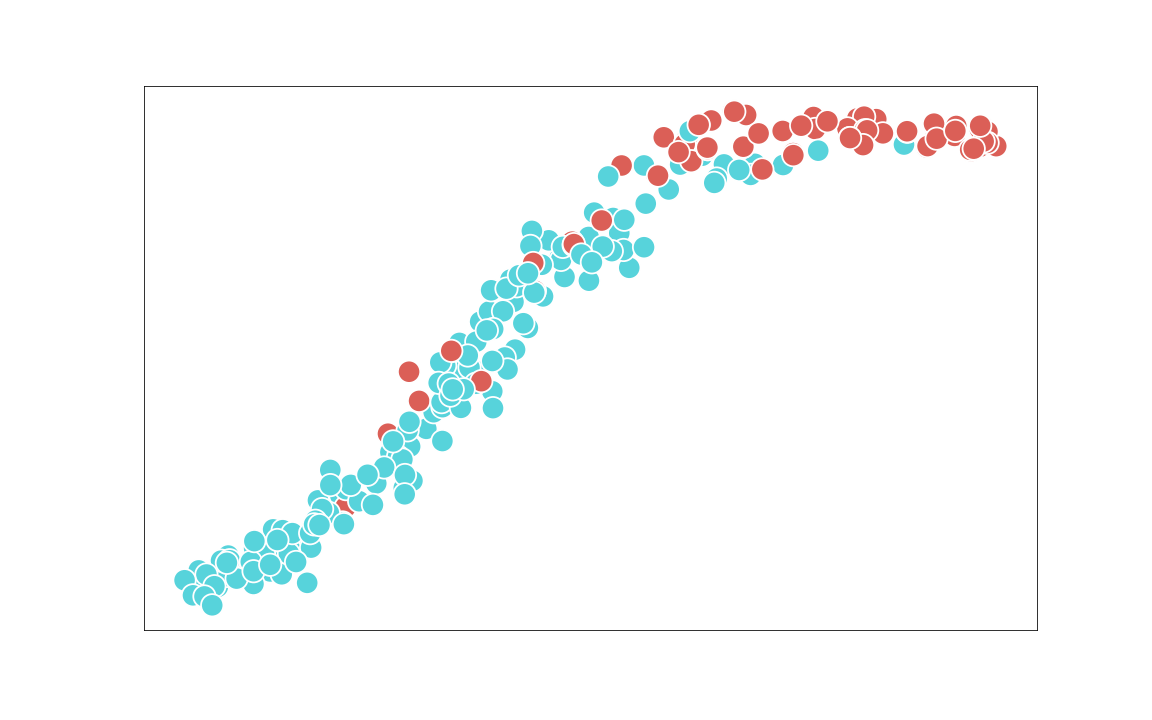}
\caption{$\phi_t$}
\end{subfigure}
\caption{t-SNE plots for ``Abuse'' in UCF Crime}
\end{figure}
\begin{figure}[htbp]
\centering
\begin{subfigure}[t]{0.32\textwidth}
\centering
\includegraphics[width=\textwidth]{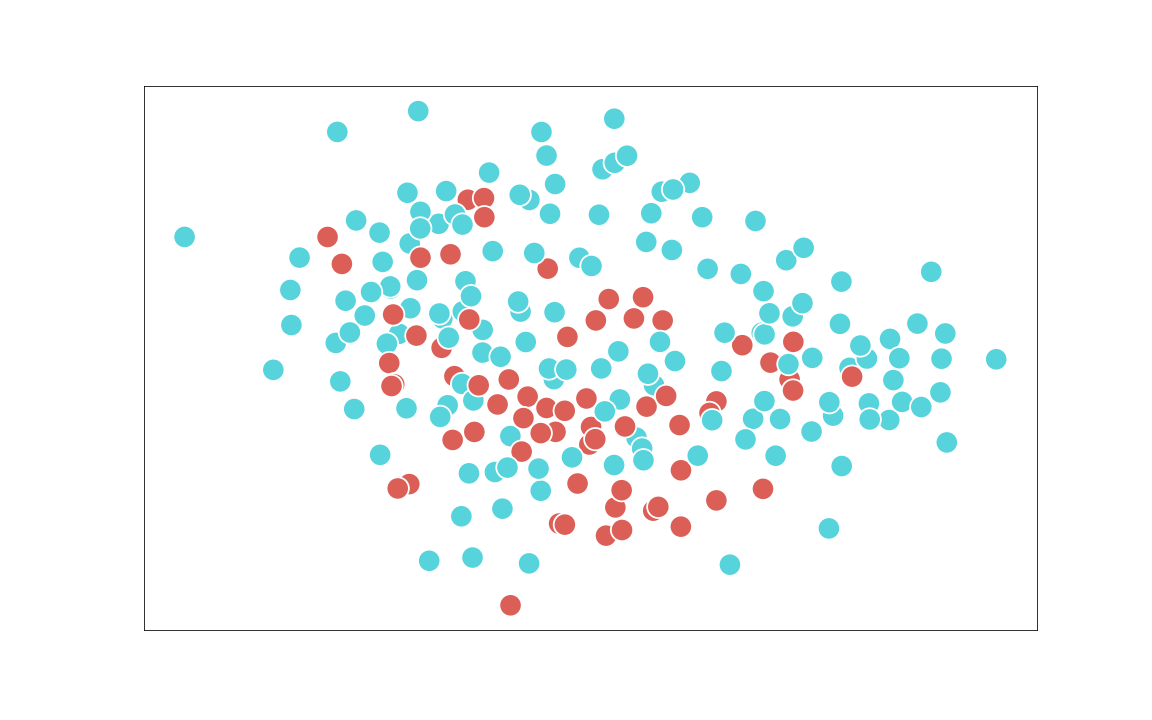}
\caption{$\phi_s$}
\end{subfigure}
\begin{subfigure}[t]{0.32\textwidth}
\centering
\includegraphics[width=\textwidth]{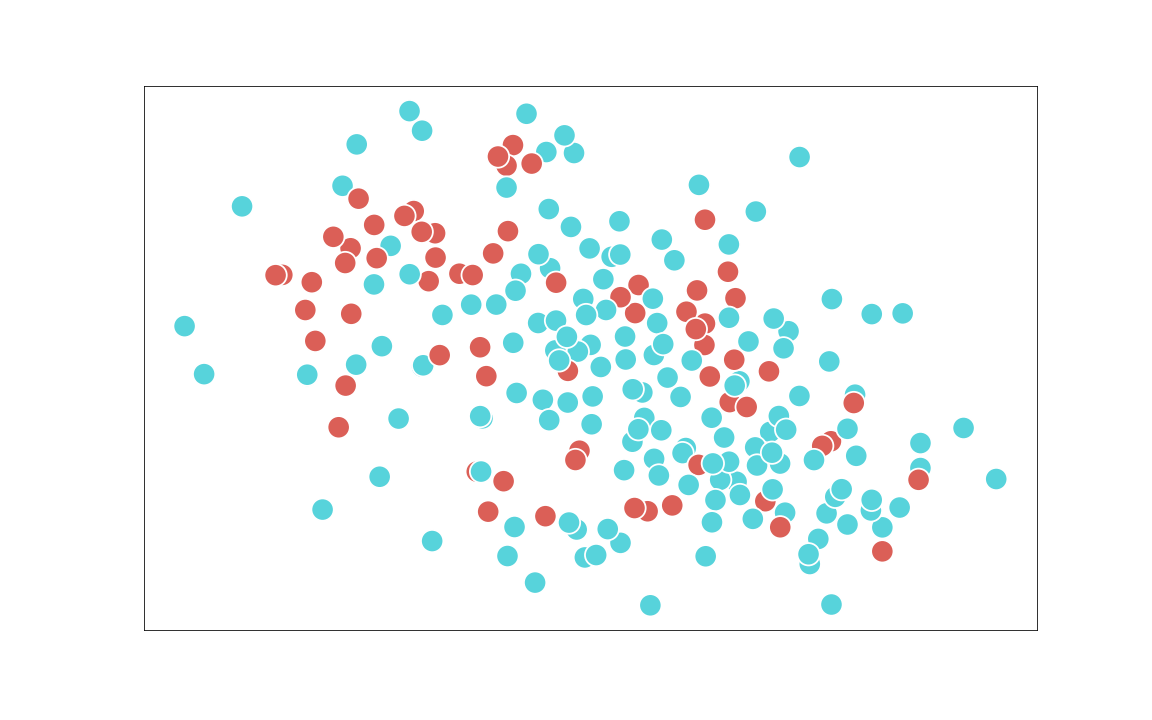}
\caption{$\phi_t^-$}
\end{subfigure}
\begin{subfigure}[t]{0.32\textwidth}
\centering
\includegraphics[width=\textwidth]{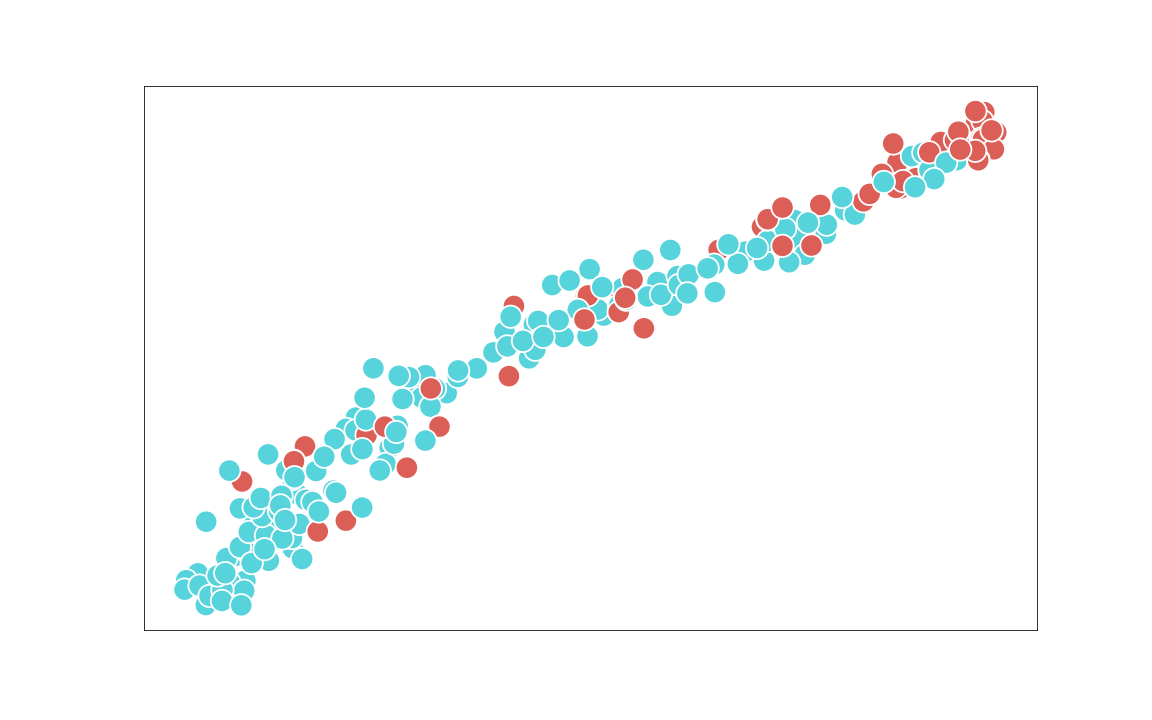}
\caption{$\phi_t$}
\end{subfigure}
\caption{t-SNE plots for ``Arrest'' in UCF Crime}
\end{figure}
\begin{figure}[htbp]
\centering
\begin{subfigure}[t]{0.32\textwidth}
\centering
\includegraphics[width=\textwidth]{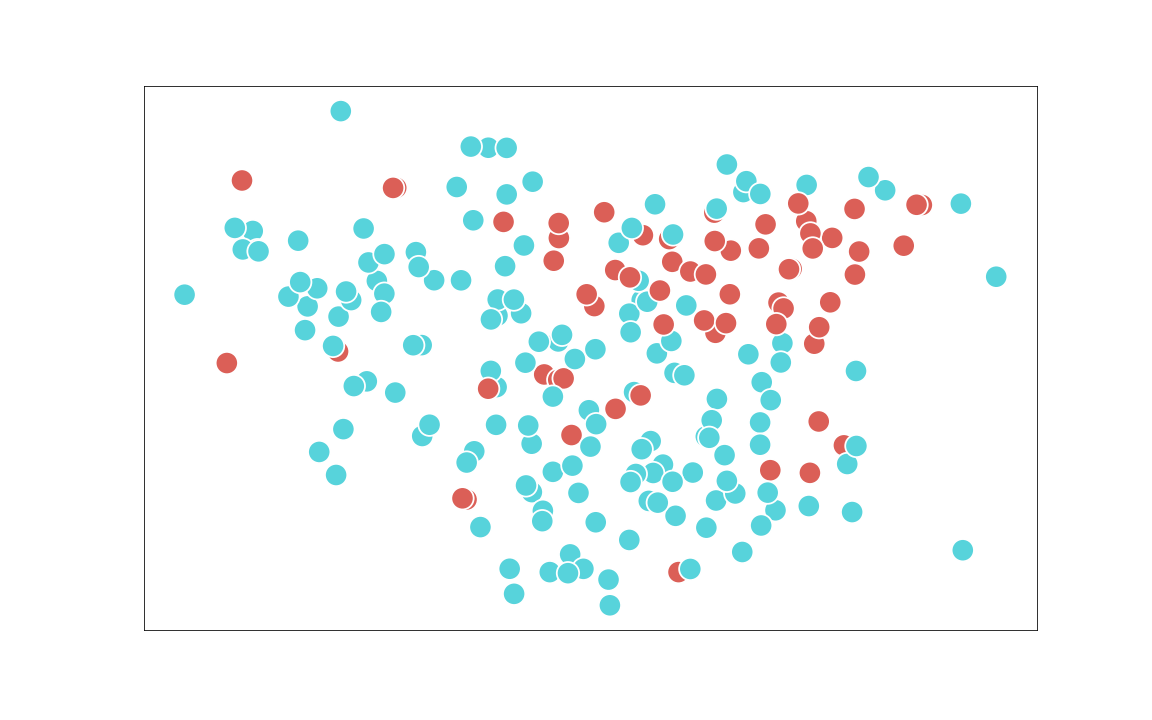}
\caption{$\phi_s$}
\end{subfigure}
\begin{subfigure}[t]{0.32\textwidth}
\centering
\includegraphics[width=\textwidth]{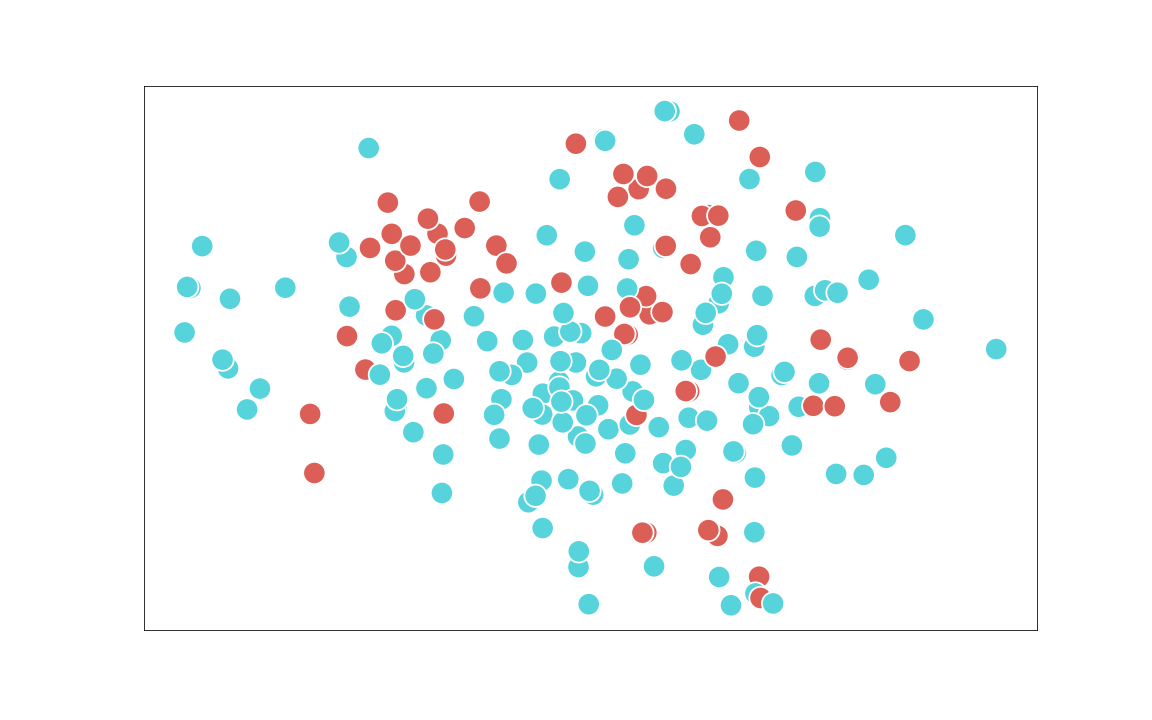}
\caption{$\phi_t^-$}
\end{subfigure}
\begin{subfigure}[t]{0.32\textwidth}
\centering
\includegraphics[width=\textwidth]{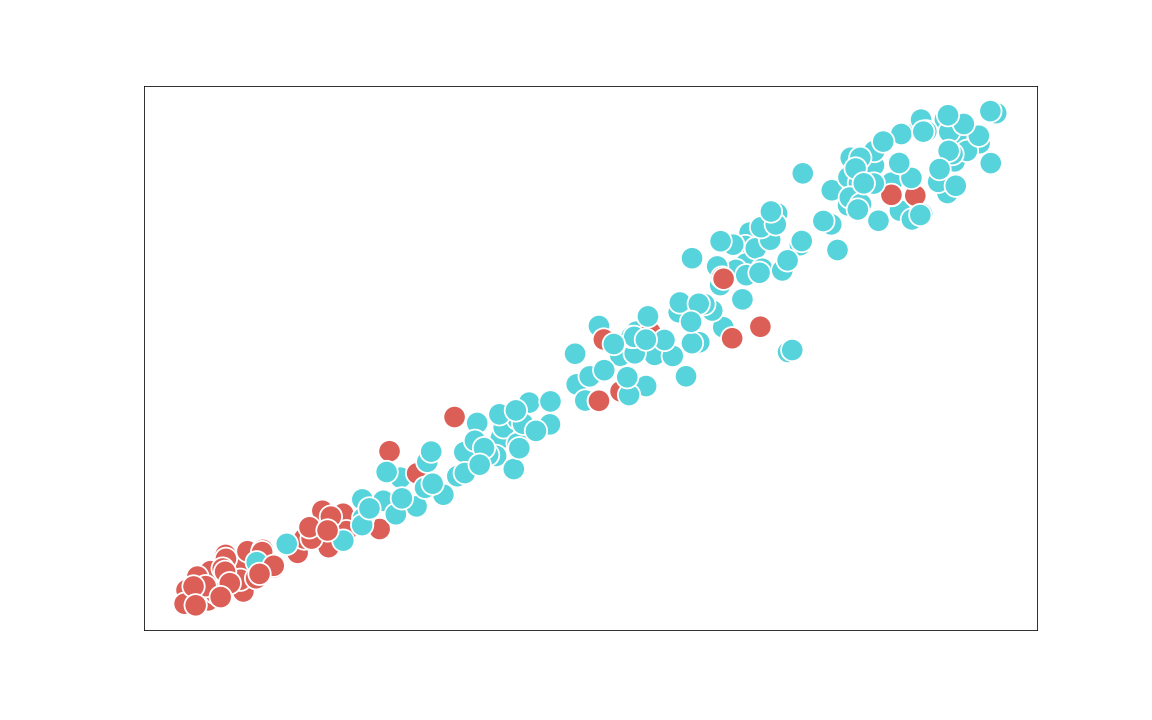}
\caption{$\phi_t$}
\end{subfigure}
\caption{t-SNE plots for ``Assault'' in UCF Crime}
\end{figure}
\begin{figure}[htbp]
\centering
\begin{subfigure}[t]{0.32\textwidth}
\centering
\includegraphics[width=\textwidth]{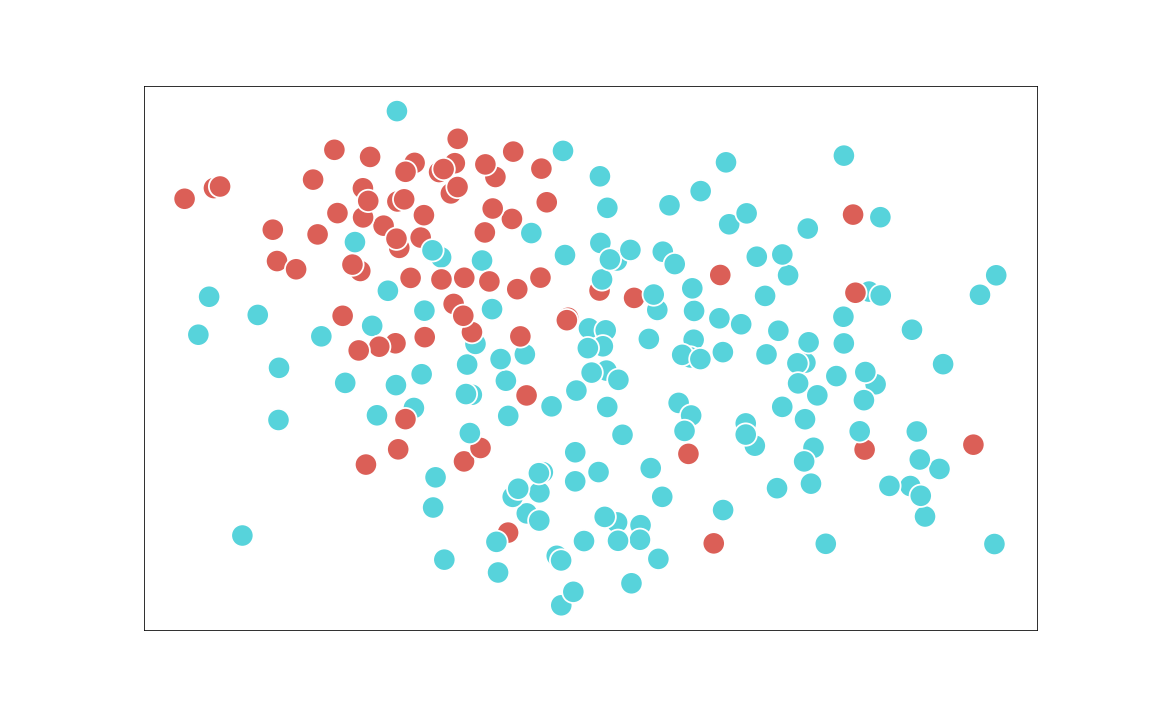}
\caption{$\phi_s$}
\end{subfigure}
\begin{subfigure}[t]{0.32\textwidth}
\centering
\includegraphics[width=\textwidth]{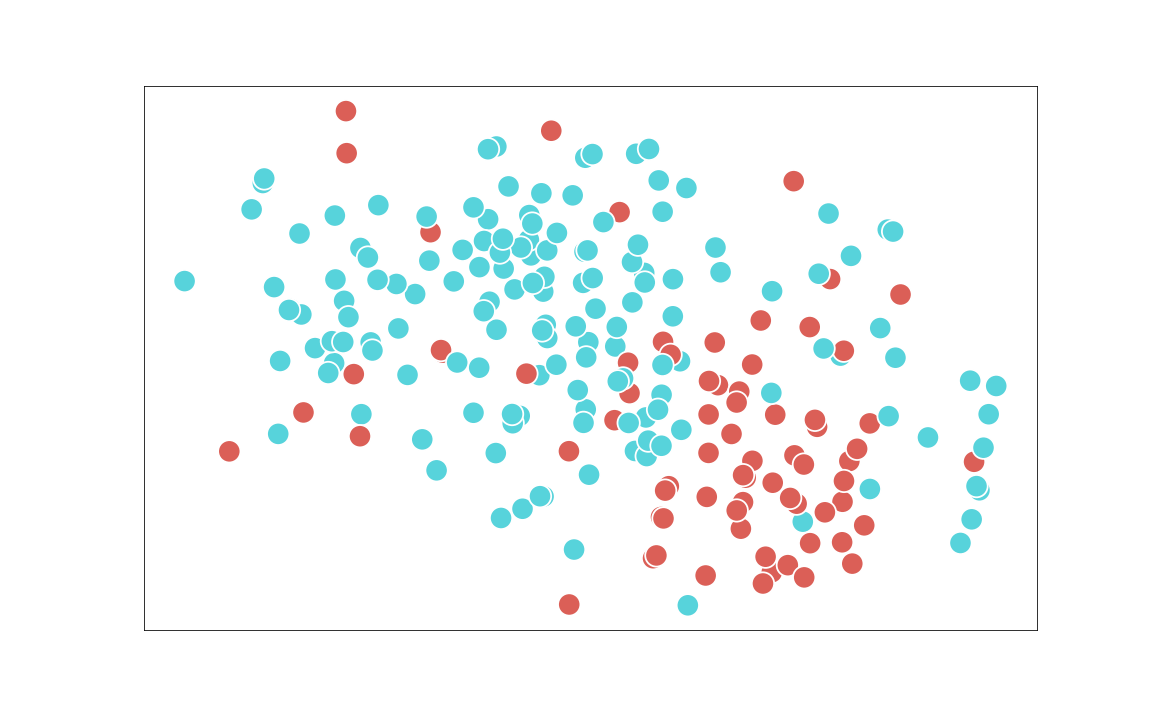}
\caption{$\phi_t^-$}
\end{subfigure}
\begin{subfigure}[t]{0.32\textwidth}
\centering
\includegraphics[width=\textwidth]{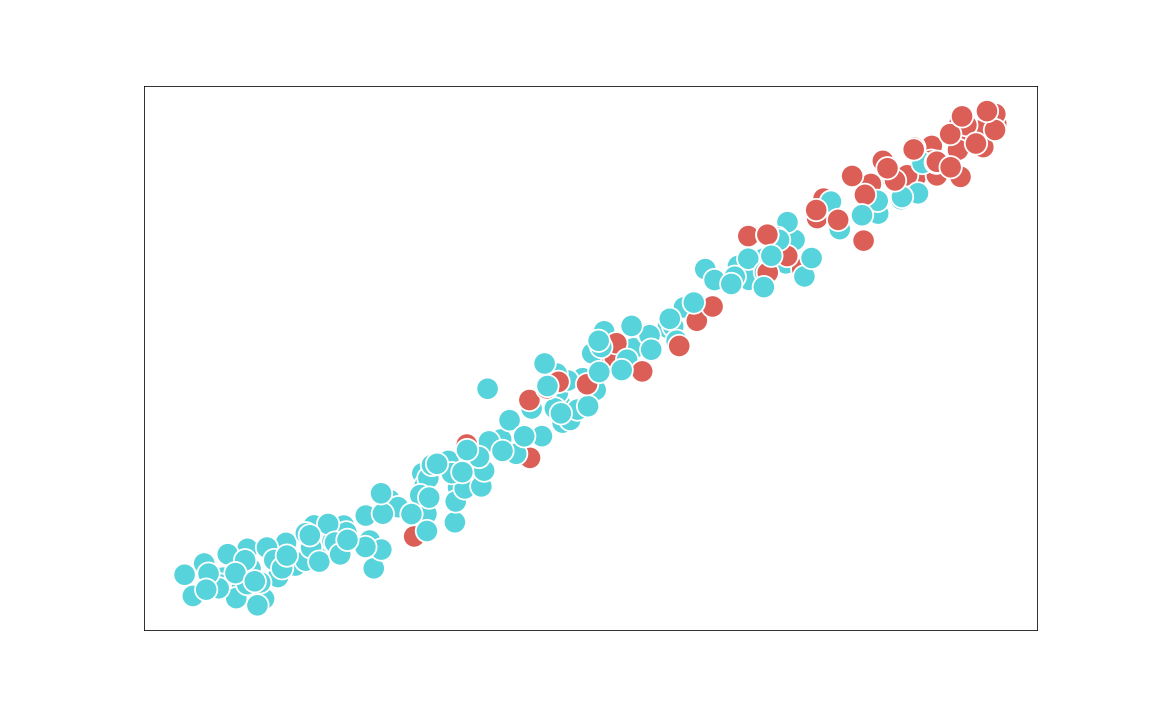}
\caption{$\phi_t$}
\end{subfigure}
\caption{t-SNE plots for ``Fighting'' in UCF Crime}
\end{figure}
\begin{figure}[htbp]
\centering
\begin{subfigure}[t]{0.32\textwidth}
\centering
\includegraphics[width=\textwidth]{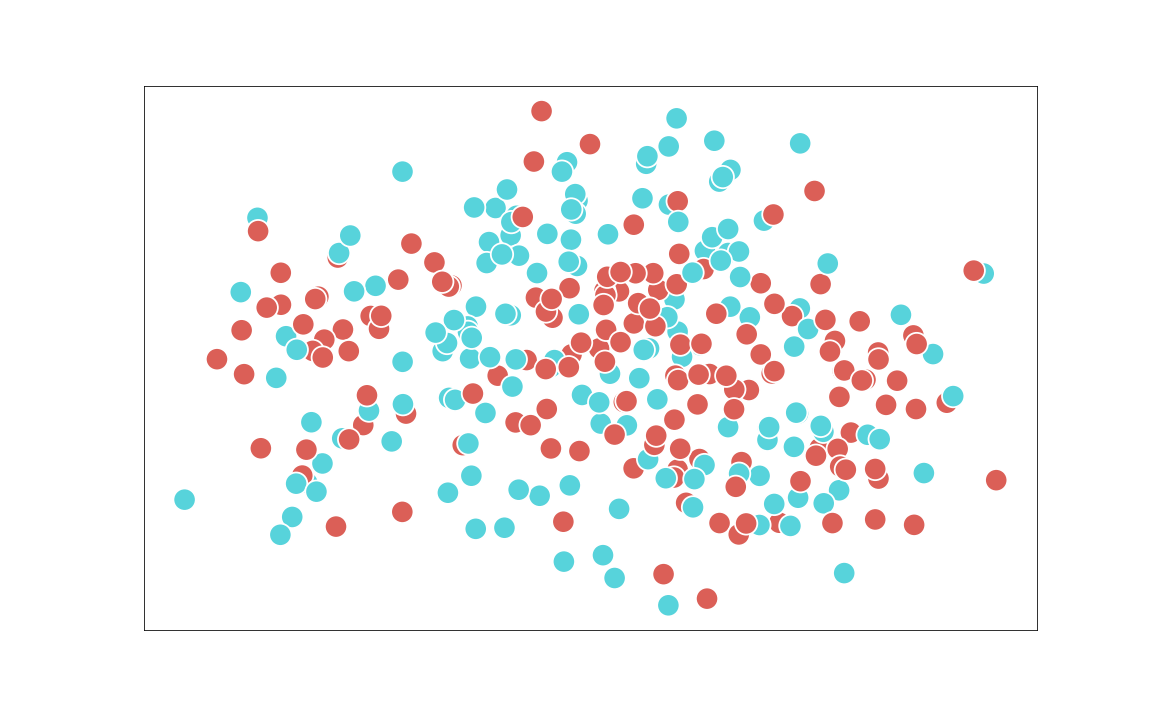}
\caption{$\phi_s$}
\end{subfigure}
\begin{subfigure}[t]{0.32\textwidth}
\centering
\includegraphics[width=\textwidth]{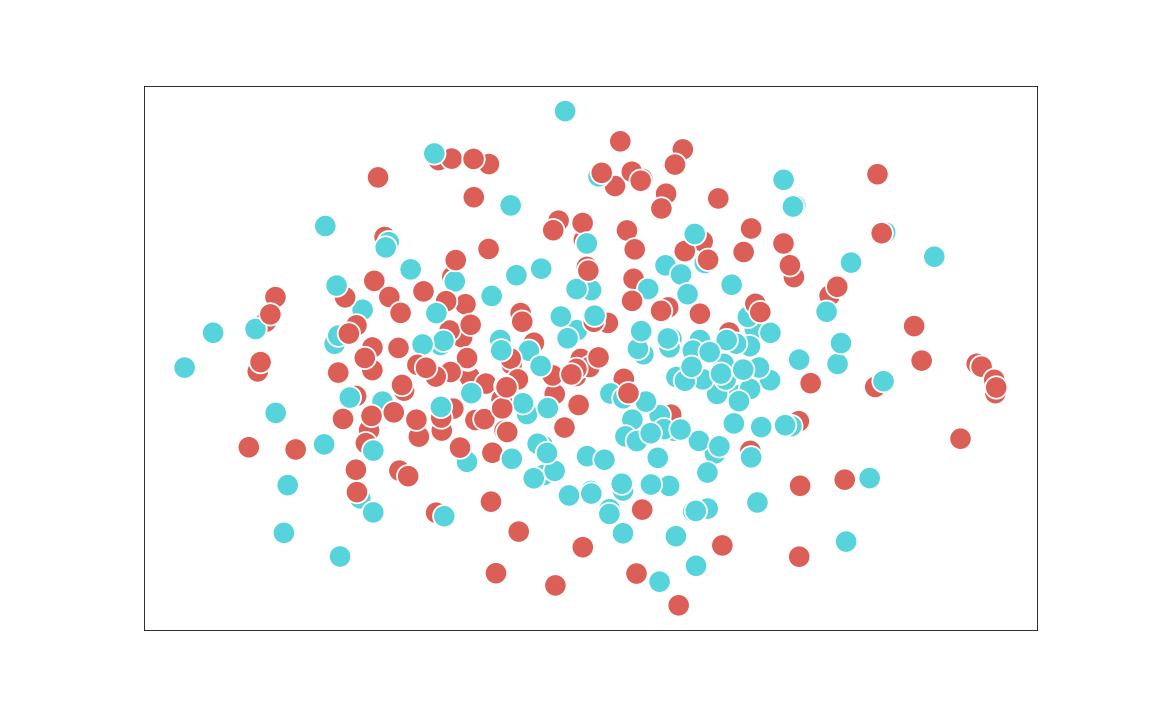}
\caption{$\phi_t^-$}
\end{subfigure}
\begin{subfigure}[t]{0.32\textwidth}
\centering
\includegraphics[width=\textwidth]{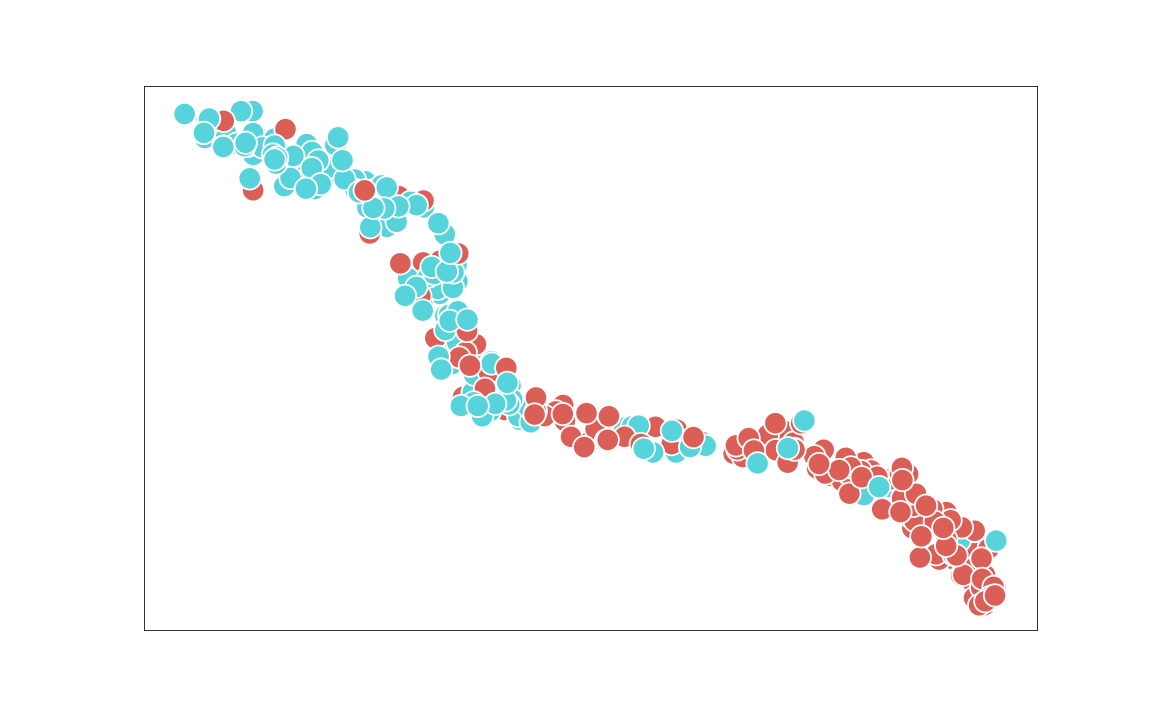}
\caption{$\phi_t$}
\end{subfigure}
\caption{t-SNE plots for ``Robbery'' in UCF Crime}
\end{figure}
\begin{figure}[htbp]
\centering
\begin{subfigure}[t]{0.32\textwidth}
\centering
\includegraphics[width=\textwidth]{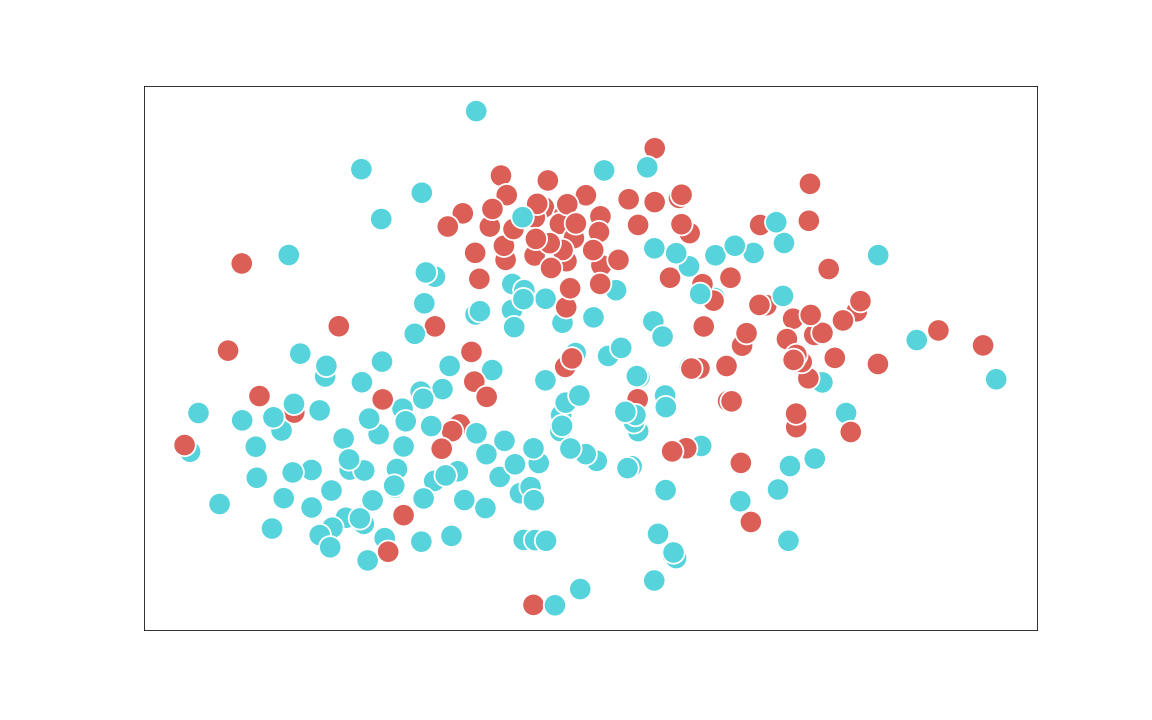}
\caption{$\phi_s$}
\end{subfigure}
\begin{subfigure}[t]{0.32\textwidth}
\centering
\includegraphics[width=\textwidth]{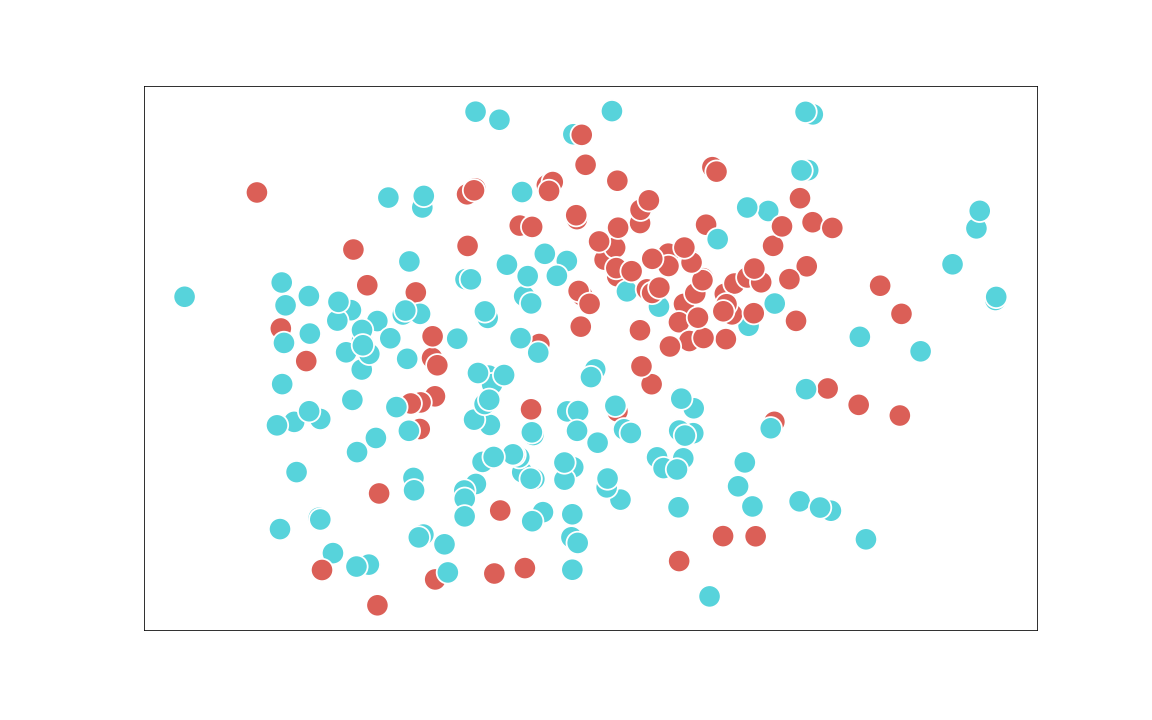}
\caption{$\phi_t^-$}
\end{subfigure}
\begin{subfigure}[t]{0.32\textwidth}
\centering
\includegraphics[width=\textwidth]{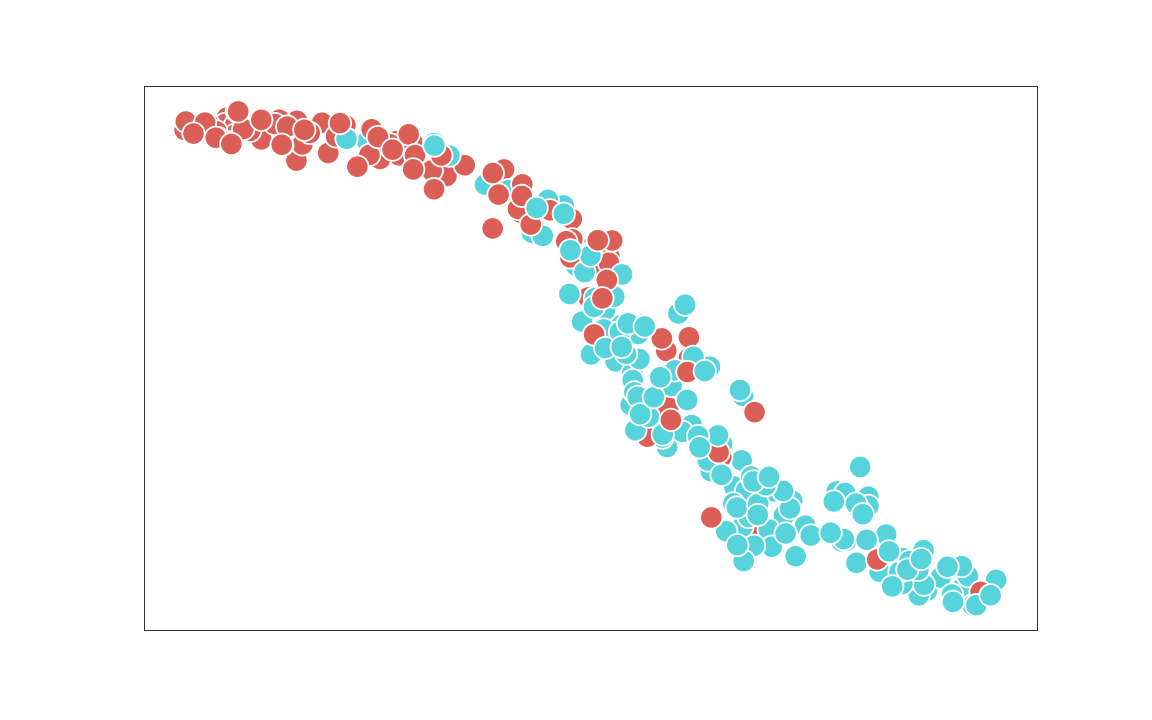}
\caption{$\phi_t$}
\end{subfigure}
\caption{t-SNE plots for ``Stealing'' in UCF Crime}
\end{figure}
\begin{figure}[htbp]
\centering
\begin{subfigure}[t]{0.32\textwidth}
\centering
\includegraphics[width=\textwidth]{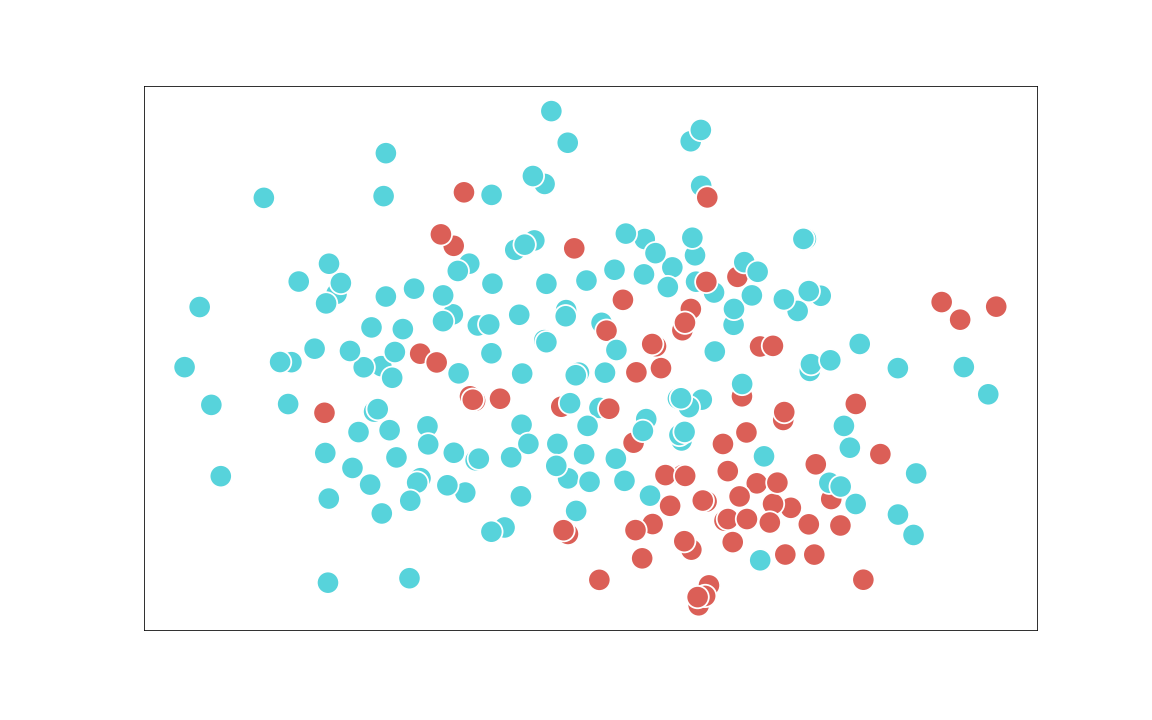}
\caption{$\phi_s$}
\end{subfigure}
\begin{subfigure}[t]{0.32\textwidth}
\centering
\includegraphics[width=\textwidth]{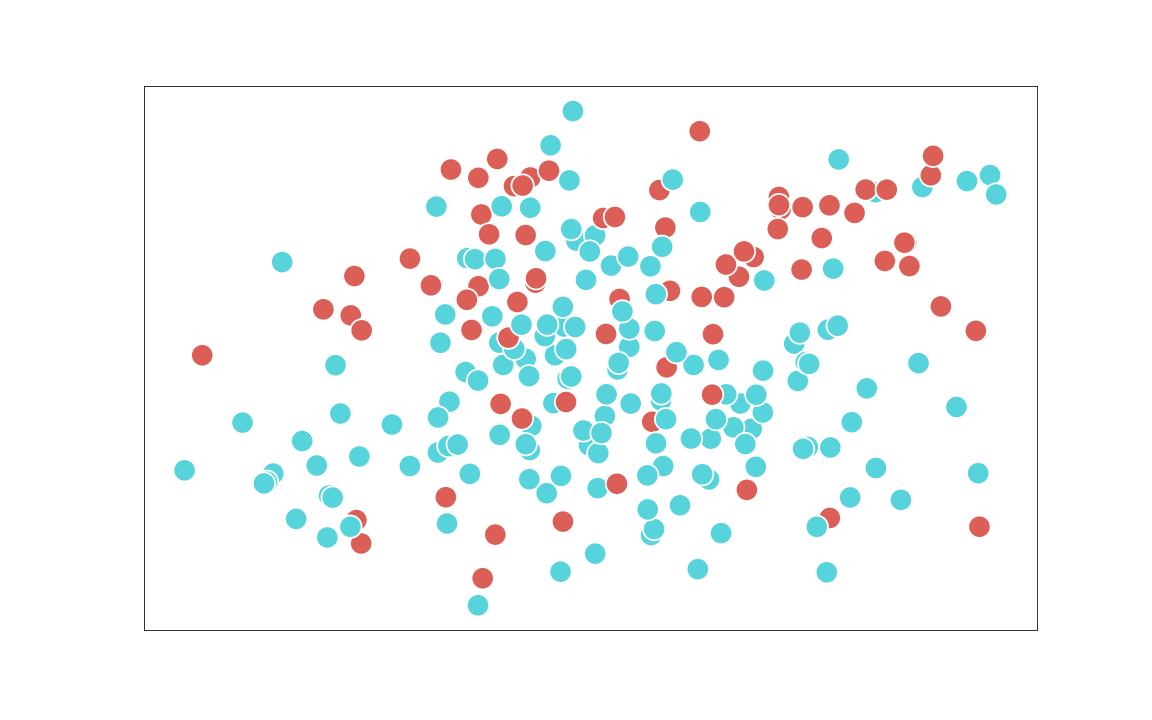}
\caption{$\phi_t^-$}
\end{subfigure}
\begin{subfigure}[t]{0.32\textwidth}
\centering
\includegraphics[width=\textwidth]{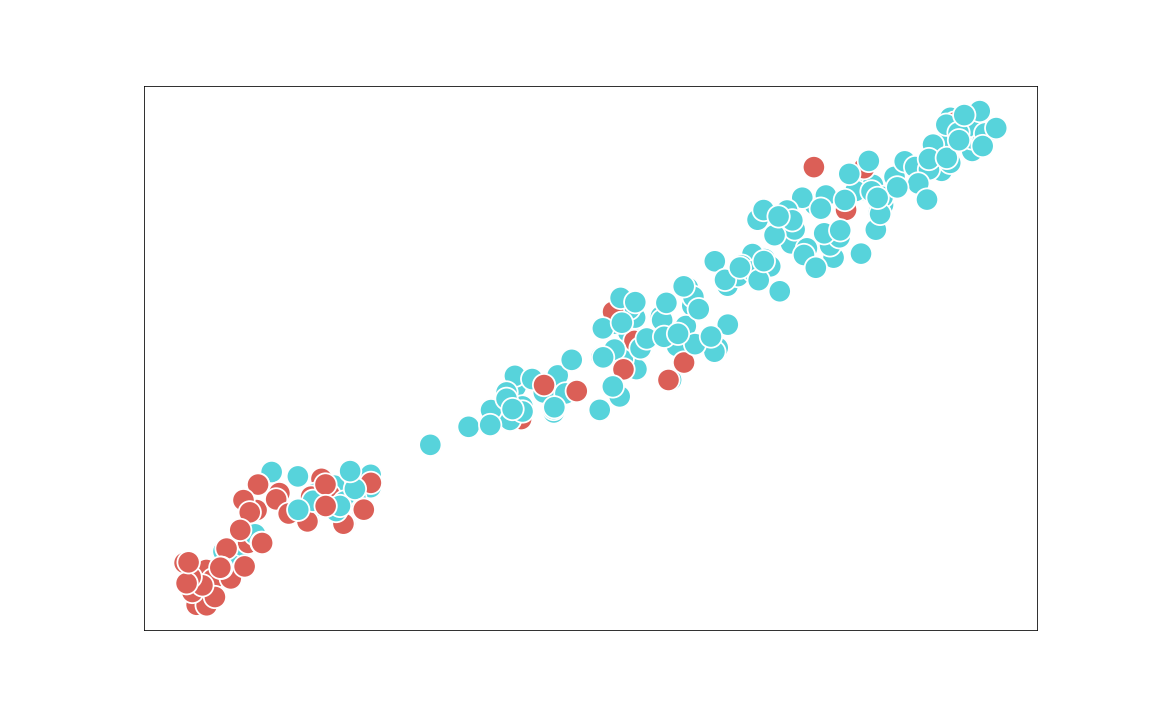}
\caption{$\phi_t$}
\end{subfigure}
\caption{t-SNE plots for ``Vandalism'' in UCF Crime}
\end{figure}

 \end{document}